\documentclass[lettersize,journal]{IEEEtran}
\usepackage{epsfig}
\usepackage{graphicx}
\usepackage{amsmath}
\usepackage{amssymb}
\usepackage{caption}
\usepackage{subcaption}
\usepackage{multirow}
\usepackage{dblfloatfix}
\usepackage{threeparttable}
\usepackage{xcolor}
\newcommand{\etal}{\textit{et al.}}
\ifCLASSOPTIONcompsoc
  \usepackage[nocompress]{cite}
\else
  \usepackage{cite}
\fi
\ifCLASSINFOpdf
\else
\fi
\begin{document}
%
% paper title
% Titles are generally capitalized except for words such as a, an, and, as,
% at, but, by, for, in, nor, of, on, or, the, to and up, which are usually
% not capitalized unless they are the first or last word of the title.
% Linebreaks \\ can be used within to get better formatting as desired.
% Do not put math or special symbols in the title.
\title{Deep Insights of Learning based Micro Expression Recognition: A Perspective on Promises, Challenges and Research Needs}

\author{Monu~Verma,
        Santosh~Kumar~Vipparthi,
        and~Girdhari~Singh,% <-this % stops a space
\IEEEcompsocitemizethanks{\IEEEcompsocthanksitem Monu Verma is with the Electrical and Computer Engineering, University of Miami, FL, USA, 33155.
 Santosh Kumar Vipparthi is with the CVPR lab of the Indian Institute of Technology Ropar Department
of Electrical Engineering, Roopnagar, India, 140001. Girdhari Singh is with the Department
of Computer Science and Engineering, Malaviya National Institute of Technology, Jaipur,
INDIA, 302017.~E-mail: monuverma.cv@gmail.com, skvipparthi@iitrpr.ac.in and gsingh.cse@mnit.ac.in
% note need leading \protect in front of \\ to get a newline within \thanks as
% \\ is fragile and will error, could use \hfil\break instead.

}% <-this % stops an unwanted space
\thanks{}}

\markboth{Journal of \LaTeX\ Class Files,~Vol.~XX, No.~XX, Month~XXXX}%
{}

\IEEEtitleabstractindextext{%
\begin{abstract}
Micro expression recognition (MER) is a very challenging area of research due to its intrinsic nature and fine-grained changes. In the literature, the problem of MER has been solved through handcrafted/descriptor-based techniques. However, in recent times,  deep learning (DL) based techniques have been adopted to gain higher performance for MER. Also, rich survey articles on MER are available by summarizing the datasets, experimental settings, conventional and deep learning methods. In contrast, these studies lack the ability to convey the impact of network design paradigms and experimental setting strategies for DL based MER. Therefore, this paper aims to provide a deep insight into the DL-based MER frameworks with a perspective on promises in network model designing, experimental strategies, challenges, and research needs. Also, the detailed categorization of available MER frameworks is presented in various aspects of model design and technical characteristics. Moreover, an empirical analysis of the experimental and validation protocols adopted by MER methods is presented. The challenges mentioned earlier and network design strategies may assist the affective computing research community in forge ahead in MER research. Finally, we point out the future directions, research needs and draw our conclusions.
\end{abstract}

% Note that keywords are not normally used for peerreview papers.
\begin{IEEEkeywords}
Micro expression recognition, Facial expression recognition, CNN models, deep learning.
\end{IEEEkeywords}}

% make the title area
\maketitle

% To allow for easy dual compilation without having to reenter the
% abstract/keywords data, the \IEEEtitleabstractindextext text will
% not be used in maketitle, but will appear (i.e., to be "transported")
% here as \IEEEdisplaynontitleabstractindextext when the compsoc 
% or transmag modes are not selected <OR> if conference mode is selected 
% - because all conference papers position the abstract like regular
% papers do.
\IEEEdisplaynontitleabstractindextext
% \IEEEdisplaynontitleabstractindextext has no effect when using
% compsoc or transmag under a non-conference mode.

% For peer review papers, you can put extra information on the cover
% page as needed:
% \ifCLASSOPTIONpeerreview
% \begin{center} \bfseries EDICS Category: 3-BBND \end{center}
% \fi
%
% For peerreview papers, this IEEEtran command inserts a page break and
% creates the second title. It will be ignored for other modes.
\IEEEpeerreviewmaketitle

\section{Introduction}\label{sec:introduction}

% Here we have the typical use of a "T" for an initial drop letter
% and "HIS" in caps to complete the first word.
\IEEEPARstart{M}{icro}-expressions (MEs) exhibit the insight into true feelings of a person even if he/she is trying to hide the genuine emotions within the manifested emotion (macro expression). 
Ekman \cite{ekman1969nonverbal} introduced various deceptive expressions (known as MEs) after investigating a depressed patient's interview video who attempted to commit suicide. Ekman observed that, the patient surpasses his intensive sadness in happiness within 1/12 seconds. However, these expressions were spotted in a few frames of the video, recorded through a standard 25 fps device but serves enough clues to sense the true sentiments of the patient. Thus, MEs can be decisive in the fields of spotting genuine psychological activities \cite{takalkar2018survey}, \textit{e.g.}, lie detection, psychoanalysis, criminal interrogation, medical diagnosis, pain detection, autism disorder, and business negotiation. \par
{MER can broadly be divided into three steps: pre-processing, feature extraction and emotion classification.\\
}
\textbf{\textit{Preprocessing Techniques:}}
  The first step of MER is to spot the MEs and then detect the RoIs from them {ME spotting is a vital step for automatic ME analysis as it locates the segments of micro-movements in a MEs video. Thus, precise ME spotting can decrease the redundant information and improve the performance of MER further. Recently, few studies \cite{zhang2018smeconvnet, liong2015automatic, tran2021micro} focused on ME spotting using deep learning methods. Li \etal \cite{li2017towards} introduced an ME spotting method for spontaneous ME datasets. Furthermore, Zhang  \etal \cite{zhang2018smeconvnet} designed a deep learning-based ME spotting method by extracting features from video clips. Tran  \etal \cite{tran2021micro}  proposed a deep sequence model for ME spotting.   Moreover, Liong  \etal  \cite{liong2015automatic}  proposed an automatic apex frame spotting model. (A more
detailed categorization and analysis of the ME spotting can be found in existing MER surveys 
\cite{oh2018survey, li2017towards, xie2020overview, ben2021video})}. Further, face alignment and noise filtration are employed to systematize the input data samples for better feature extraction and learning \cite{xie2020overview}. Some of the MER frameworks also utilized the motion magnification (MM) \cite{xia2020revealing, li2020joint} and temporal normalization \cite{zhou2019dual, xia2019spatiotemporal} techniques to enhance the visibility of the minute temporal variations and normalize the frames. {Recently, deep learning (DL) based approaches require huge dataset for training. However, all available MEs datasets are far from the enough data samples. Therefore, data augmentation techniques such as: random crop and rotation in terms of the spatial domain, shifting, magnification \cite{xia2019spatiotemporal} and synthetic data generation using generative adversarial networks (GANs) \cite{zhang2018joint, yu2020ice} are also gaining the attention to enhance the data samples.} The more details can be seen in \cite{xie2020overview, goh2020micro}. \par
{Based on feature extraction and classification methods, MER approaches can be categorized into traditional and deep learning based approaches.\\ 
\textbf{\textit{Traditional handcrafted MER methods:}}
The traditional MER methods rely on the predesigned feature descriptors to encode the spatial and temporal changes from the MEs video sequences. In literature many robust spatio-temporal feature descriptors: local binary pattern (LBP) and its variants: three orthogonal planes (LBP-TOP) \cite{zhao2007dynamic}, LBP with six intersection points (LBP-SIP) \cite{wang2014lbp}, spatiotemporal LBP with integral projection (STLBP-IP) \cite{huang2015facial}, revisited integral projection (DiSTLBP-RIP) \cite{huang2017discriminative} \textit{etc.} were proposed for MER. Furthermore, some optical flow based descriptors: main directional mean optical flow (MDMO) \cite{liu2015main}, sparseMDMO \cite{liu2018sparse}, FHOFO \cite{happy2017fuzzy}, facial dynamic map (FDM) and color based descriptors: TICS \cite{wang2015micro} were introduced to encode the features of MEs. After that, the encoded features are forwarded to the classifiers such as support vector machine (SVM), neural networks (NN), \textit{\textit{etc.}}, which learn the distinctive properties of the emotion classes. Sensitivity and specificity of the traditional descriptors have gained good performance as compared to professionally trained specialists. However, it is still difficult to manually design a robust descriptor
for capturing quick subtle changes in MEs. The detailed summary of the traditional MER approaches are listed in the supplementary draft (supplementary: Table III), A more detailed categorization of traditional methods and classifiers can be found in \cite{li2017towards,ben2021video}.}\\

\textbf{\textit{Deep Learning based MER methods:}} The supervised techniques of deep learning adaptively learn the features from the raw data and classify the emotion classes accordingly. This paper aims to describe the details of network design strategies followed in the literature in-terms of downsampling, multi-stream, multi-scale, deep or shallow networks, kernels depth, sizes, \textit{etc.}, for MER. Moreover, there is no standard evaluation protocol, class settings and metrics for the fair comparison of the models were not present in the literature. Therefore, it is difficult to come up with a common conclusion of the performances for the existing state-of-the-art approaches. Thus, these factors motivated us to present a detailed survey by addressing the effects of selecting the input formats, evaluation strategies, implementation settings, evaluation metrics on MER performance. More details of input formats such as apex frame, onset-apex-offset frames, compressed single instance image, and image sequences are studied and discussed their effect on the model's overall performance. Similarly, the evaluation strategies like person dependent, a person independent, composite, cross-domain, and class settings like 3-, 4-, 5- and 7- emotion classes, and implementation setting like learning rates, data augmentation, evaluation metrics: recognition accuracy, F1-score, unweighted F1-score, unweighted average recall, mean diagonal value of the confusion matrix and its impact on overall models performances is discussed in detailed.
\footnote[1]{{In this paper, terms learning and deep learning are often used interchangeably.}}

\begin{table*}[!t]
\centering
\caption{Summarization of existing surveys in last decade.}
\label{tab:1}
\begin{tabular}{lccc}
\hline \noalign{\smallskip}
\textit{\textbf{No.}} & \textit{\textbf{Pub-Year}} & \textit{\textbf{Title}}                                                             & \textit{\textbf{Highlights}}                                                                                                                                                             \\ \noalign{\smallskip}\hline \hline

1.                    &  \begin{tabular}[c]{@{}c@{}}Fr. Psy. 2018 \cite{oh2018survey} \\(Frontiers) \end{tabular}           & \begin{tabular}[c]{@{}c@{}}A Survey of Automatic Facial Micro-\\Expression   Analysis: Databases, Methods,\\ and  Challenges\end{tabular}                            & \begin{tabular}[c]{@{}c@{}}A survey of databases and traditional feature based methods \\ for MEs spotting and recognition. The survey also presents \\ the evaluation metrics (accuracy and f1 score), validation \\ strategies and challenges.\end{tabular}                                                                                    \\\hline
2.                     &\begin{tabular}[c]{@{}c@{}} T-AFF-2018 \cite{li2017towards}\\ (IEEE) \end{tabular}               & \begin{tabular}[c]{@{}c@{}}Towards Reading Hidden Emotions: A \\Comparative Study  of Spontaneous Micro-\\Expression Spotting and Recognition\\ Methods\end{tabular} & \begin{tabular}[c]{@{}c@{}}A survey of various preprocessing methods and feature \\ descriptors for spontaneous ME spotting and recognition. \\ The survey also conducted the competitive analysis of \\ conventional MER approaches.\end{tabular}                                                                                                \\\hline
3.                     & Arxiv-2018 \cite{merghani2018review}                & \begin{tabular}[c]{@{}c@{}}A Review on Facial Micro-Expressions \\Analysis: Datasets,  Features and Metrics\end{tabular}                                         & \begin{tabular}[c]{@{}c@{}}A survey of the various feature extraction algorithms,  classifiers \\ and evaluation metrics for MER. The survey  also  presents\\ the datasets , challenges in data accumulation, comparative\\ analysis between available datasets, evaluation  metrics,\\ validation strategies and analysis of learning  approaches as \\compared to traditional approaches.\end{tabular}                                                                                                    \\\hline
4.                     & \begin{tabular}[c]{@{}c@{}}Vis.Comp-2020 \cite{goh2020micro} \\(Springer) \end{tabular}           & \begin{tabular}[c]{@{}c@{}}Micro-expression recognition: an updated \\review of current  trends, challenges and\\ solutions\end{tabular}                           & \begin{tabular}[c]{@{}c@{}}A survey of traditional approaches for MEs detection and \\ recognition by categorizing solutions into low-level, \\ mid-level, and high-level solutions.\end{tabular}                                                                                                                                                 \\\hline
5.                    & \begin{tabular}[c]{@{}c@{}}IVC-2020 \cite{zhou2021survey}   \\(Elsevier) \end{tabular}               & A survey of micro-expression recognition                                                                                                                         & \begin{tabular}[c]{@{}c@{}}A survey of datasets, preprocessing techniques and existing\\  MER algorithms based on the issues: overfitting, data \\ unbalancing and robustness.\end{tabular}                                                                     \\\hline
6.                     & Arxiv-2020 \cite{xie2020overview}                & \begin{tabular}[c]{@{}c@{}}An Overview of Facial Micro-Expression \\Analysis: Data,  Methodology and Challenge\end{tabular}                                      & \begin{tabular}[c]{@{}c@{}}A survey of datasets, preprocessing techniques, handcrafted\\ features, ME spotting and learning based MER. The \\learning based MER discussed by dividing into three \\ aspects:: macro to- micro adaptation, recognition \\based on key apex frames, and recognition \\ based on facial action units. \end{tabular}                                                                                                                                   \\\hline
7.                     &\begin{tabular}[c]{@{}c@{}} T-PAMI-2021 \cite{ben2021video}\\ (IEEE) \end{tabular}                & \begin{tabular}[c]{@{}c@{}}Video-based Facial Micro-Expression\\ Analysis: A Survey of  Datasets, Features \\and Algorithms\end{tabular}                           & \begin{tabular}[c]{@{}c@{}}A review that highlighted the key differences between \\ macro- and micro-expressions, video-based  micro-expression \\analysis, neuropsychological basis for MEs and datasets.\\ The survey focused on the comprehensive study of existing \\ MEs dataset and  introduce a new MMEW dataset. Also,\\ the survey included a brief summary of features spotting \\ algorithms, recognition algorithms, applications and \\evaluation metrics for  MER  approaches.\end{tabular} \\\hline
8.                     & Arxiv-2021 \cite{li2021deep}             & \begin{tabular}[c]{@{}c@{}}Deep Learning for Micro-expression \\ Analysis: A Survey\end{tabular}                           & \begin{tabular}[c]{@{}c@{}}A review that highlighted deep learning based MER \\ methods, challenging datasets, and comparative analysis \\between most infuential DL based MER methods.\\ The survey also included remaining challenges,\\ and future direction of MER.\end{tabular} \\\hline
9.                     & Sensors-2022 \cite{guerdelli2022macro}             & \begin{tabular}[c]{@{}c@{}}Macro-and Micro-Expressions Facial \\ Datasets: A Survey \end{tabular}                           & \begin{tabular}[c]{@{}c@{}}A survey of facial expression datasets, including \\  both macro and micro expressions.\end{tabular} \\\hline
10.                    & The proposed               & \begin{tabular}[c]{@{}c@{}}Deep Insights of Learning based MER\\ Frameworks:  A Perspective on Promises, \\Challenges and Research Needs\end{tabular}              & {\begin{tabular}[c]{@{}c@{}} {The comparative review of the existing deep learning} \\ {frameworks with research needs and the deep study} \\{of the technical  characteristics like, end-to-end vs two-stage }\\ {architecture, downsampling, multi-stream/scale structure, }\\{deeper vs shallow network, etc. has done.  Moreover, the MER}\\ {approaches used  different metric calculation or platforms}\\{ or number of emotion classes or input strategies;}\\ {therefore it is difficult to compare the performance}\\ {of MER frameworks. We reviews these factors in detail} \\{and highlighted the effects on the performance of MER.} \end{tabular}}\\ \hline                                    
\end{tabular}
\end{table*}
\subsection{Comparison with previous reviews}
In past years, many notable surveys related to MER approaches have been published and the details are summarized in  Table  \ref{tab:1}. Many articles focus to summarise the details of datasets, preprocessing  techniques,  traditional  MEs  spotting  along with  the  feature  extraction  algorithms,  classifiers  and  experimental  settings.  Also, the survey of the MER framework is available in \cite{merghani2018review, goh2020micro, zhou2021survey, xie2020overview, ben2021video}.
Firstly, Merghani \etal \cite{merghani2018review} briefly described CNN frameworks and the technical differences with the traditional approaches in MER. Further, a  detailed study of MEs databases and the comparative analysis of the data with challenges in data accumulation and labeling, evaluation metrics, and strategies are discussed. Similarly, Goh \etal  \cite{goh2020micro} presented a survey by including datasets,  preprocessing techniques,  MEs spotting, and feature extraction algorithms. The main focus of the study is to highlight the ME features by dividing them into three categories: low, mid, and high-level features. Also, Zhou et al \cite{zhou2021survey} presented a brief survey of the available traditional and deep learning techniques with preprocessing techniques and datasets. Apart from the similar data collection, evaluation matrix, and categorization of the conventional and CNN methods, Xie et al \cite{xie2020overview} present the details of the macro-to-micro feature adoption and synthetic data generation to balance MEs data samples. {Guerdelli \etal \cite{guerdelli2022macro} prsented a detailed survey of facial expression datasets by including both macro and micro expressions.} In recent times, Ben et ac. \cite{ben2021video} presented a “micro-and-macro expression warehouse (MMEW)” dataset by incorporating both micro and macro expressions. Also, a detailed study of the challenges while creating the datasets is presented.  The apparent technical differences between the conventional approaches in the literature are presented for ME.  In addition, a brief study of the CNN techniques is given in this article. From the above details, it is clear that the available survey articles focus on presenting a categorization of the available models and datasets. {Similarly, Li \etal \cite{li2021deep} presented a detailed study of deep learning methods for MER, challenging datasets, and comparative analysis between  most influential MER methods. Moreover, the study also detailed the remaining challenges and future scope of the MER.}  However, they fail to comprehensively analyze the essential aspects like model designing, evaluation strategies, challenges, and research needs for deep learning models. Therefore, this article focuses on presenting a detailed survey on:

\begin{enumerate}
\item The detailed survey of DL-based MER approaches has been presented by categorizing into three major categories:  multi-stage,  end-to-end, and transfer learning-based MER. Further, subcategorization of these categories is offered by considering the CNN frameworks like 2D-CNN, multi-stream/scale, capsule, 3D-CNN, CNN-LSTM, Graph-based, NAS based, etc. Also, the detailed technical characteristics of these frameworks are discussed.
\item Promises towards the different modules of DL based model designing such as two stage and end-to-end learning, downsampling, kernel sizes, shallow and deep networks, to design and develop an effective and efficient deep learning framework for MER are discussed and observations are concluded.

\item A detailed discussion on validation strategies (PDE:  LOVO,  80/20split,  PIE:  LOSO,  composite  LOSO,  CDE), evaluation metrics (accuracy,  F1-score,  Recall,  UF1,  UAR,  confusion  matrix) and their significance on MER frameworks are presented. 

\item A  detailed  study  of different validation  setups:  PDE,  PIE,  and  CDE,  and  other experimental settings: data augmentation, input selection,  number  of  emotion  classes,  etc., is conducted to analyze their effects  over  the performance  of  the  MER  approaches.  Thus,  new researchers  will  get  awareness  about  the  selection of experimental settings in DL-based MER.

\end{enumerate}
To the best of our knowledge this is the first attempt to comparatively  analyze  the  role  of  various  designing  modules, evaluation strategies, and experimental settings for learning based  MER  frameworks.  The  main  aim of  this  study is to help researchers or affective computing community to concentrate on the effective designing modules and experimental settings to design a robust MER framework. The detailed comparison between recently published survey \cite{ben2021video} and proposed survey is presented in the supplementary file. In addition, the supplementary file also included the visual presentation of the different DL-based MER approaches. The detailed information about the handcrafted MER approaches is tabulated in  Table III. of the supplementary file.

\begin{figure*}
    \centering
    \includegraphics[width=0.92\linewidth, height=3in]{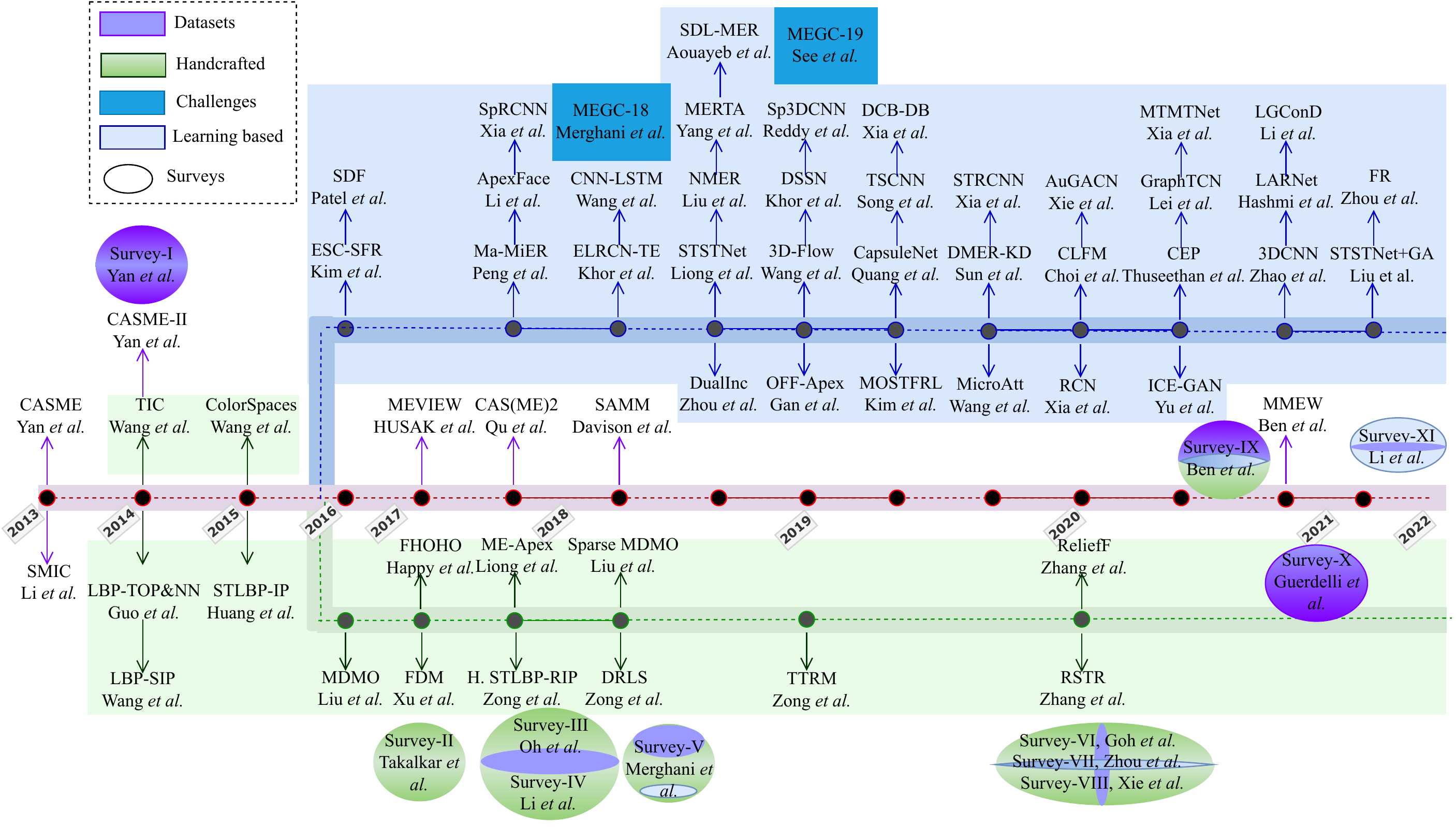}
    \caption{Milestones of MER, including datasets , traditional , recent deep learning methods, challenges and survey paper.  Most listed papers are highly cited and published in top journals or conferences. The main focus of this study is highlighted in blue color.}
    \label{fig:1}
\end{figure*}
\section{Deep Learning based Micro Expression Recognition}\label{categories}
\begin{figure}[!t]
    \centering
    \includegraphics[width=0.85\linewidth, height=2.5in]{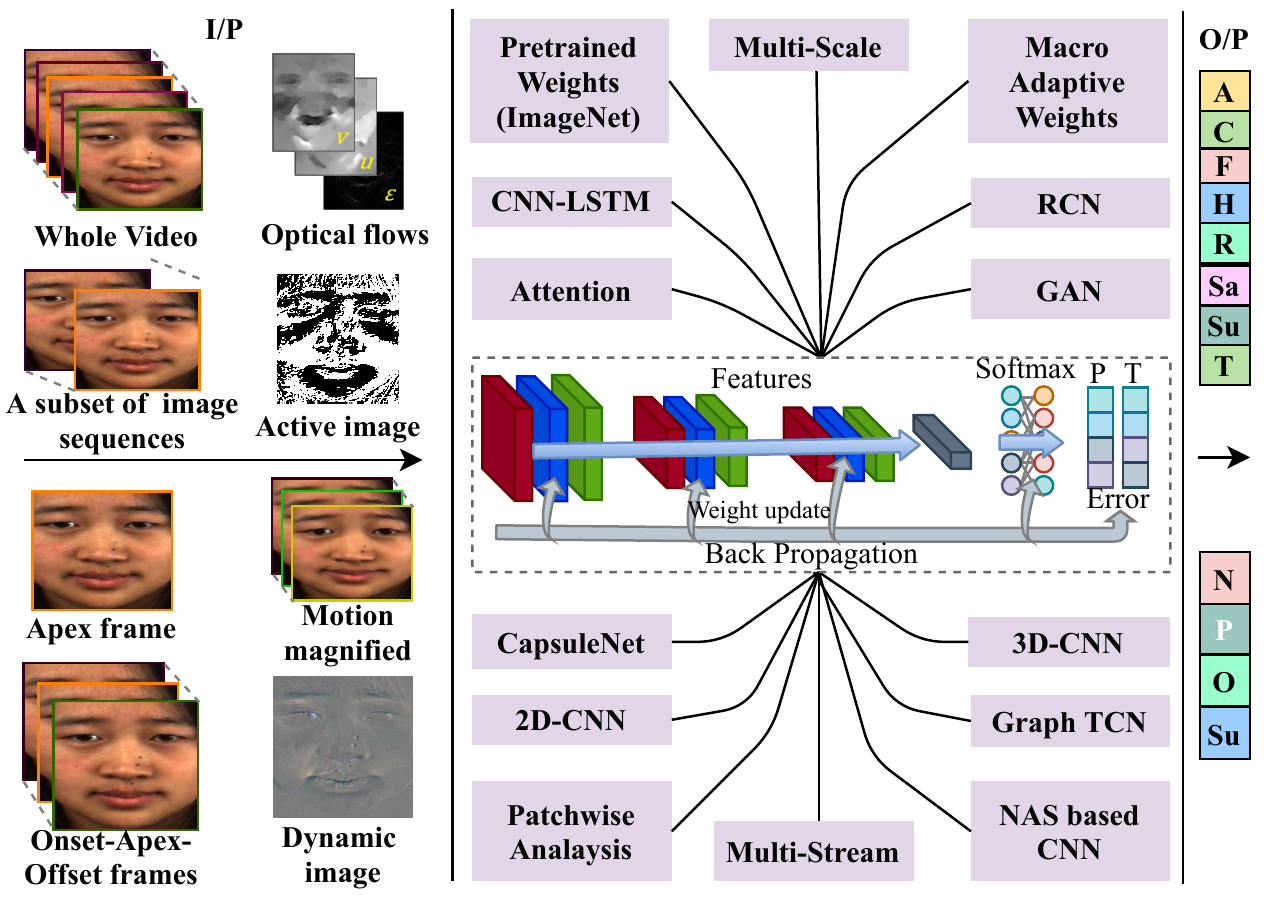}
    \caption{A flow diagram of deep learning frameworks for generic MER. \textit{Here, I/P and O/P implies for the input and output.}}
    \label{fig:2}
\end{figure}

  The   MER   frameworks   demand   spatio-temporal   feature learning  with  momentary  changes  to  capture  the  subtle variations of MEs. These factors make the design and development of deep learning models for MER an incredibly challenging task.  In  this  section,  we  present  an  empirical  review  of deep  learning  methods  highlighted  in  Fig. \ref{fig:1}.  Based  on available framework  architectures in the literature, we categorize these frameworks into  three  broad  categories:  multi-stage  (Section  \ref{multistage}), an end-to-end (Section \ref{endtoend}), and transfer learning (Section \ref{transferlearning}) based  MER  frameworks  as  shown  in  Table  \ref{tab:3}.  We  further divide the categories into subcategories based on different network characteristics: 2D-CNN, multi-scale/stream, capsule, recurrent convolutional, 3D-CNN, CNN-LSTM, graph based, generative adversarial networks (GANs) and neural architecture search (NAS) based networks as shown in Fig. \ref{fig:2}.  Further,  a  deep  analysis  of  DL  based  MER  frameworks has been done and observations are discussed in Section \ref{techChar}.

\subsection{Multi-Stage Frameworks}\label{multistage}
{ In MER, most of the multi-stage frameworks are two-stage, where, in the first stage, the handcrafted descriptors are employed to extract the primary features.  In the second stage, the CNN network is used to learn the expressive features of the MER.}  Based  on  characteristics  of  these  deep  learning  MER models, two-stage frameworks are further classified into sub-categories: 2D-CNN, multi-stream/scale, capsule, RCN and 3D-CNN. More details of the sub categorization is discussed as follows. 
\subsubsection{2D-CNN Models}
Almost all deep learning based MER approaches adopt the 2D-CNN  networks  \cite{verma2019learnet, vermaAff, verma2020non, li2020joint, khor2019dual}.  Learning  the spatio-temporal  features  from  2D-CNN  is  an  insignificant problem. Therefore, to maintain the compatibility with 2D-CNN,  researchers  have  designed  two-stage  frameworks \cite{verma2019learnet, song2019recognizing, gan2019off, vermaAff, verma2020non, li2020joint, khor2019dual}.  Where,  in  the  first stage  optical  flow  or  single  instance  image  is  computed by  applying  handcrafted  approaches.  While  in  the  second stage the 2D-CNN model is designed to learn the MEs specific features.  Specifically,  Verma \etal{et  al.}  \cite{verma2019learnet}  introduced  a  2D-CNN MER framework. The framework first computed the single  image  instances  by  applying  the  dynamic  imaging. Furthermore, the 2D-CNN model LEARNet is designed to learn the MEs specific features. {Gupta \cite{gupta2021merastc}  introduced the 2D-CNN MERASTC approach for MER by encoding the subtle deformations through action units (AUs), landmarks, gaze, and appearance features of MER.}\\
\textbf{\textit{Discussion:}} 
For MER, 2D-CNN models require an auxiliary first stage to process the spatio-temporal information into 2-dimensional formats (Supplementary Fig. 2). However, the  2D-CNN  are  easy  to  design  as well as require very less computational cost as compared to spatio-temporal networks like CNN-LSTM, RCN, 3D-CNN \textit{etc.} In addition 2D-CNN approaches \cite{song2019recognizing, gan2019off, gupta2021merastc} have shown impressive performance as shown in Table \ref{tab:5}-\ref{tab:6}. Therefore, 2D-CNN networks gain much attention as compared to other techniques in MER as reported in Table \ref{tab:3}. 
\subsubsection{Multi-Stream Networks}
Multi-stream networks capture a diverse range of features from  different  streams  to  learn  salient  edge  variations  of the MEs. In the literature, many two-stage MER frameworks \cite{khor2018enriched, liu2020offset, yang2019merta, song2019recognizing} exploit the hybrid feature learning capability of the multi-stream networks. Li et at. \cite{li2020joint} designed two streams (local and global path)  based  networks  to  use  the  coupled features  of  local  (sub-regions)  and  global  (whole  face),  respectively. Song \etal \cite{song2019recognizing} proposed a three streams-based CNN network consisting of a static-spatial stream, local-spatial  stream,  and  dynamic-temporal  stream  to  capture three different clues in three different frames. Furthermore, spatial features are concatenated and fed to the single LSTM to  learn  the  temporal  features.  Khor  \etal  \cite{khor2018enriched}  introduced a CNN model with two streams, channel wise stacking for spatial  enrichment  and  feature  wise  stacking  for  temporal enrichment. Liong \etal \cite{ststnet} designed a shallow three stream network  to  learn  the  optical  flows  guided  features.  Similarly, to use the optical flow guided features, Ganet \etal \cite{liu2020offset} introduced two stream networks for MER. Yang \etal \cite{yang2019merta} exploit the feature discriminative capability of the VGGNet-16  to  capture  the  spatial  feature  of  MEs  in  three  streams: ME  sequences,  optical  flow,  and  optical  strain.  Liu  et  al. \cite{liu2020offset}  introduced  a  five-stream  network  with  capsuleNet  to improve the performance of MER. Also, a detailed comparison of the performances of multi-stream networks is tabulated in Table \ref{tab:5}-\ref{tab:6}.\\
\textbf{\textit{Discussion:}} In literature Multi-Stream networks achieve high performance  in  MER over sequential 2D-CNN networks.  The  multi-stream  networks  can  capture  the  diverse  range  of  features  from  different streams  and  boost  the  efficiency  of  the  network.  Moreover, the  multi-stream  networks  benefited  MER  frameworks  to learn enough features with shallow networks and small data samples.  From Table \ref{tab:5}-\ref{tab:6} it is evident that multi-stream networks outperform the sequential networks in terms of performance. 
\begin{table*}[!htp]
\centering
\caption{The comparison of the CNN based MER approaches based on network design. }
\label{tab:3}
 \begin{threeparttable}
\begin{tabular}{lcccccccc}
\hline
\textbf{Pub-Yr}      & \textbf{Input}                                                               & \textbf{Prepr.}                                                          & \textbf{\begin{tabular}[c]{@{}c@{}}Handcrafted\\ Support\end{tabular}}                                                    & \textbf{N/W Type}                                                               & \textbf{E-to-E} & \textbf{\begin{tabular}[c]{@{}c@{}}Macro \\Support\end{tabular}}                                                 & \textbf{\begin{tabular}[c]{@{}c@{}}P.T. \\Weights\end{tabular}} & \textbf{Protocol}                                                         \\ \hline \hline
ACM-2016 \cite{kim2016micro}   & Video                                                                        & No                                                                       & No                                                                              & CNN-LSTM                                                                        & Yes                 & No                                                                    & No                           & LOSO                                                                                       \\
ICPR-16 \cite{patel2016selective}    & Video (20F)                                                                  & ASM                                                                      & No                                                                              & 2D-CNN                                                                          & Yes                 & Yes                                                        & No                           & 4Fold-LOSO                                                                                \\
FG-18 \cite{peng2018macro}       & Apex                                                                         & C.S, R, S& No                                                                              & Transfer-Learning                                                               & Yes                 & Yes & Yes                          & \begin{tabular}[c]{@{}c@{}}SD-LOSO, \\ CDE\end{tabular}                                   \\
ICIP-18 \cite{li2018can}     & Apex                                                                         & EVM                                                                      & No                                                                              & Transfer-Learning                                                               & Yes                 & No                                                                    & Yes                          & SD-LOSO                                                                                    \\
IPTA-18 \cite{xia2018spontaneous}    & Video (30F)                                                                  & ASM, EVM                                                                 & No                                                                              & RCN                                                                             & Yes                 & No                                                                    & No                           & 5Fold-LOSO                          \\
FG-18 \cite{khor2018enriched}      & Video (10F)                                                                  & DLib, TIM                                                                & Yes: O. F.                                                                      & \begin{tabular}[c]{@{}c@{}}CNN-LSTM, \\ Multi-Stream\end{tabular}               & No                  & No                                                                    & Yes                          & \begin{tabular}[c]{@{}c@{}}SD-LOSO,\\    \\ CD-LOSO, \\ CDE\end{tabular}                 \\
FG-19 \cite{zhou2019dual}       & Onset, Apex                                                                  & {TV-L1}                                                               & Yes: O. F.                                                                      & \begin{tabular}[c]{@{}c@{}}2D-CNN, \\ Multi-Stream, \\ Multi-scale\end{tabular} & No                  & No                                                                    & No                           & \begin{tabular}[c]{@{}c@{}}SD-LOSO, \\ CD-LOSO\end{tabular}                                \\
FG-19 \cite{ststnet}       &  {Onset, Apex}     & {-}   & Yes: O. F. & 3D-CNN  & No   & No  & No                          & \begin{tabular}[c]{@{}l@{}}SD-LOSO, \\ CD-LOSO\end{tabular}                                \\
FG-19 \cite{liu2019neural}       & {Onset-Apex}                                                                        & -                                                                & Yes: O. F.                                                                      &  Multi-Stream                & No                  & No                                                                    & No                           & \begin{tabular}[c]{@{}l@{}}SD-LOSO, \\ CD-LOSO\end{tabular}                                \\
MTA-19 \cite{yang2019merta}     & Video (10F)                                                                  & TIM                                                                      & Yes: O. F.                                                                      & \begin{tabular}[c]{@{}c@{}}CNN-LSTM, \\ Multi-Stream\end{tabular}               & No                  & No                                                                    & Yes                          & SD-LOSO                                                                                                                                        \\
PAA-19 \cite{Li3D-Flow}     & Video (10-15F)                                                               & TIM                                                                      & Yes: O.F.                                                                       & 3D-CNN                                                                          & No                  & No                                                                    & No                           & SD-LOSO                                                                                   \\
ICIP-19 \cite{khor2019dual}     & Onset-Apex                                                                  & TV-L1,                                                                   & Yes: O.F                                                                        & \begin{tabular}[c]{@{}c@{}}2D-CNN, \\ Multi-Stream\end{tabular}                 & No                  & No                                                                    & No                           & SD-LOSO                                                                                   \\
IJCNN-19 \cite{reddy2019spontaneous}    & \begin{tabular}[c]{@{}c@{}}Video \\    \\ (96F)\end{tabular}                 & -                                                                        & No                                                                              & \begin{tabular}[c]{@{}c@{}}3D-CNN, \\ Multi-Stream\end{tabular}                 & Yes                 & No                                                                    & No                           & 80/20 Split                                                                               \\
FG-19 \cite{van2019capsulenet}      & Apex                                                                         & FacialToolkit                                                            & No                                                                              & Capsule                                                                         & Yes                 & No                                                                    & Yes                          & \begin{tabular}[c]{@{}c@{}}SD-LOSO, \\ CD-LOSO\end{tabular}                                \\
IEEE-Acc-19 \cite{song2019recognizing} & Video                                                                        & -                                                                        & \begin{tabular}[c]{@{}c@{}}Yes: UPLBP,\\ O. F.\end{tabular}                                                               & \begin{tabular}[c]{@{}c@{}}2D-CNN, \\ Multi-Stream, \\ Multi-Scale\end{tabular} & No                  & Yes                                                          & Yes                          & SD-LOSO                                                                                  \\
ICBEA-19 \cite{xia2019cross}    & {Onset-Apex}                                                                        & EVM                                                                      & Yes: O. F.                                                                      & RCN                                                                             & No                  & No                                                                    & No                           & \begin{tabular}[c]{@{}c@{}}SD-LOSO, \\ CD-LOSO\end{tabular}                               \\
Neu.Co-19 \cite{wang2020micro}   & Apex                                                                         & AAM                                                                      & No                                                                              & \begin{tabular}[c]{@{}c@{}}2D-CNN, \\ Multi-Scale\end{tabular}                  & Yes                 & Yes & Yes                          & \begin{tabular}[c]{@{}c@{}}SD-LOSO, \\ CD-LOSO, \\ CDE\end{tabular}                       \\
TIP-19 \cite{verma2019learnet}      & Video                                                                        & Voila-Jones                                                              & Yes: DI                                                                         & \begin{tabular}[c]{@{}c@{}}2D-CNN, \\ Multi-Scale\end{tabular}                  & No                  & No                                                                    & No                           & 80/20 Split                                                                              \\
MLSP-19 \cite{aouayeb2019spatiotemporal}     & -                                                                            & DLib                                                                     & No                                                                              & CNN-LSTM                                                                        & Yes                 & No                                                                    & No                           & \begin{tabular}[c]{@{}l@{}}SD-LOSO, \\ CD-LOSO\end{tabular}                                \\
IJCNN-20 \cite{verma2020non}      & Video                                                                        & Voila-Jones                                                              & Yes: AI                                                                         & \begin{tabular}[c]{@{}c@{}}2D-CNN, \\ Multi-Scale\end{tabular}                  & No                  & No                                                                    & No                           & SD-LOSO                                                                                  \\
ToM-20 \cite{xia2019spatiotemporal}     & Video                                                                        & \begin{tabular}[c]{@{}c@{}}ASM, TIM,\\ EVM \end{tabular}                                                            & Yes: O. F                                                                       & RCN                                                                             & No                  & No                                                                    & No                           & \begin{tabular}[c]{@{}l@{}}SD-LOSO, \\ LOVO\end{tabular}                             \\
TIP-20 \cite{xia2020revealing}      & {Onset-Apex}                                                                        & EVM                                                                      & Yes: O. F                                                                       & RCN                                                                             & No                  & No                                                                    & No                           & \begin{tabular}[c]{@{}c@{}}SD-LOSO, \\ CD-LOSO\end{tabular}                            \\
TAFF-20 \cite{sun2020dynamic}     & \begin{tabular}[c]{@{}c@{}}Video, \\ Apex, \\ On-A-off\end{tabular} & OpenFace                                                                 & No                                                                              & Teacher-Student                                                                 & Yes                 & Yes                                                           & Yes                          & SD-LOSO                                                                                   \\
ACMMM-20 \cite{xie2020assisted}   & Video                                                                        & -                                                                        & No                                                                              & GAN, 3D-CNN                                                                     & Yes                  & Yes                                                                   & No                           & \begin{tabular}[c]{@{}c@{}}SD-LOSO, \\ CDE,    \\ LOVO\end{tabular}               \\
ACMMM-20 \cite{xia2020learning}   & Apex                                                                         & -                                                                        & No                                                                             & \begin{tabular}[c]{@{}c@{}}2D-CNN, \\ Encoder-Decoder\end{tabular}              & Yes                  & Yes        & Yes                          & \begin{tabular}[c]{@{}c@{}}SD-LOSO, \\ CD-LOSO\end{tabular}                               \\
ACMMM-20 \cite{lei2020novel}   & {Onset-Apex}                                                                         & -                                                                        & No                                                                              & Graph-TCN                                                                       &Yes                  & No                                                                    & No                           & SD-LOSO                                                                                  \\
Arxiv-20 \cite{yu2020ice}   & On-A-Off                                                            &\begin{tabular}[c]{@{}c@{}}Landmark\\ Detection \end{tabular}                                                        & No                                                                              & GAN, Capsule                                                                    & Yes                  & No                                                                    & Yes                          & \begin{tabular}[c]{@{}c@{}}SD-LOSO, \\ CD-LOSO\end{tabular}                               \\
IEEE-Acc-20 \cite{choi2020facial} & Video                                                                        & -                                                                        & Yes: LMF                                                                        & CNN-LSTM                                                                        & No                  & Yes                                                             & Yes                          & \begin{tabular}[c]{@{}c@{}}SD-LOSO, \\ CD-LOSO, \\ CDE\end{tabular}                        \\
TIP-21 \cite{li2020joint}      & Apex                                                                         & EVM                                                                      & Yes: 3D-FFT                                                                     & 2D-CNN                                                                          & No                  & No                                                                    & Yes                          & SD-LOSO                                                                                   \\
IEEE-Acc-21 \cite{thuseethan2020complex} & Image                                                                        & AAM, CLAHE                                                               & Yes                                                                             & 2D-CNN                                                                          & No                  & No                                                                    & No                           & \begin{tabular}[c]{@{}c@{}}SD-LOSO, \\ 90/10 Split\end{tabular}                     \\

IEEE-MM-21 \cite{vermaAff}    & Video                                                                        & Viola Jones                                                              & Yes: DI                                                                         & \begin{tabular}[c]{@{}c@{}}2D-CNN, \\ Multi-Scale\end{tabular}                  & No                  & No                                                                    & No                           & \begin{tabular}[c]{@{}c@{}}SD-LOSO, \\ CDE\end{tabular}                             \\
TAFF-21 \cite{gupta2021merastc}      & Video                                                                        & Viola Jones                                                              & \begin{tabular}[c]{@{}c@{}}Yes: Landmark\\ $\&$,  Gaze feature\end{tabular} & 2D-CNN                                                                          & No                  & No                                                                    & No                           & \begin{tabular}[c]{@{}c@{}}SD-LOSO, \\ CD-LOSO\end{tabular}                              \\
IEEENLS-21 \cite{verma2021automer}    & Video                                                                        & Viola Jones                                                              & No                                                                              & NAS-3DCNN                                                                       & Yes                 & No                                                                    & No                           & SD-LOSO  
 \\
 Neu.Co-21 \cite{zhao2021two}    & Video   & {-}  &Yes: O.F. & 3D-CNN    & No     & No  & No  &\begin{tabular}[c]{@{}c@{}}SD-LOSO, \\ CD-LOSO\end{tabular}
 \\
  Sig. Proc. \cite{liu2021micro}    &    {Onset, Apex}   & {-}  &Yes: O. F.   & 3D-CNN+GA &No                  & No & No  &CD-LOSO
 \\PR-22 \cite{zhou2022feature}    & Apex                                                                         & LibFace, TV-L1                                                           & Yes: O. F                                                                       & Transfer-Learning                                                               & No                  & No                                                                    & Yes                          & \begin{tabular}[c]{@{}c@{}}SD-LOSO, \\ CD-LOSO\end{tabular}                                \\
 \hline                                                  
\end{tabular}
\begin{tablenotes}[para,flushleft]
   \textit{Here, Prepr., E-to-E, P. T. weights, N/W, On-A-Off, GA, and EVM represents preprocessing, end-to-end, pre-trained weights, network, Onset-Apex-Offset, Genetic Algorithm, and eulerian motion video motion.
   }
  \end{tablenotes}
  \end{threeparttable}
\end{table*}
\begin{table*}[!t]
\centering
\caption{Experimental settings and performance based comparison of existing DL approaches on PDE validation setup.}
\label{tab:4}
\begin{threeparttable}
\begin{tabular}{lcccccccccc}
\hline
\multirow{2}{*}{\textbf{Pub-Yr}} & \multirow{2}{*}{\textbf{Input Size}} & \multirow{2}{*}{\textbf{LR}} & \multirow{2}{*}{\textbf{Data Aug.}}                                                                 & \multirow{2}{*}{\textbf{Protocol}} & \multicolumn{3}{c}{\textbf{SMIC}}       & \multicolumn{3}{c}{\textbf{CASME-II}}   \\ \cline{6-11}
                                 &                                      &                              &                                                                                                     &                                    & \textbf{E} & \textbf{Acc} & \textbf{F1} & \textbf{E} & \textbf{Acc} & \textbf{F1} \\ \hline\hline
IJCNN-19 \cite{reddy2019spontaneous}                & $64\times 64$                                & Fixed                        & No                                                                                                  & 80/20 Split                        & 3          & 64.82        & N/A         & N/A        & N/A          & N/A         \\
TIP-19 \cite{verma2019learnet}                   & $112\times 112$                              & Fixed                        & F, R, T                          & 80/20 Split                        & 3          & 91.09        & N/A         & 7          & 76.57        & N/A         \\
ToM-20 \cite{xia2019spatiotemporal}                  & $300\times 245$                              & Adaptive                     & \begin{tabular}[c]{@{}c@{}}Temporal\\ connectivity based\end{tabular}                               & LOVO                               & 3          & 74.9         & 71.4        & 4          & 83.3         & 80.7        \\
ACMMM-20 \cite{xie2020assisted}               & N/A                                  & N/A                          & GAN based                                                                                           & LOVO                               & N/A        & N/A          & N/A         & 7          & 51.9         & 42.4        \\
IEEE-Acc-21 \cite{thuseethan2020complex}             & $128\times 128$                              & Adaptive                     & F, R, T & 90/10 Split                        & N/A        & N/A          & N/A         & 5          & 90.0         & 87.0       \\ \hline
\end{tabular}
\begin{tablenotes}[para,flushleft]
   \textit{Here, F, R, T, C.S, and  stands for flipping, rotating, transmitting, color shift and smoothing respectively.}
  \end{tablenotes}
  \end{threeparttable}
\end{table*}
\subsubsection{Multi-Scale Networks}\label{mul2stage}
Multi-scale  feature  representations  have  been  successfully used in two-stage MER \cite{zhou2019dual, vermaAff}. Zhou \etal \cite{zhou2019dual} a dual-inception network that operates in three scales ($1\times 1$, $3\times 3$ and $5\times 5$)  for  feature  encoding  of  MEs.  Verma et  al.  \cite{verma2019learnet}  introduced  a  multi-scale  based  lateral  assertive hybrid  network  to  capture  the  micro-level  features  of  an expression  in  the  facial  regions.  Zhai \etal \cite{zhai2021displacement} extended the the LearNet approch and introduced the displacement generating module based MER (DGMER) framework. Song  et  al.  \cite{song2019recognizing}  encodes spatio-temporal features by employing $5\times 5$ and $3\times 3$ sized filters in a consecutive manner. Verma \etal \cite{verma2020non} improve the robustness of MER with hybrid (fusion of $3\times 3$ and $5\times 5$) local receptive feature blocks. Furthermore, Verma \etal \cite{vermaAff} designed the AffectiveNet by incorporating MICRoFeat block to conserve the scale-invariant features with$3\times 3$, $5\times 5$, $7\times 7$ and $11\times 11$ sized convolution (conv) filters.\\
\textbf{\textit{Discussion:}} Similar to multi-stream MER frameworks, multi-scale  MER  frameworks also  achieve  impressive  performance.  The  multi-scale  convolution  layers  guide  the network  towards  both  minute  and  abstract  level  features,which  are  significant  to  describe  the  distinctive  features of  different  micro  expressions. The  more  detailed  analysis based on literature results have been included in section \ref{techChar} and Table \ref{tab:5}-\ref{tab:6}. 

\subsubsection{Capsule Networks}
CNNs have shown impressive performance in literature. However, CNN models are computationally expensive and need a lot of data to train a model for specific-domain tasks. Moreover, the CNN model can pay attention to the translation in variance but failed to learn the rotation in variance. To resolve these issues, Sabour \etal \cite{sabour2017dynamic} introduced the concept of Capsule. Capsule is a group of neurons to maintain the part-whole relationship and handle the viewpoint in variance. Some two-stage MER approaches \cite{van2019capsulenet, liu2020offset}  have exploited the capability of Capsule networks. Very first, Quang \etal \cite{van2019capsulenet} used the Capsule networks along with ResNet 18 and secured 4$^{th}$ position in MEGC-2019 challenge). Liu \etal \cite{liu2020offset}  utilized the Capsule module with a multi-stream CNN for MER.\\
\textbf{\textit{Discussion:}}
The capsule based networks can handle both translation and rotation variations and design a more robust MER network as compared to CNNs. Moreover, capsule networks facilitate the more concrete features represented, which can be interpreted to understand the behaviour of the network (how the network is learning the MEs’ features). However, the capsule based MER approaches are not widely notable due to complex nature and computational cost, though they have shown great promise. The capsule-based networks are still evolving and there is lots of scope for the researchers to create better and faster architectures so that it will be the baseline for solving any expression (MaEs or MEs) classification problem.

\subsubsection{Recurrent Convolutional Networks}
Recurrent convolutional networks (RCNs) enable every unit to incorporate context information in an arbitrarily large region in the current conv layer and allows learning microlevel edge variations. In literature, Xia \etal \cite{ xia2019cross, xia2019spatiotemporal, xia2020revealing} exploit the RCN by following two-stage architecture to learn the representation of subtle facial movements from image sequences. In these studies the recurrent connection
within the feed-forwarded conv layers are employed to learn the temporal variations of image sequences extracted by multiple-scale receptive fields. \\
\textit{\textbf{Discussion:}}
CNN’s were inspired by early findings in the study of biological vision and share properties of the visual system of the brain. One notable distinction is that CNN is often a feed-forward design, whereas the visual system of the brain is abundant with recurrent connections. Thus, RCN-based architectures are benefited with more microbiologically realistic than their feed-forward counterparts. In addition, the activities of RCN units evolve over time as the activity of each unit is modulated by the activities of its neighbouring units, which allows the network to learn distinctive edge variations with temporal information in MEs under challenging conditions over CNN as reported in Table \ref{tab:4}-\ref{tab:7}.
\subsubsection{3D-CNN Models}\label{3d2stage}
Most of the existing MER approaches \cite{van2019capsulenet, li2020joint, liu2019neural} rely only on the apex frame/single instance for the analysis of MEs through 2D-CNNs. However, some studies emphasize the importance of dynamic aspects for detecting the subtle changes \cite{ambadar2005deciphering} and its effect on the performance of MER. In MEs video, each frame has its own significance towards the identification of the emotion class. Whereas some other CNN models \cite{wang2020micro, yang2019merta, khor2018enriched}, exploit the capability of 2D CNN and LSTM/RNN to elicit the spatial and temporal features, respectively. However, these models are not capable of extracting joint features of spatial and temporal variations, simultaneously \cite{reddy2019spontaneous}. Therefore to overcome the above issues, recently, some of MER approaches \cite{ststnet, Li3D-Flow, zhao2021two, liu2021micro} have taken advantage of the 3D-CNN network to capture, both spatial and temporal features simultaneously by adopting two stage architecture. \\
\textbf{\textit{Discussion:}} The 3D convolutional layers are reasonable to learn the spatio-temporal information simultaneously. However, 3D-CNNs do not gain much attention in MER as it holds huge parameters and requires more computation power as compared to others such as 2D-CNN, CNN-LSTM, RCN \textit{etc.} Also, deciding the number of hyper parameters such as layers, 3D down-sampling, and number of filters,  in the network is a challenging task.  
\subsection{End-to-End Frameworks}\label{endtoend}
An end-to-end model means that the CNN model takes the raw data as input and gives the final response without any aid of external modules or blocks (see supplementary Fig. 5a). Based on characteristics of the deep learning models, end-to-end frameworks are also categorized as: multi-scale, CNN-LSTM, 3D-CNN, Graph based, and NAS based MER.

\subsubsection{Multi-Scale Networks}
As discussed in Section \ref{mul2stage}, multi-scale networks have shown great performance in two-stage networks. Similarly, to exploit the capability of multi-scale conv layers, Wang \etal \cite{wang2020micro} proposed an end-to-end micro-attention for MER by utilizing two scales $1\times 1$ and $3\times 3$ to encode the micro expressive features.\\
\textbf{\textit{Discussion:}}
Similar to multi-stage networks, the  multi-scale  convolution  layers also achieve impressive performance in case of end-to-end MER models. MEs are holding very sensitive and subtle information, thereby extracting the minute muscle changes within coarse facial features. It is important to learn minute as well as abstract features of the facial appearance and multi-scale convolution layers learn the distinctive features of true emotions.  Thus, multi-scale convolutional layers can be embedded in any type of deep learning MER models  to  extract  the  disparities between different  micro  expressions.

\subsubsection{CNN-LSTM Models} 
In CNN-LSTM networks, CNN is used to extract the spatial features and LSTM is included to learn the time-scale dependent information that resides along with the frame sequences. First, Kim \etal \cite{kim2016micro} used 2D-CNN followed by LSTM to encode the spatial and temporal features in MEs videos. Similarly, other work \cite{khor2018enriched,wang2018micro, yang2019merta} also utilized the combination of CNN and LSTM based conv networks to design end-to-end MER frameworks. CNN is employed to encode the MEs frames into spatial feature vectors and then MEs classes are predicted by passing the resultant features through the LSTM module. Choi \etal \cite{choi2020facial} introduced an integrated framework of CNN and LSTM for landmark feature map-based MER.\\
\textbf{\textit{Discussion:}} The CNN-LSTM based MER framework allows to capture and classify the spatio-temporal features of MEs in an end-to-end manner. However, CNN-LSTM networks first learn the spatial features and then temporal features. Therefore, sometimes these networks fail to correlate the spatial and temporal features simultaneously and fail to achieve good performance. Thus, 3D-CNN based models are introduced to resolve the issue of CNN-LSTM and learn the spatio-temporal features simultaneously. The detailed insights of models are discussed in Section  \ref{techChar}.

\subsubsection{3D-CNN Models}
As discussed in Section \ref{3d2stage}, cascaded MER architectures such as CNN with LSTM or RNN are not capable of extracting joint features of spatial and temporal variations, simultaneously \cite{reddy2019spontaneous}. The capability of 3D-CNN to describe the spatio-temporal features for MEs in an end-to-end manner was first presented in \cite{reddy2019spontaneous}. 
{Xie \etal \cite{xie2020assisted} adopts the 3D ConvNet based Pseudo-3D to design a light weighted end-to-end architecture. Furthermore, AU node features are extracted and processed through the AU graph relation learning module to describe the emotion classes of MEs.}\\
 
\subsubsection{Graph based CNN models}
Recently, the graph based end-to-end CNN approaches have achieved attention and attracted the researchers \cite{lei2020novel, xie2020assisted, kumar2021micro, lei2021micro} in the field of MER. Lei \etal \cite{lei2020novel} exploited the capability of landmarks and proposed a graph temporal CNN (Graph-TCN) to capture the local muscle movements of the MEs. The Graph-TCN method consists of two streams: node and edge feature extraction streams. Finally, both node and edge features are merged to classify the emotion label of MEs. Xie \etal \cite{xie2020assisted} used the AUs relation graph to learn the subtle facial muscle movements for MER.\par

{Each expression originated due to facial muscle movements and divided the face into small regions named action units (AUs) to represent the affective expression regions, defined in Facial Action Coding System (FACS) \cite{ekman1978facial} and Micro-Expression Training Tool (METT) \cite{ekman2003mett}. Thus, AUs are frequently used to describe how emotions are physically expressed. Most of the graph-based deep learning models incorporated the AUs relationship. Lo \etal \cite{lo2020mer} introduced MER-GCN, an AU-oriented MER architecture based on Graph Convolutional Network (GCN) \cite{welling2016semi}, where GCN layers are able to explore the dependency laying between AU nodes for MER. Similarly, Xie \etal \cite{xie2020assisted} and Lei \etal \cite{lei2021micro} exploit utilized the GCN to discover the AUs relationship. Lei \etal \cite{lei2020novel} utilized the graph structure for node and edge feature extraction.\\}
\textbf{\textit{Discussion:}} The graph based models are growing gradually in MER. The new affective computing researchers have a lot of scope in the graph based FER/MER approaches. The graph based models are complex to understand but need very less computation cost with great efficiency, which is the future demand to work with memory limited or hand-held devices.

\subsubsection{GANs based MER}\label{gan}
With the advancement in learning based methods, the performance of the MER is improved but still limited due to the lack of large-scale training data, computation and design expertise. Some recent works \cite{yu2020ice, xie2020assisted, xia2020learning}, have been focused on these issues and provide solutions by exploiting the power of GANs. Yu \etal \cite{yu2020ice} introduced a capsule enhanced GAN to generate the synthetic MEs with identity aware faces to increase the data-samples for better training. Xie \etal \cite{xie2020assisted} came up with the AU intensity controller GAN to produce synthetic data for resolving the problem of limited and biased data-samples.\\
\subsubsection{NAS based MER}
The robust deep learning models are designed manually based on trial-and-error engineering and need expert knowledge, further this is very time consuming and requires high level domain expertise in designing CNN networks. Very first, Verma \etal \cite{verma2021automer} focused on these factors and came up with a conclusion that instead of spending time and effort designing the best possible CNN, in the hope of improved performance, it is prudent to design algorithms to search for the best CNN model for MER. \\
\textbf{\textit{Discussion:}} NAS algorithms are able to search and automatically design the best optimal CNN model with minimum human intervention. Initially, NAS based algorithms \cite{liu2018darts,liu2019auto} required huge computation and took many days to train a task specific CNN model. However, recent NAS based approaches \cite{zheng2021migo} focus on the faster searching and training architectures. Specifically, in MEs, NAS based algorithms need extra efforts to design MEs feature adaptive inner as well as outer architecture search. The MER field is still far from automatic model designing and has a lot of scope to develop better and faster NAS based MER algorithms. Also, selecting robust operation and number of cells for MER application is one of the prominent steps for the performance improvements in MER. 
\begin{table*}[!t]
\centering
\caption{ Experimental settings and performance based comparison of existing DL approaches on PIE validation setup.}
\label{tab:5}
\begin{tabular}{lcccccccccc}
\hline
\multirow{2}{*}{\textbf{Pub-Yr}} & \multirow{2}{*}{\textbf{Input Size}} & \multirow{2}{*}{\textbf{LR}} & \multirow{2}{*}{\textbf{Data Aug.}}                                                        & \multirow{2}{*}{\textbf{Protocol}} & \multicolumn{3}{c}{\textbf{SMIC}}           & \multicolumn{3}{c}{\textbf{CASME-II}}        \\ \cline{6-11}
                                 &                                      &                              &                                                                                            &                                    & \textbf{E}   & \textbf{Acc}  & \textbf{F1}  & \textbf{E} & \textbf{Acc}   & \textbf{F1}    \\ \hline \hline
ACM-2016 \cite{kim2016micro}               & $64\times 64$                               & {N/A}                 & F, R, T   & {LOSO}                      & {N/A} & {N/A}  & {N/A} & {5} & {60.98} & {N/A}   \\
ICPR-16 \cite{patel2016selective}                 & $48\times 48$                                & N/A                          & N/A                                                                                        & 4Fold-LOSO                         & {3}   & {53.6} & N/A          & {5} & {47.30} & N/A            \\
FG-18 \cite{peng2018macro}                   & $224\times 224$                              & Adaptive                     & F, R, T              & SD-LOSO                            & {N/A} & {N/A}  & {N/A} & {3} & {75.68} & {65.0}  \\
ICIP-18 \cite{li2018can}                & N/A                                  & N/A                          & N/A                                                                                        & SD-LOSO                            & {N/A} & {N/A}  & {N/A} & {5} & {63.30} & {N/A}   \\
IPTA-18 \cite{xia2018spontaneous}                & $1966\times 30$                              & Adaptive                     & N/A                                                                                        & 5Fold-LOSO                         & {3}   & {57.1} & t{N/A} & {4} & {63.2}  & {N/A}   \\
FG-18 \cite{khor2018enriched}                  & $224\times 224$                              & Adaptive                     & No                                                                                         & SD-LOSO                            & {N/A} & {N/A}  & {N/A} & {5} & {57.00} & {41.07} \\
MTA-19 \cite{yang2019merta}                & N/A                                  & Adaptive                     & N/A                                                                                        & SD-LOSO                            & {N/A} & {N/A}  & {N/A} & 5          & 60.54          & N/A            \\
SP-IC-10 \cite{gan2019off}               & $28\times 28$                                & Fixed                        & No                                                                                         & SD-LOSO                            & 3            & 67.8          & 67.09        & 5          & 88.28          & 86.27          \\
PAA-19 \cite{Li3D-Flow}                 & $226\time 226$                              & N/A                          & No                                                                                         & SD-LOSO                            & 3            & 55.49         & {N/A} & 5          & 59.11          & {N/A}   \\
ICIP-19 \cite{khor2019dual}                 & N/A                                  & Fixed                        & No                                                                                         & SD-LOSO                            & 3            & 63.41         & 64.62        & 5          & 70.78          & 72.97          \\
IEEE-Acc-19 \cite{song2019recognizing}            & $48\times 48$                                & Fixed                        & N/A                                                                                        & SD-LOSO                            & 3            & 72.74         & 72.36        & 5          & 80.97          & 80.70          \\
Neu.Co-19 \cite{wang2020micro}               & $224\times 224$                              & Adaptive                     & C.S, R, S               & SD-LOSO                            & 3            & 49.4          & 49.6         & 5          & 65.9           & 53.9           \\
IJCNN-20 \cite{verma2020non}                 & $112\times 112$                              & Fixed                        & F, R, T      & SD-LOSO                            & N/A          & N/A           & N/A          & 4          & 62.09          & N/A            \\
ToM-20 \cite{xia2019spatiotemporal}                  & $300\times 245$                              & Adaptive                     & Temporal connectivity                     & SD-LOSO                            & 3            & 72.3          & 69.5         & 4          & 80.3           & 78.6           \\
TAFF-20 \cite{sun2020dynamic}                 & $132\times 132$                              & N/A                          & N/A                                                                                        & SD-LOSO                            & 3            & 76.06         & 77.0         & 5          & 72.61          & 67.0           \\
ACMMM-20 \cite{xie2020assisted}              & N/A                                  & N/A                          & GAN Based                                                                                  & SD-LOSO,                           & 3            & 70.2          & 43.3         & 7          & 56.1           & 56.1           \\
ACMMM-20 \cite{xia2020learning}               & $224\times 224$                              & Fixed                        & No                                                                                         & SD-LOSO                            & 3            & 75.6          & 70.1         & 5          & 75.6           & 70.1           \\
ACMMM-20 \cite{lei2020novel}                & $256\times 256$                              & N/A                          & F, R, T      & SD-LOSO                            & N/A          & N/A           & N/A          & 5          & 73.98          & 72.46          \\
IEEE-Acc-20 \cite{choi2020facial}            & $21\times 21$                                & N/A                          & No                                                                                         & SD-LOSO                            & 3            & 71.34         & 71.34        & 5          & 71.54          & 70.26          \\
TIP-21 \cite{li2020joint}                 & $224\times 224$                              & 224                          & \begin{tabular}[c]{@{}c@{}}5 Nearest frame \\ to apex\end{tabular}                         & SD-LOSO                            & 3            & 63.41         & 62.1         & 5          & 62.14          & 60.0           \\
IEEE-Acc-21 \cite{thuseethan2020complex}            & $128\times 128$                              & Adaptive                     & F, R, T & SD-LOSO                            & N/A          & N/A           & N/A          & 5          & 86.0           & 82.0           \\
IEEE-MM-21 \cite{vermaAff}                & $112\times 112$                              & Fixed                        & F, R, T      & SD-LOSO                            & N/A          & N/A           & N/A          & 4          & 68.74          & N/A            \\

TAFF-21 \cite{gupta2021merastc}                  & N/A                                  & Adaptive                     & F, R, T      & SD-LOSO                            & 3            & 79.3          & 79.0         & 5          & 85.4           & 86.2           \\
TNLS-21 \cite{verma2021automer}                  & $120\times 120$                              & Adaptive                     & {No}                                                                                & SD-LOSO                            & 3            & 81.20         & 51.7         & 4          & 74.15          & 55.44       \\ 
Neu. Co-21 \cite{zhao2021two} & $112\times 112$ & Adaptive  & {Yes} & SD-LOSO                            & N/A            & N/A        & N/A        & 5          & 81.89          & 83.00       \\PR-22 \cite{zhou2022feature}               & $28\times 28$                                & N/A                          & {N/A}                                                                               & SD-LOSO                            & 3            & 57.90         & N/A          & 5          & 62.38          & N/A            \\ \hline  
\end{tabular}
\end{table*}
\subsection{Transfer learning based Networks}\label{transferlearning}
{All available datasets for MER are relatively smaller as compared to other computer vision tasks. However, it is a well-known fact that the direct training of deep networks from scratch over smaller datasets is prone to overfitting. To mitigate the effect of overfitting, many studies have taken advantage of pre-trained weights (fine-tuning) of well-known models: AlexNet \cite{krizhevsky2012imagenet}, VGG \cite{Simonyan2014VeryDC}, ResNet \cite{He2015DeepRL}, \textit{\textit{etc.}}, which are trained over large-scale datasets such as ImageNet, FaceNet, \textit{etc.} However, Patel \etal  \cite{patel2016selective}  explore that all pre-trained weights are not fitting well to discriminate against the MEs due to the low intensity of facial movements. Although, features of MaEs and MEs share some feature similarities in facial texture and muscle movements. Therefore, to exploit the capability of pre-trained features by considering domain adaption, Patel \etal  \cite{patel2016selective}  retrained ImageNet weights for macro datasets: CK+ and SPOS to train the model for expressive features. Next, resultant features are fine-tuned over micro expression datasets. On the same hypothesis, in MEGC-18 \cite{yap2018facial} and MEGC-19 challenge \cite{see2019megc}, researchers \cite{peng2018macro, wang2020micro, liu2019neural} utilized the pre-trained weights of ResNet and its variants \cite{He2015DeepRL}, trained over ImageNet and further to learn the expression specific features model retrained over MaE datasets: CK+, OULU, JAFFE, and MUG. Further, by adopting the domain features of the expressions from MaEs the CNN models are fine-tuned for ME tasks. The pertained weights of ResNet-18 have been widely used in the literature \cite{peng2018macro, liu2019neural, xia2020learning} due to their feature learning capability. Further, Wang \etal \cite{wang2018micro} designed a CNN network: transferring long-term convolutional neural networks (TLCNN) by combining 2DCNN and LSTM. First, TLCNN is trained over MaE large-scale datasets: KDEF, MMI, and TFID to ac- quire the knowledge of emotion-specific features. Afterward, TLCNN weights are used to retrain the model for MER. Khor  \etal \cite{khor2018enriched} fine-tuned the VGG-16 with VGG-face weights to capture the enriched spatial feature of MEs. Moreover, Yang \etal \cite{yang2019merta} introduced a MERTA network by combining VGG-16 and LSTM to exploit the spatio-temporal features of ME sequences and their respective optical flow and -strain. Li \etal \cite{li2020joint}  have benefited from pre-trained weights VGG Face to guide the deep CNN network for smaller MEs datasets. Besides fine-tuning and domain adaption, knowledge distillation is another effective transfer learning approach. To provide a compact solution for training data and computation, Sun \etal \cite{song2019recognizing}  proposed knowledge distillation to transfer knowledge of AUs to MER through teacher-student CNN learning. The main aim of the framework is to guide the shallow student network by transferring the knowledge of features from a pre-trained deep teacher network.}\\
\textbf{\textit{Discussion:}} The pre-trained weights or transfer learning is a sure-fire concept to solve the problem of overfitting and speed up the learning process with smaller sized datasets. The well-trained models: VGG-16, ResNet, GoogleNet \textit{etc.} benefited the MER approaches to reduce the problem of overfitting up-to some extent. However, most of these models are trained over ImageNet dataset, which has contrast data samples related to MEs with low muscle intensity, subtle and rapid changes. Therefore, pre-trained weights of ImageNet are not suitable for MER. Whereas pre-trained weights over face images or macro datasets are more advisable as these datasets hold analogous features in terms of facial structure and shape. Moreover, pre-trained weights can be utilized in both multi-stage and end-to-end MER networks. More detailed technical characteristics are discussed in section \ref{techChar}.

\subsubsection{End-to-end vs multi-stage frameworks}
In a two-stage network, a handcrafted feature descriptor such as dynamic imaging \cite{verma2019learnet}, affective motion imaging \cite{vermaAff}, optical flows \cite{ststnet, Li3D-Flow, liu2021micro} \textit{etc.}, is used in the first stage to capture the primary features. While, in the second stage, the CNN network is used to learn stage-1 features. For example, Verma \etal  \cite{verma2019learnet, vermaAff} utilized the dynamic and affective motion imaging to capture the spatiotemporal features into a single instance and then CNN network is designed to learn the MEs features (see supplementary Fig. 2a and 3b). Therefore, the performance of the CNN models is also dependent on the handcrafted feature descriptors. Moreover, the two-stage networks require auxiliary computation to compute the handcrafted features and inconvenience in real-time applications. Whereas, end-to- end MER architectures process the data into single shot and advisable for real-time applications. The model categorization in terms of end-to-end and two-stage is represented in Table \ref{tab:3}. Moreover, efficacy of end-to-end and two-stage networks over different experimental settings: number of classes, input formats, validation protocols, learning rates, data augmentation, \textit{etc.} is reported in Table \ref{tab:4}-\ref{tab:7}.

\subsection{Discussion on technical characteristics of DL based MER}\label{techChar}
This section represents the deep insights of technical characteristics in various decisive aspects such as down sampling, multi-scale/-stream CNNs, shallow v/s deeper CNNs, and effects of kernel sizes in convolutional layers of the deep learning-based MER approaches.

\begin{figure}[!t]
     \centering
     \begin{subfigure}[b]{0.48\linewidth}
         \centering
         \includegraphics[width=\textwidth, height=1.6in]{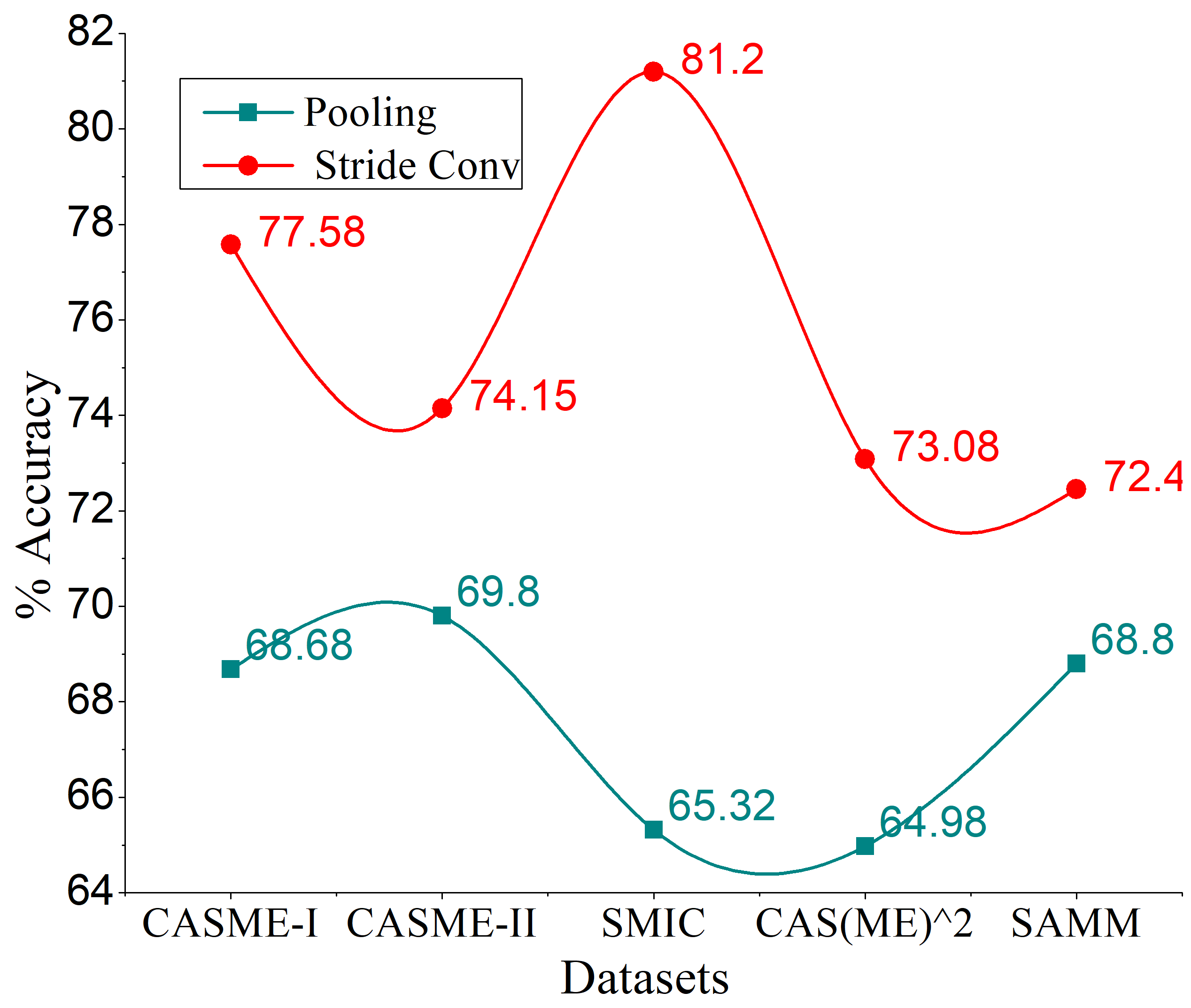}
         \caption{}
         \label{fig:9a}
     \end{subfigure}
     \hfill
     \begin{subfigure}[b]{0.48\linewidth}
         \centering
         \includegraphics[width=\textwidth, height=1.6in]{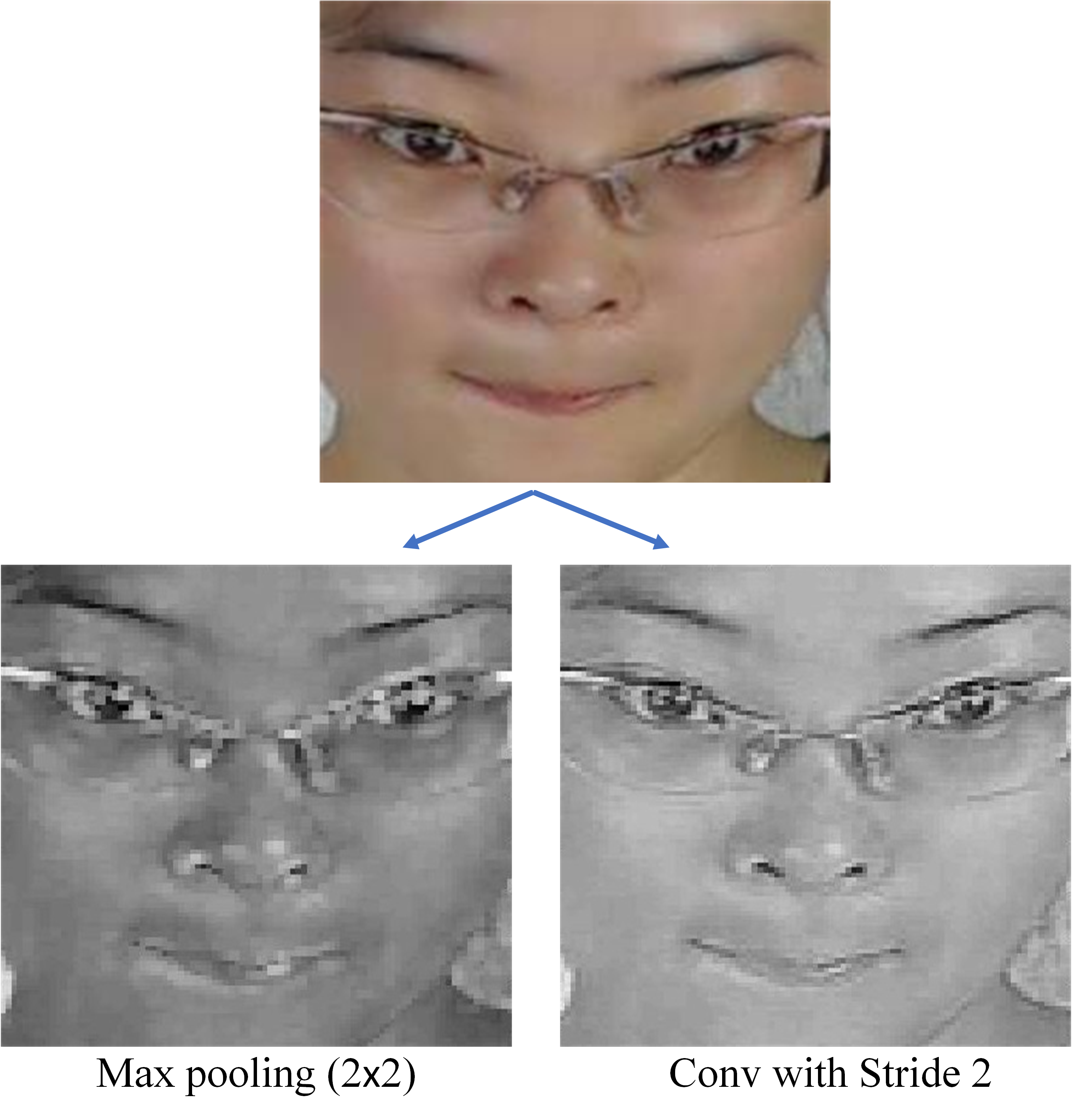}
         \caption{}
         \label{fig:9b}
     \end{subfigure}
     
        \caption{The performance analysis with pooling and conv layer with stride 2 in (a) quantitative \cite{verma2021automer} and (b) qualitative \cite{verma2019learnet}, manner.}
        \label{fig:9}
\end{figure}
\subsubsection{Impact of downsampling with convolution and pooling}
Down sampling plays a significant role to reduce the number of parameters and ensures higher computational speed in CNN framework. In the case of MER, majorly, max pooling and convolution with stride operations were used for dimensionality reduction. The max-pooling layer captures the high-level edge information by applying max operation. However, max pooling operations tend to focus on high level stack information by ignoring minors, which play a key role in the recognition of micro level features. However, the convolution adds to the inter-feature dependencies and reduces the dimension by parameter learning among channels rather than fixing it. The impact of convolution with strides over max pooling operations on MER datasets are shown in Fig. \ref{fig:9}. The Fig. \ref{fig:9a} represents that the convolution with stride gains significant improvement in accuracy as compared to pooling over CASME-I, CASME-II, SMIC, CAS(ME)$^2$ and SAMM datasets. Moreover, Fig. \ref{fig:9b}  depicts the visual effects of the max pooling over convolution with stride. From Fig. \ref{fig:9b}, it is clearly visible that the response map of pooling loses more information as compared to convolution. The more details about the technical differences with qualitative and quantitative measures are discussed in \cite{verma2018expertnet, verma2019learnet, vermaAff}. From Fig. \ref{fig:9} and \cite{verma2018expertnet,verma2019learnet, vermaAff} it is clear that convolution with stride outperforms the max pooling operation in MER.

\begin{figure}[!t]
     \centering
     \begin{subfigure}[b]{1\linewidth}
         \centering
         \includegraphics[width=0.9\textwidth, height=0.8in]{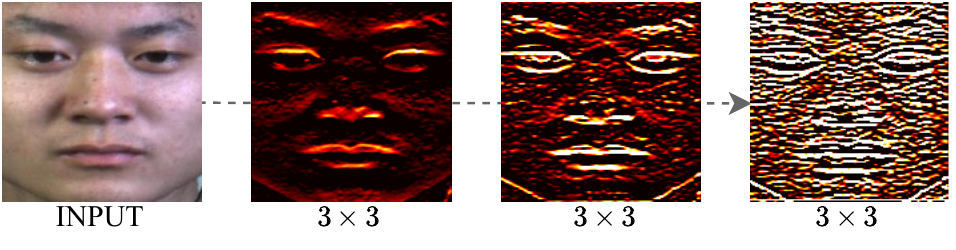}
         \caption{}
         \label{fig:10a}
     \end{subfigure}
     \hfill
     \begin{subfigure}[b]{1\linewidth}
         \centering
         \includegraphics[width=0.9\textwidth, height=3.6in]{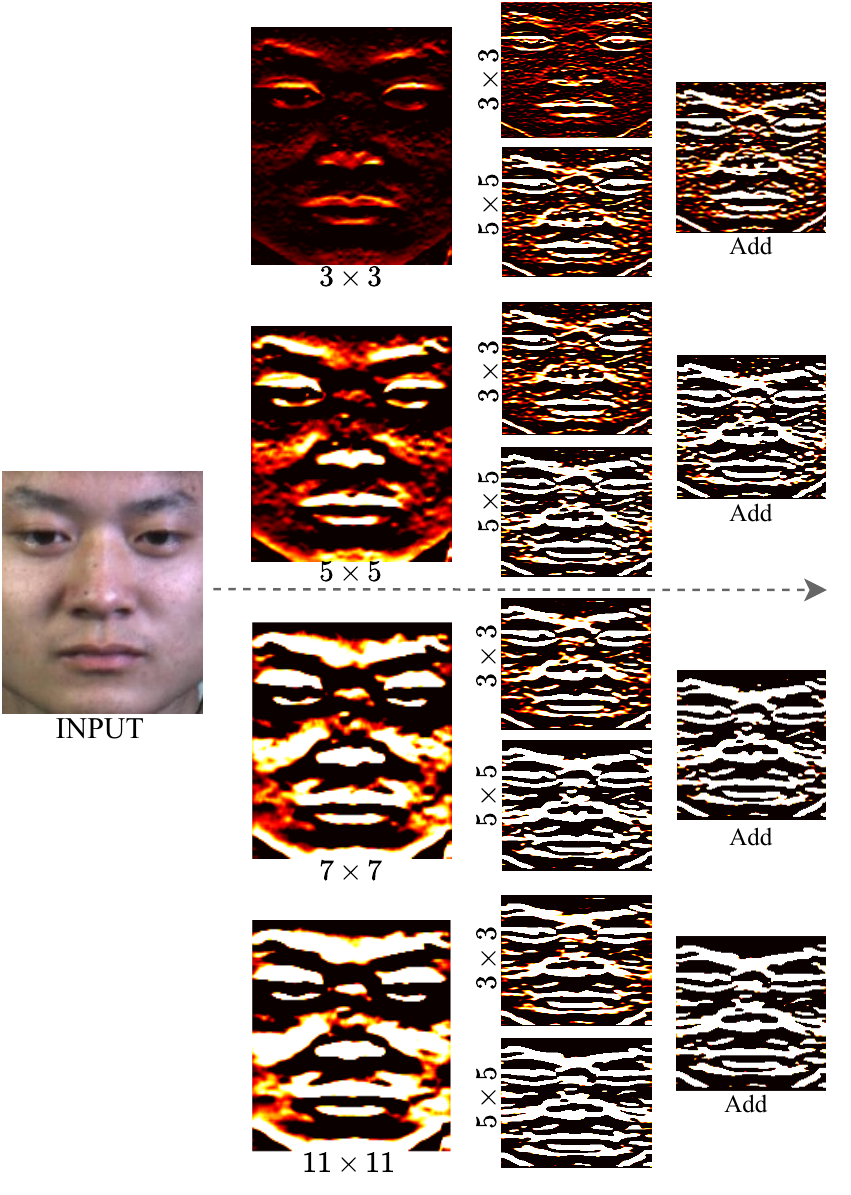}
         \caption{}
         \label{fig:10b}
     \end{subfigure}
     
        \caption{The feature maps responses generated by employing (a) single-scale and (b) multi-scale conv layer in multi-stream and linear framework, respectively.}
        \label{fig:10}
\end{figure}

\begin{figure*}[!t]
     \centering
     \begin{subfigure}[b]{0.18\linewidth}
         \centering
         \includegraphics[width=\textwidth,height=1.3in]{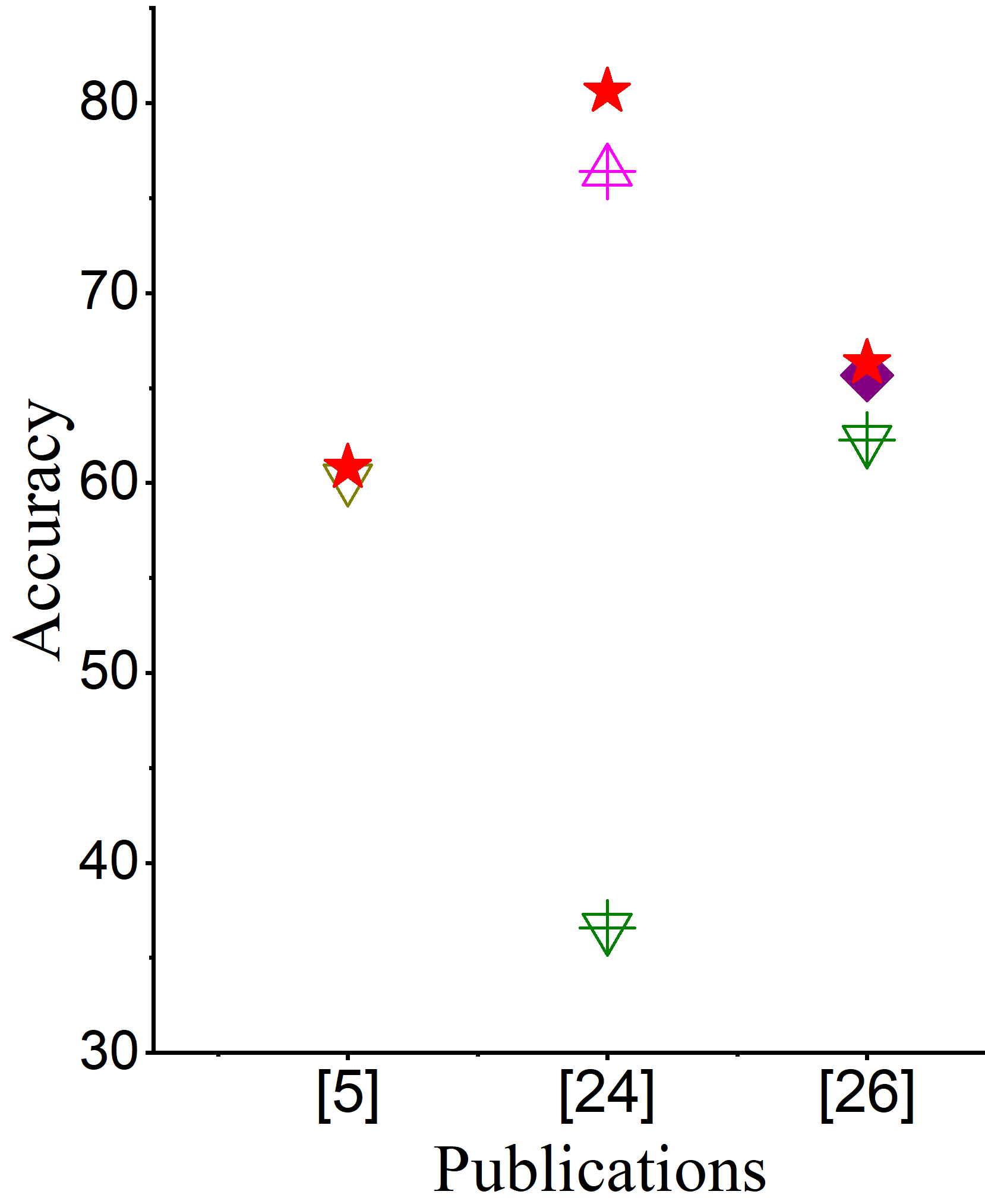}
         \caption{}
         \label{fig:11a}
     \end{subfigure}
     \hfill
     \begin{subfigure}[b]{0.18\linewidth}
         \centering
         \includegraphics[width=\textwidth,height=1.3in]{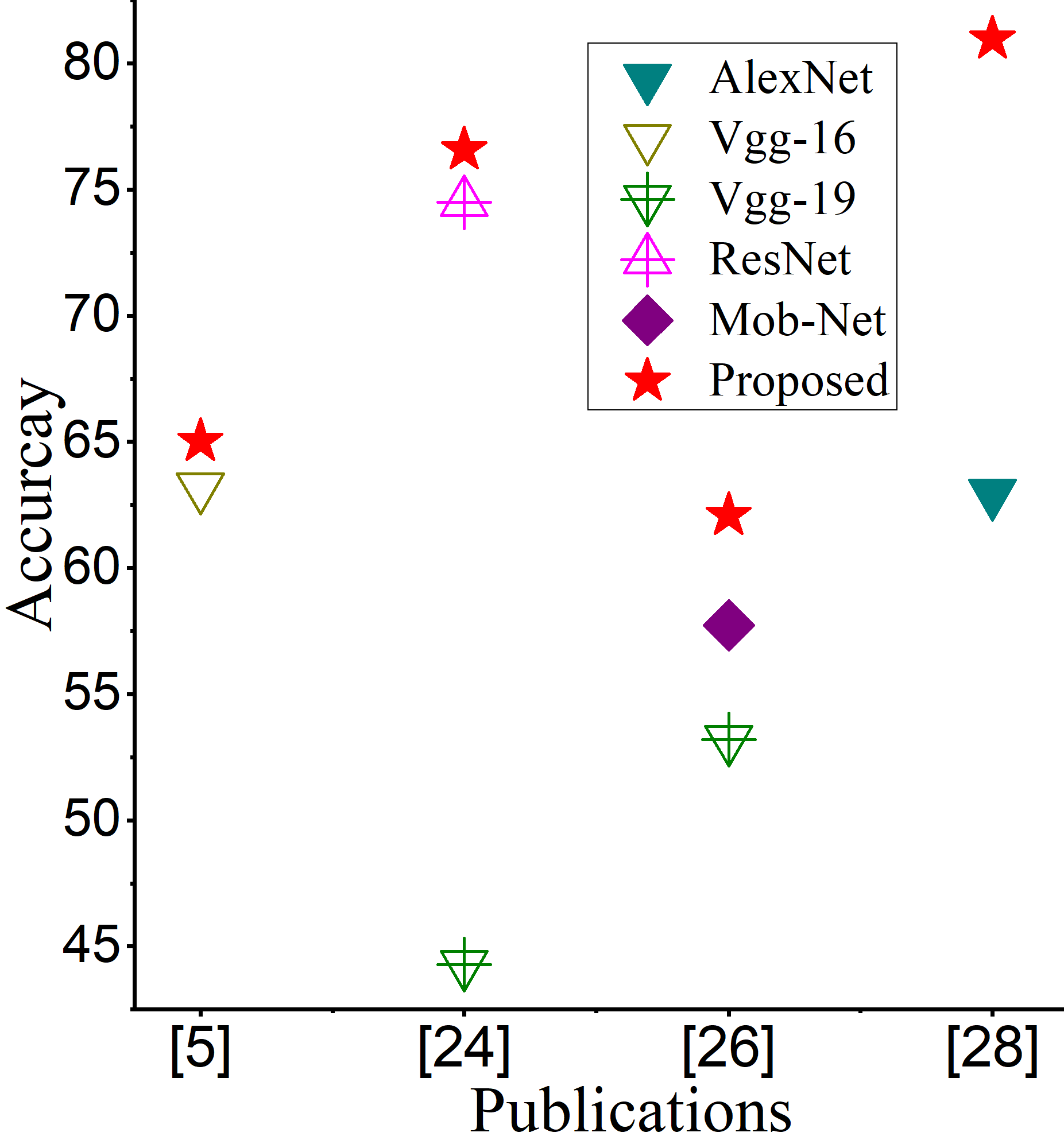}
         \caption{}
         \label{fig:11b}
     \end{subfigure}
      \hfill
     \begin{subfigure}[b]{0.18\linewidth}
         \centering
         \includegraphics[width=\textwidth,height=1.3in]{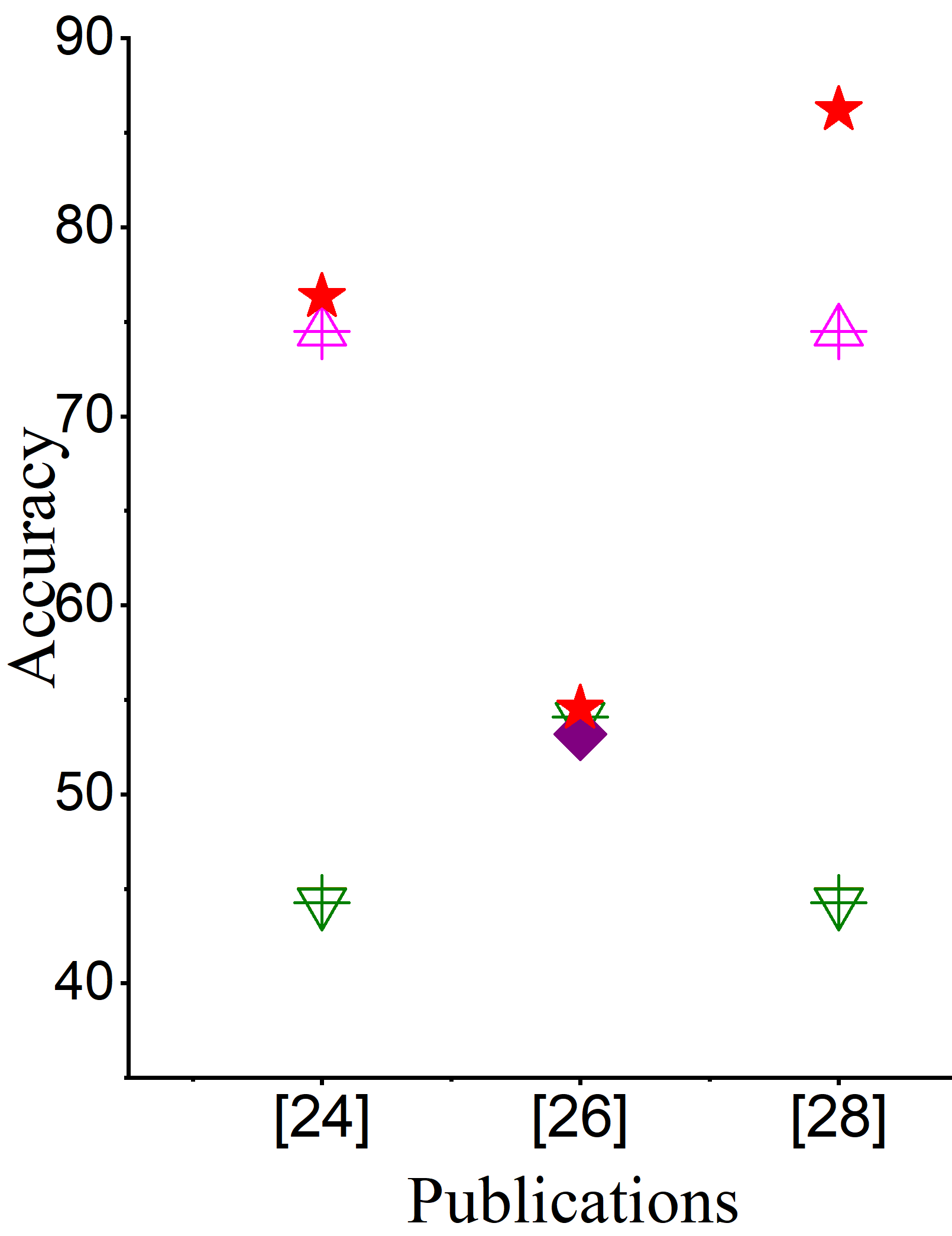}
         \caption{}
         \label{fig:11c}
     \end{subfigure}
      \hfill
     \begin{subfigure}[b]{0.18\linewidth}
         \centering
         \includegraphics[width=\textwidth,height=1.3in]{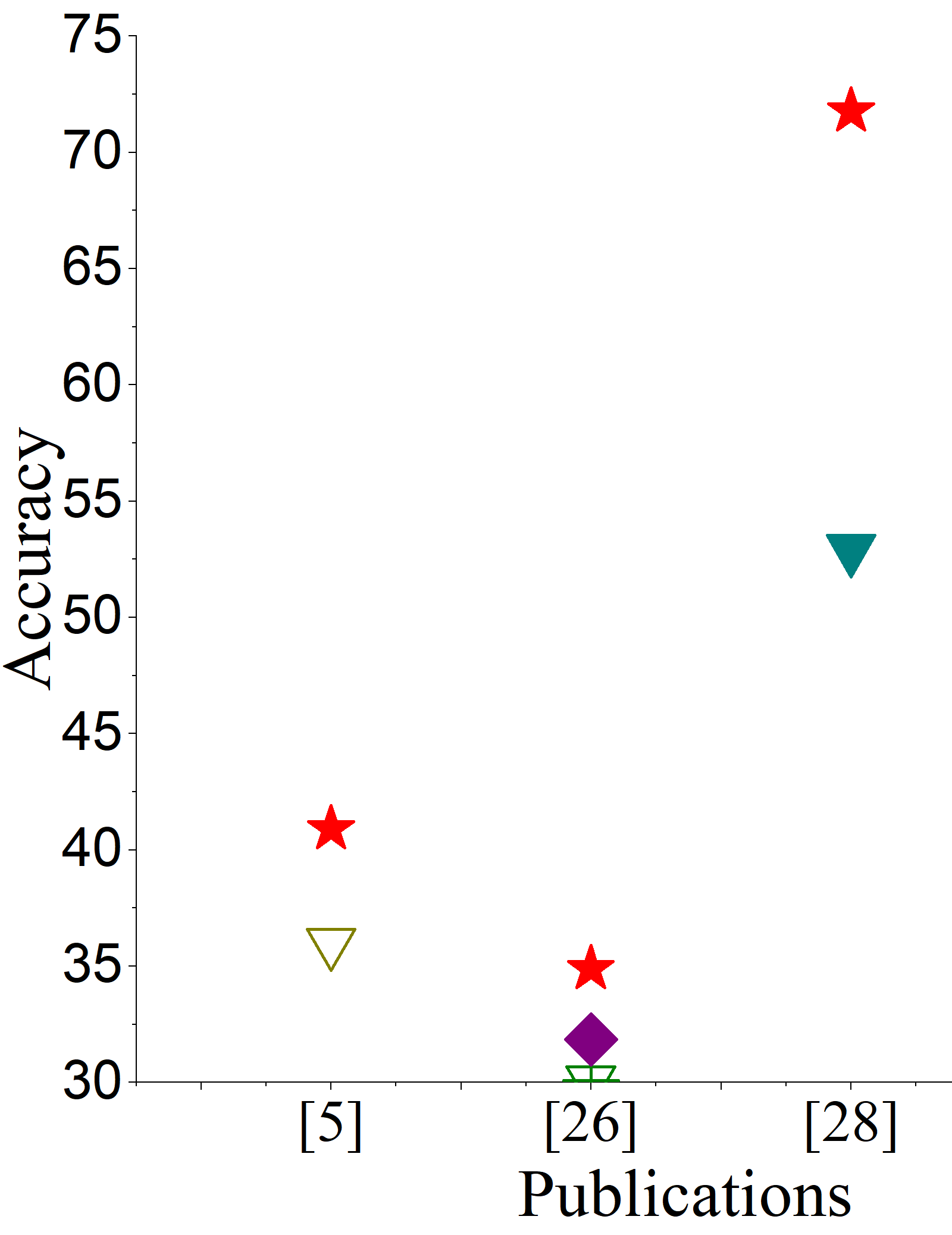}
         \caption{}
         \label{fig:11d}
     \end{subfigure}
      \hfill
     \begin{subfigure}[b]{0.18\linewidth}
         \centering
         \includegraphics[width=\textwidth,height=1.3in]{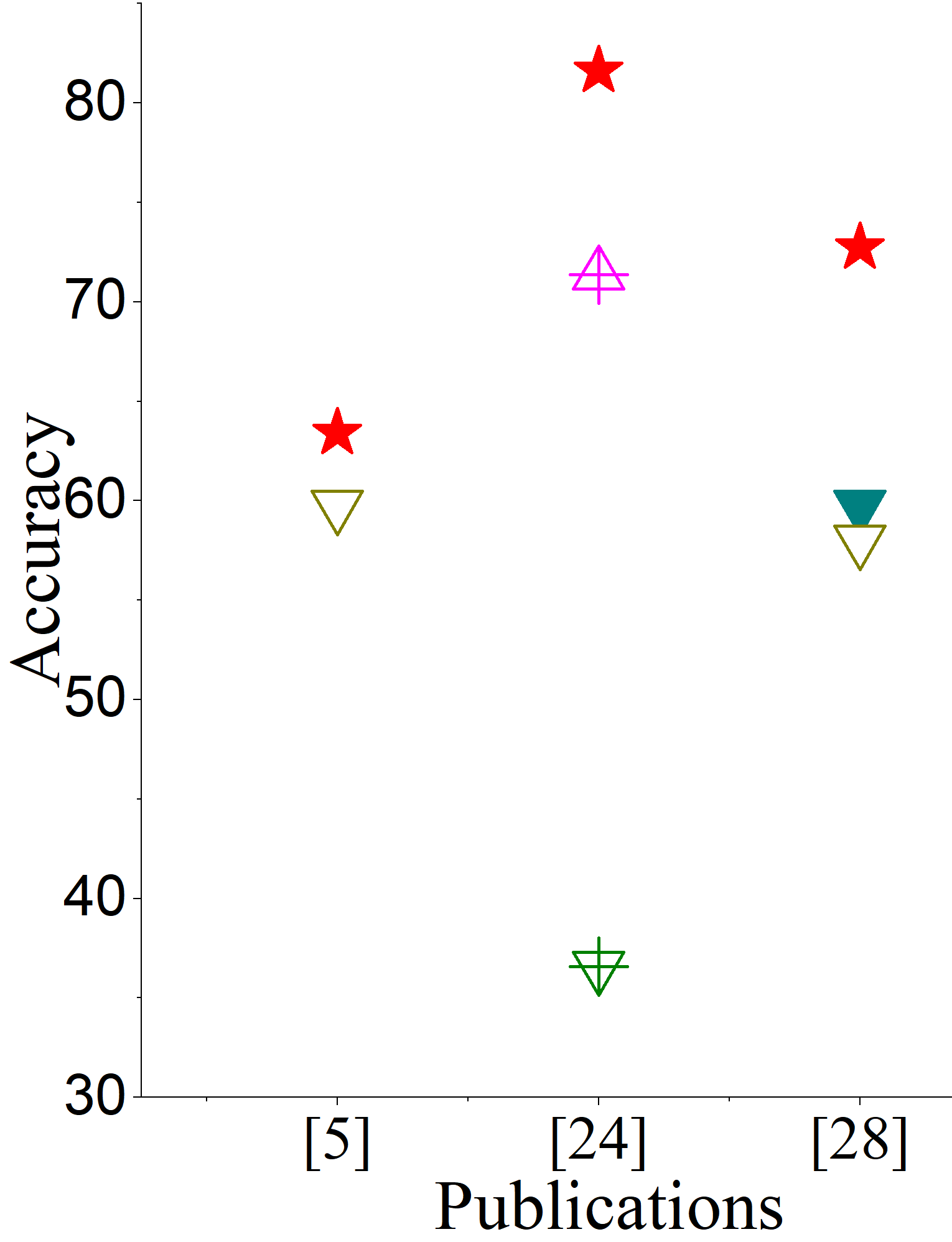}
         \caption{}
         \label{fig:11e}
     \end{subfigure}
     
        \caption{The recognition rate analysis of multi-scale/stream (LearNet \cite{verma2019learnet}, OrigiNet \cite{verma2020non}, LGConv \cite{li2020joint} and TSCNN \cite{song2019recognizing}) and single-scale/ linear (AlexNet \cite{krizhevsky2012imagenet}, VGG-16 \cite{Simonyan2014VeryDC}, VGG-19 \cite{Simonyan2014VeryDC}, ResNet \cite{He2015DeepRL} and Mob-Net \cite{mobilenet}) CNN frameworks on five datasets (a) CASME-I, (b) CASME-II, CAS(ME)$^2$, SAMM and SMIC. \textit{Here, Proposed implies for the proposed approach in that particular publication (X coordinate).}}
        \label{fig:11}
\end{figure*}
\begin{figure}[!bp]
    \centering
    \includegraphics[width=0.9\linewidth, height=2in]{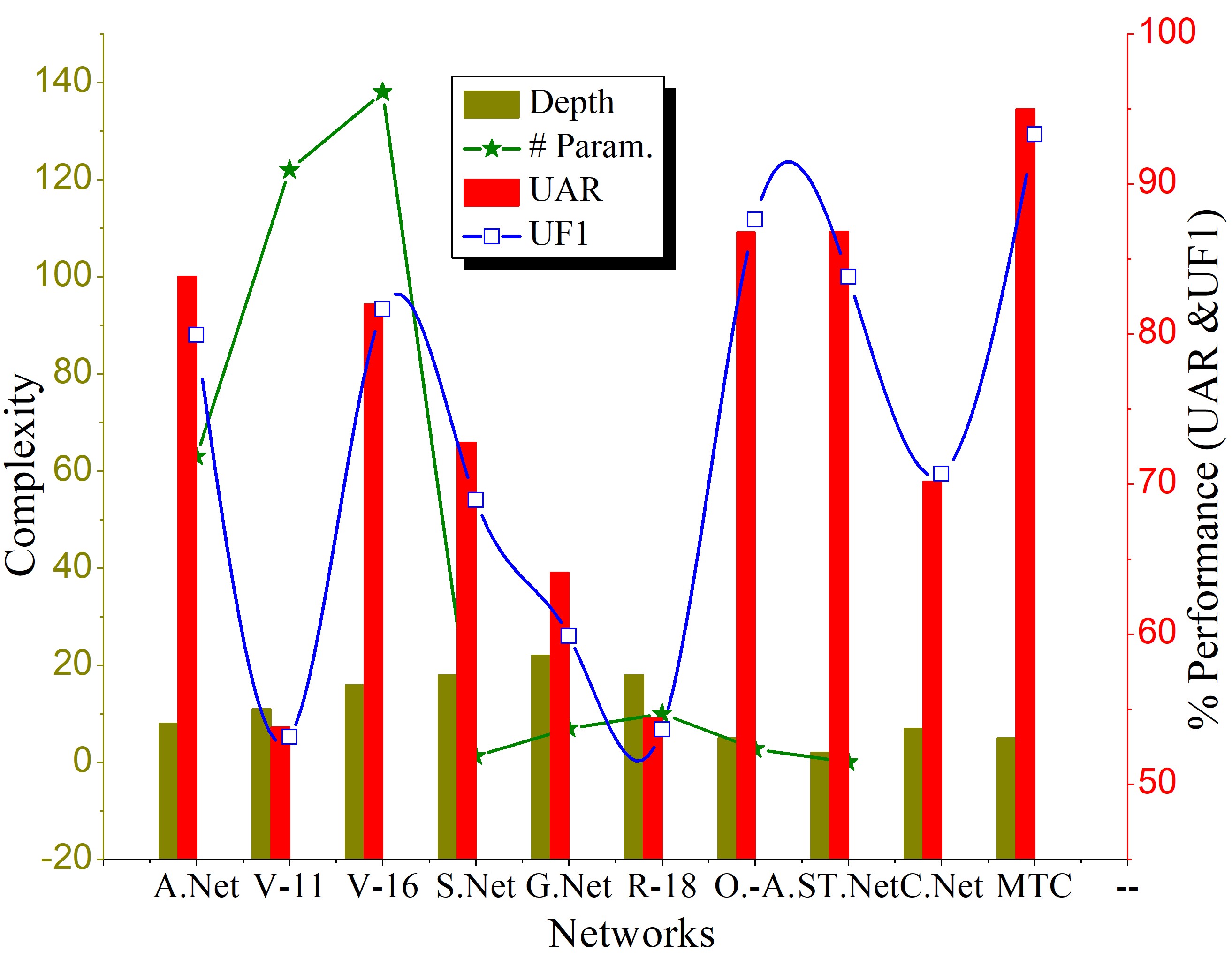}
    \caption{The performance analysis of various deep \cite{krizhevsky2012imagenet, Simonyan2014VeryDC,iandola2016squeezenet, Giannopoulos2018DeepLA, He2015DeepRL } and shallow \cite{gan2019off, ststnet, van2019capsulenet, gupta2021merastc} networks in terms of depth, UAR, UF1 and total number of parameters {over CASME-II dataset with 3 expressions (positive, negative and surprise) LOSO evaluation setup}. \textit{Here, A. Net, V-11, V-16, S.Net, G. Net, R-18, O.\_A, ST.Net, C.Net and MTC implies for AlexNet, VGG-11, VGG-16, SqueezeNet, GoogleNet, ResNet-18, OffApex, STSTNet, capsuleNet and MERASTC.}}
    \label{fig:12}
\end{figure}

\subsubsection{Impact of Multi-Scale/-Stream and Sequential CNN Frameworks on MER}
Multi-scale/stream feature representations have been successfully used in MER and achieve good performance in the literature \cite{zhou2019dual, song2019recognizing, yang2019merta}. From the literature \cite{zhou2019dual, song2019recognizing, yang2019merta}, it is evident that linearly coupled conv layers with identical filter size (VGG-16, VGG-19, ResNet) have failed to capture homogeneous scaled receptive fields and avoid fine-tuned edge variations. However, multi-scale/stream CNN models could capture detailed features from small to extensive regions, by applying multi-conv layers with different scale filters. The qualitative comparison between single-scale/stream and multi-scale/stream conv layers are depicted in Fig. \ref{fig:10}. From Fig. \ref{fig:10a}, it is quite clear that CNN based models built on single branch linearly connected conv layers, lack in gathering adequate features of facial appearance due to repetitive cross-correlation operation. Whereas, Fig. \ref{fig:10b} represents the capability of multi scale/stream conv layers in learning of significant discriminable features from the expressive regions of the MEs. Moreover, the quantitative results for models \cite{verma2019learnet, verma2020non, li2020joint, song2019recognizing} on different datasets: CASME-I, CASME-II, CAS(ME)$^2$, SAMM and SMIC, are represented in Fig. \ref{fig:11}. The quantitative results have also proven a higher generalization capabilities of the mutli-scale/stream over single-scale/linear CNN frameworks. Based on both qualitative and quantitative results analysis, we can conclude that multi-scale/stream CNNs acquired more MEs features and outperformed the single stream/linear CNN frameworks. The existing multiscale/ stream models are detailed in Table \ref{tab:3} and corresponding results for different datasets over single-domain leave one subject out, composite-domain leave one subject out and cross-domain/dataset validation protocols are tabulated in Table \ref{tab:4}-\ref{tab:7}.

\subsubsection{Does deeper network affect the performance of the MER?}
Yes, in general, the CNN model requires a large amount of data samples for efficient training. However, publicly available datasets for MER consist of limited data samples and tend to cause over-fitting. Moreover, deep/dense networks like AlexNet, VGG-11, VGG-16, SqueezeNet, GoogleNet and ResNet-18, etc also failed to capture minute features but were liable in emotion classification. Deep/dense networks \cite{Verma2019HiNetHI, ststnet} may vanish the micro-level features of the expressive regions due to progressive convolution and pooling operation. Therefore, most state-of-the-art MER approaches \cite{gan2019off, ststnet, van2019capsulenet, gupta2021merastc}  adopt shallow and light weighted CNN models for MER. The quantitative results of literature study are analyzed in Fig. \ref{fig:12}. In Fig. \ref{fig:12}, we can observe that, deep/dense networks: AlexNet, VGG-11, VGG-16, SqueezeNet, GoogleNet and ResNet-18 have failed to achieve superior performance as compared to shallower networks: OffApex, STSTNet, CLF, MTC. Based on results, we can conclude that shallow networks are preferable and achieve high performance in terms of accuracy as well as F1-score as compared to deeper networks. Moreover, deeper networks are computationally expensive as compared to shallow networks.

\subsubsection{Does kernel sizes have any impact on CNN layers?}
\begin{figure}[!t]
    \centering
    \includegraphics[width=0.8\linewidth, height=1.8in]{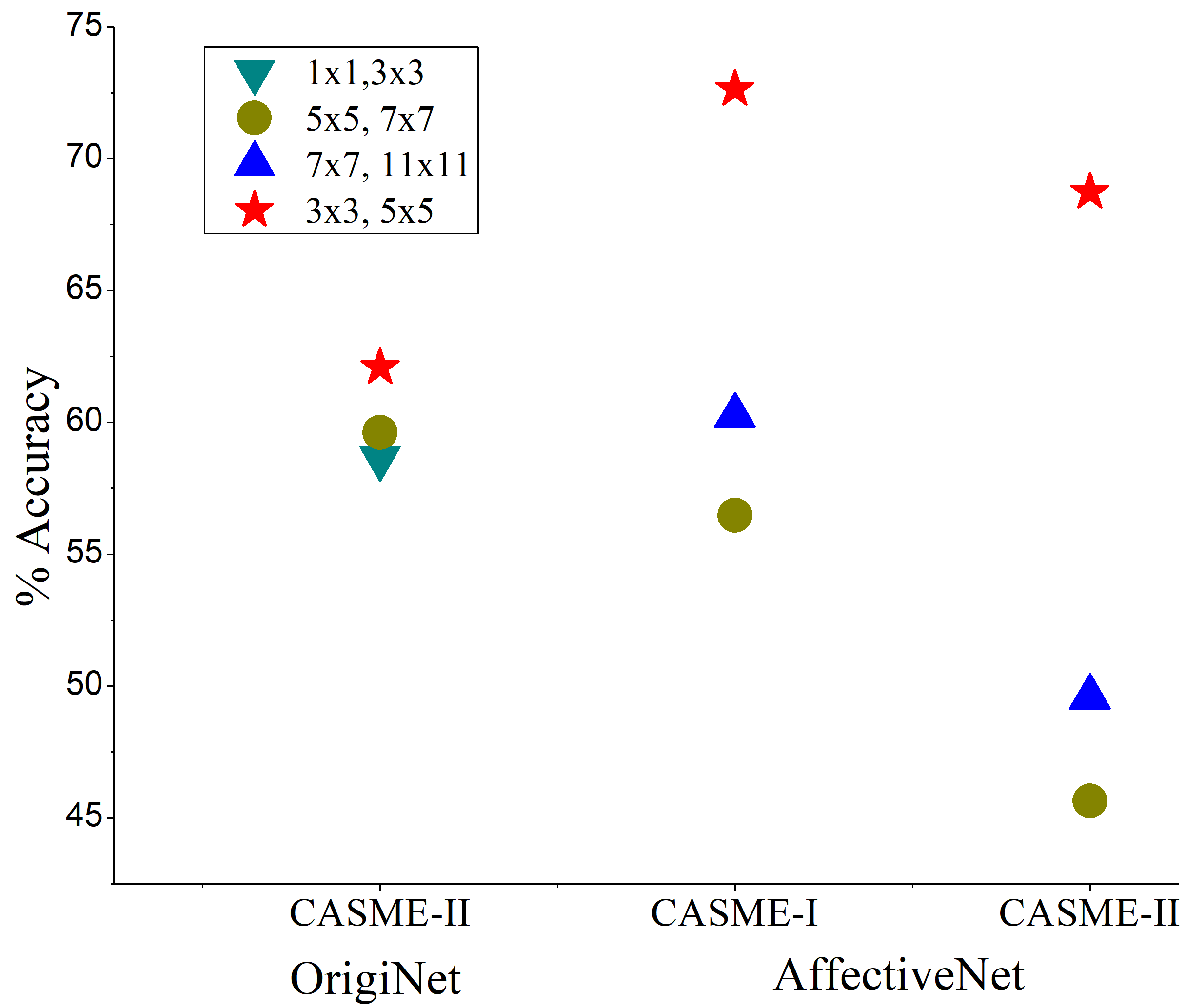}
    \caption{{The accuracy performance analysis for different kernel: hybrid ($1\times 1$ and $3\times 3$), hybrid ($5\times 5$ and $7\times 7$), hybrid ($7\times 7$ and $11\times 11$), and hybrid ($3\times 3$ and $5\times 5$),  sizes on CASME-I and CASME-II datasets over SD-LOSO setup.}}
    \label{fig:14}
\end{figure}

Yes, the kernel sizes also come under important paradigms of the CNN model designing. Kernel sizes directly affect the performance of the model along with computation cost. The comparative analysis between different sized kernels on datasets: CASME-I and CASME-II for MER frameworks \cite{vermaAff, verma2020non}  are demonstrated in Fig. \ref{fig:14}. Moreover, the qualitative effect of various kernel sizes is depicted in Fig. \ref{fig:10}. Based on literature study \cite{verma2020non}, it is clear that kernel sizes $3\times 3$ and $5\times 5$ are more capable to define the MEs features and achieve higher performance in MER. From the observations the smaller kernel sizes are preferable for MER applications. Kernels with large scales ($7\times 7$ and $11\times 11$) have a larger receptive field per layer and allow the extraction of generic features spread across the image. Therefore, these filters focus on abstract transitional information and skip the minute information, which is quite important in MER.
\section{Training and Evaluation strategies}\label{sec:exp}
The performance of any framework is affected by two factors: 1) the model uncertainty in architecture design 2) the evaluation strategy, i. e. the strictness in the data division to validate the generalization strength of the framework. The technical aspects of designing paradigms are already discussed in the previous section (learning based MER models and studying technical characteristics). In this section, first we focus on the datasets and evaluation metrics (section \ref{sec:eval}). Further, we discussed the validation strategies available in the literature to validate the robustness of the MER frameworks (section \ref{setups}). Moreover, we have studied the different paradigms of experimental strategies in detail and observations are provided in section \ref{exstr}.
\subsection{Datasets and Evaluation Metrics}\label{sec:eval}
The current research needs across all the computer vision applications are motivated by the success of deep learning algorithms. The success of the deep learning algorithms highly depends on the availability of sufficient training data with variations of the populations and environments as much as possible. The higher the diversity in the present training data, the more robustly one can estimate the model parameters. In this section, we primarily discuss the publicly available MEs datasets (highlighted by purple color in the Fig. \ref{fig:1}) that have been used for evaluating the MER methods. In the literature, ME datasets can be broadly classified into a dataset in-lab environment and a dataset in-wild.

\subsubsection{Traditional Datasets (Lab environment)} 
The six traditional datasets: CASME-I, CASME-II, CAS(ME)$^2$, SMIC, SAMM and MMEW have been widely used in the literature for MER evaluation. Among them, the CASME-II dataset is most prominently used in literature. The CASME-II dataset holds two sets: Part 1-247 and Part 2-255 with 5 and 8 emotion classes, respectively. The CASME-I elicited 195 image sequences of 35 participants (22 males and 13 females) with 8 emotions. Moreover, CAS(ME)$^2$ dataset is prepared to spot and recognize MEs in long videos. The CAS(ME)$^2$ dataset contains two parts: part A and part B. Part A includes 87 long video sequences of macro and micro expressions which are used for MEs spotting tasks. While part B included 300 macro and 57 micro expressions, which are used to evaluate MER tasks. The CAS(ME)$^2$ dataset samples were collected from 22 participants (9 males and 13 females). All the samples of CAS(ME)$^2$ dataset are annotated by using AUs and 4 emotion labels: Positive (8), Negative (21), Surprise (9), and Others (19). Moreover, to provide more challenging scenarios, SMIC has been introduced with three different illumination conditions: high speed (HS), visual (VIS) and near infrared (NIR) including 164, 71 and 71 samples of 16, 8 and 8 subjects, respectively. The SMIC dataset is the second most utilized dataset to validate the performance of MER models. All the above datasets involve either one or three ethnic participants. Therefore, to include the challenge of diverse ethnicity SAMM dataset has been furnished with 32 participants having 13 different ethnicities. The SAMM dataset comprises 159 video samples annotated with 7 emotion classes: Happiness (24), Surprise (13), Anger (20), Disgust (8), Sadness (3), Fear (7), and Others (84). Recently, a micro-and-macro expression warehouse (MMEW) \cite{ben2021video} dataset with the largest pool of MEs was introduced. The MMEW dataset contains 300 macro and micro image sequences of 36 participants. The MMEW dataset is annotated with FACS and 7 emotion classes: Happiness (36), Anger (8), Surprise (89), Disgust (72), Fear (16), Sadness (13), and Others (102).

\begin{figure}[!t]
    \centering
    \includegraphics[width=0.78\linewidth, height=3.8in]{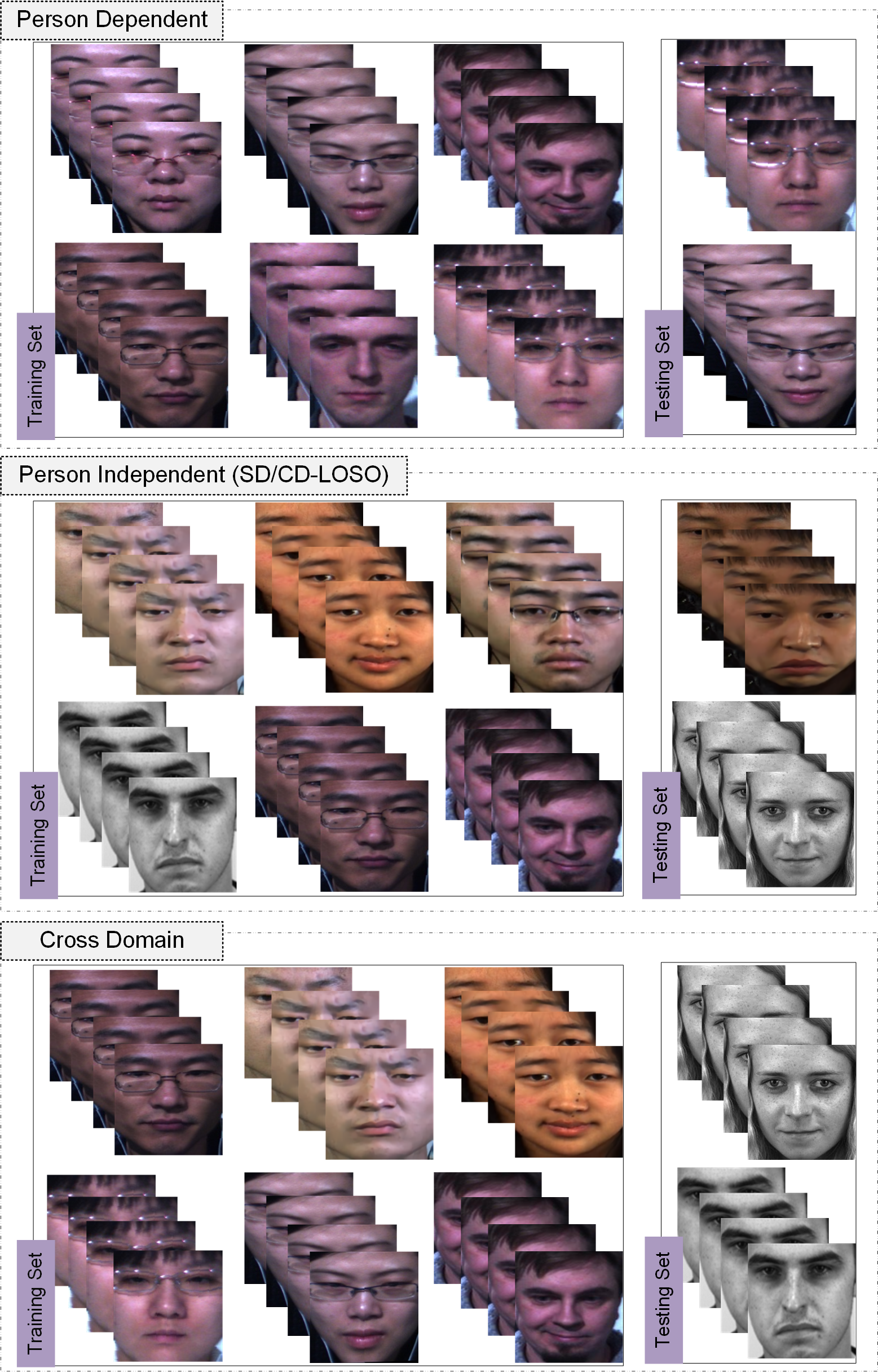}
    \caption{Differences between person dependent, person independent and cross domain validation setups.}
    \label{fig:15}
\end{figure}
\subsubsection{Wild Dataset}
All traditional datasets samples are elicited in a lab-controlled environment and lack the detail and divinity of real-life challenges. The MEVIEW dataset \cite{husak2017spotting} is the first MEs dataset that incorporated wild environment challenges: occlusion, illumination variations, candid faces \textit{etc.} The video samples are collected from the website. Specifically, all the samples are downloaded from YouTube videos of poker games. The database contains 31 samples from 16 subjects, having both macro and micro expressions. More detailed analysis of datasets can be found in \cite{husak2017spotting}.

\subsection{Evaluation Metrics and significance}
The widely used evaluation metrics for MER are recognition accuracy (Acc.), weighted f1- score (F1), weighted average recall (WAR), Un-weighted f1-score (UF1), Unweighted average recall (UAR), and mean diagonal value of the confusion matrix. The accuracy is calculated by computing average hit rate across the all emotion class samples. Let $TP$, $TN$, $FP$ and $FN$ be the true positive, true negative, false positive and false negative, respectively. The recognition accuracy is computed by using Eq. [\ref{eq:1}].
\begin{equation}\label{eq:1}
    Recognition \;\;accuracy=\frac{TP+TN}{TP+TN+FP+FN}
\end{equation}
The accuracy can be largely contributed by a large number of true negatives and not focus on false negative and false positive. Thus, Acc. is liable to bias data and reflect partial effectiveness of the MER frameworks. Whereas, F1 score is a better measure to balance between $TP$, $TN$, $FP$ and $FN$. The F1-score is calculated by using Eq. [\ref{eq:2}].
\begin{equation}\label{eq:2}
    F_{1}\;Score=\sum_{i=1}^{c}\frac{T_{i}}{T}\times \frac{2\times TP_{i}}{TP_{i}+FP_{i}+FN_{i}}
\end{equation}
Where, T and C represents the total number of samples and emotion classes, respectively. F1 Score is considered as a better measure because of its balancing nature with uneven class distribution (large number of actual negatives).\par

Though MER datasets have heavy imbalanced annotations, both Acc and F1 score failed to justify the efficacy of the MER Models. Recently, UAR and UF1 have drawn much attention due to their unbiased nature of evaluation. Both UAR and UF1 computed the performance of a model w.r.t number of classes without consideration of samples per class. The UAR and UF1 are calculated by using Eq. [\ref{eq:3}] and Eq. [\ref{eq:4}], respectively.\par
\begin{equation}\label{eq:3}
    UAR=\frac{1}{C}\sum^{C}_{i=1}\frac{TP_{i}}{T_{i}}
\end{equation}
\begin{equation}\label{eq:4}
    UF1=\frac{1}{C}\sum^{C}_{i=1}\frac{2\times TP_{i}}{\left(2\times TP_{i}\right)+FP_{i}+FN_{i}}
\end{equation}

Moreover, some of the methods \cite{peng2018macro} utilized the WAR for evaluation. The WAR is computed by using Eq. [\ref{eq:5}].\par
\begin{equation}\label{eq:5}
    WAR=\frac{\sum^{C}_{i=1}{TP_{i}}}{T}
\end{equation}

Some of the MER frameworks \cite{li2018can,van2019capsulenet,choi2020facial, li2020joint} also adopt the mean diagonal value of the confusion matrix to show the detailed generalization of the model.
\begin{figure}[!t]
    \centering
    \includegraphics[width=0.7\linewidth, height=1.45in]{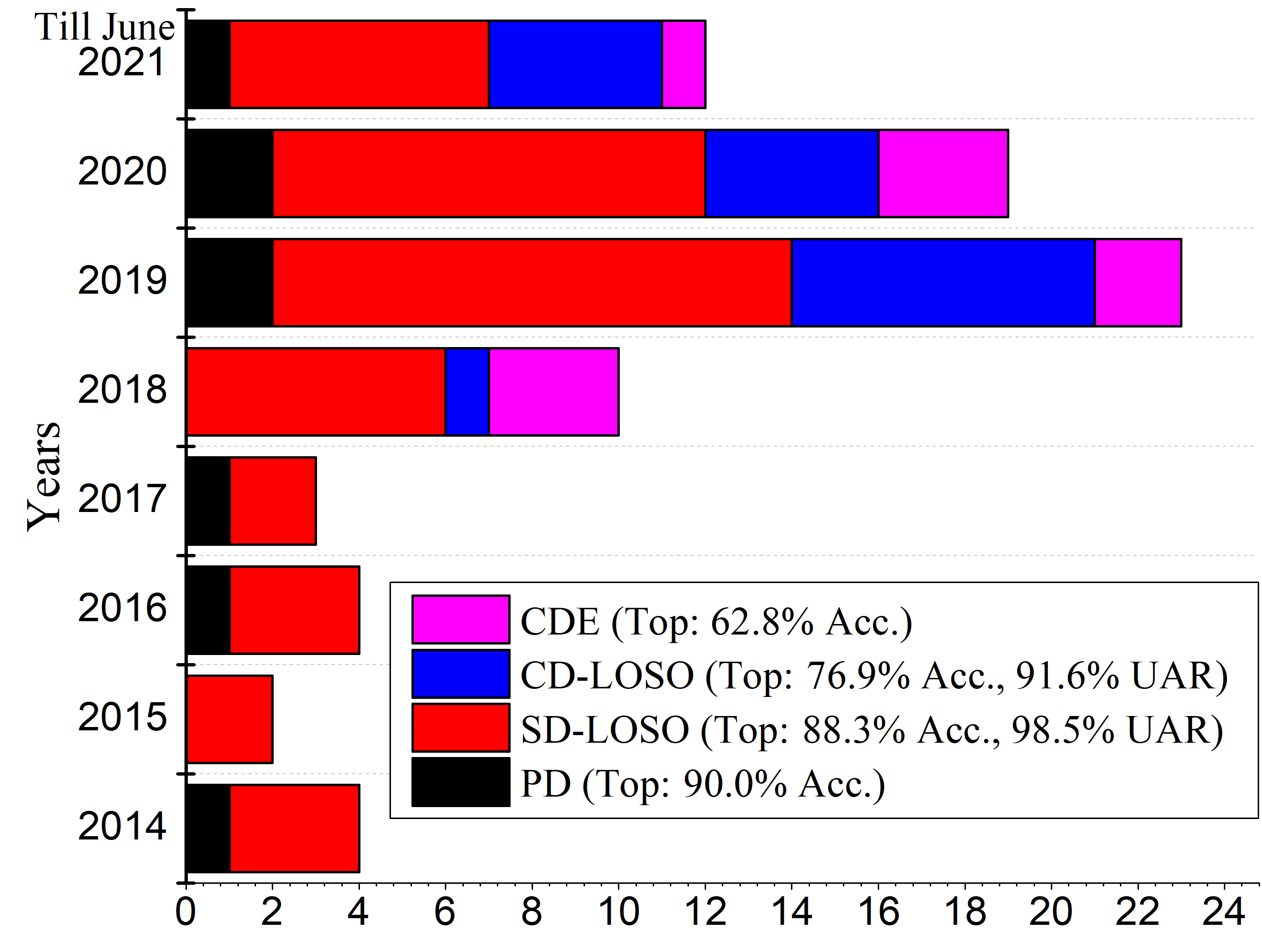}
    \caption{The evolution in validation strategies in past years.}
    \label{fig:16}
\end{figure}
\subsection{Validation Strategies/protocols}\label{setups}
In literature, many state-of-that-art techniques adopted diverse data-division strategies to prove the robustness of the algorithm. Therefore, the supervised techniques we broadly categorize into person dependent evaluation (PDE), person independent evaluation (PIE) and cross-data evaluation (CDE). In ‘Person Dependent setup’ the training and testing set contains frames from the same category, whereas in Person independent setup’ completely unseen data is used for testing. Similarly, in cross-data evaluation train the model with one dataset and test the model accuracy with a completely different dataset.
The sample technical differences between PDE, PIE and CDE are depicted in Fig. \ref{fig:15}. Also, the impact of data-division strategies over the model’s performances are tabulated in the Table \ref{tab:3}- Table \ref{tab:7}. More details of the data-division strategies are discussed as follows. 

\subsubsection{Person Dependent Evaluation (PDE)}
Based on literature study, PDE setup can be further divided into two categories 1) k-fold cross validation and 2) leave one video out (LOVO). In k-fold cross validation data is randomly divided into a ratio of 100-P:P. Where, in each iteration P\% of data is used for inference to validate the performance of the model and remaining data samples are reserved for the training. Some studies \cite{verma2019learnet, reddy2019spontaneous} follow the PDE setup with 80:20 ratio. Recently, Thuseethan \etal \cite{thuseethan2020complex} used a 90:10 ratio with 10-fold cross-validation to validate the effectiveness of the framework. In LOVO, one expression video of a person is used for inference and remaining all data samples are used for training. Thus, there are immense chances to use the same person's expression in both training and testing data. The articles \cite{liu2015main, happy2017fuzzy, xia2020revealing} in literature adopted LOVO setup to prove the robustness of their models.

 \begin{figure}[!t]
     \centering
     \begin{subfigure}[b]{0.49\linewidth}
         \centering
         \includegraphics[width=\textwidth]{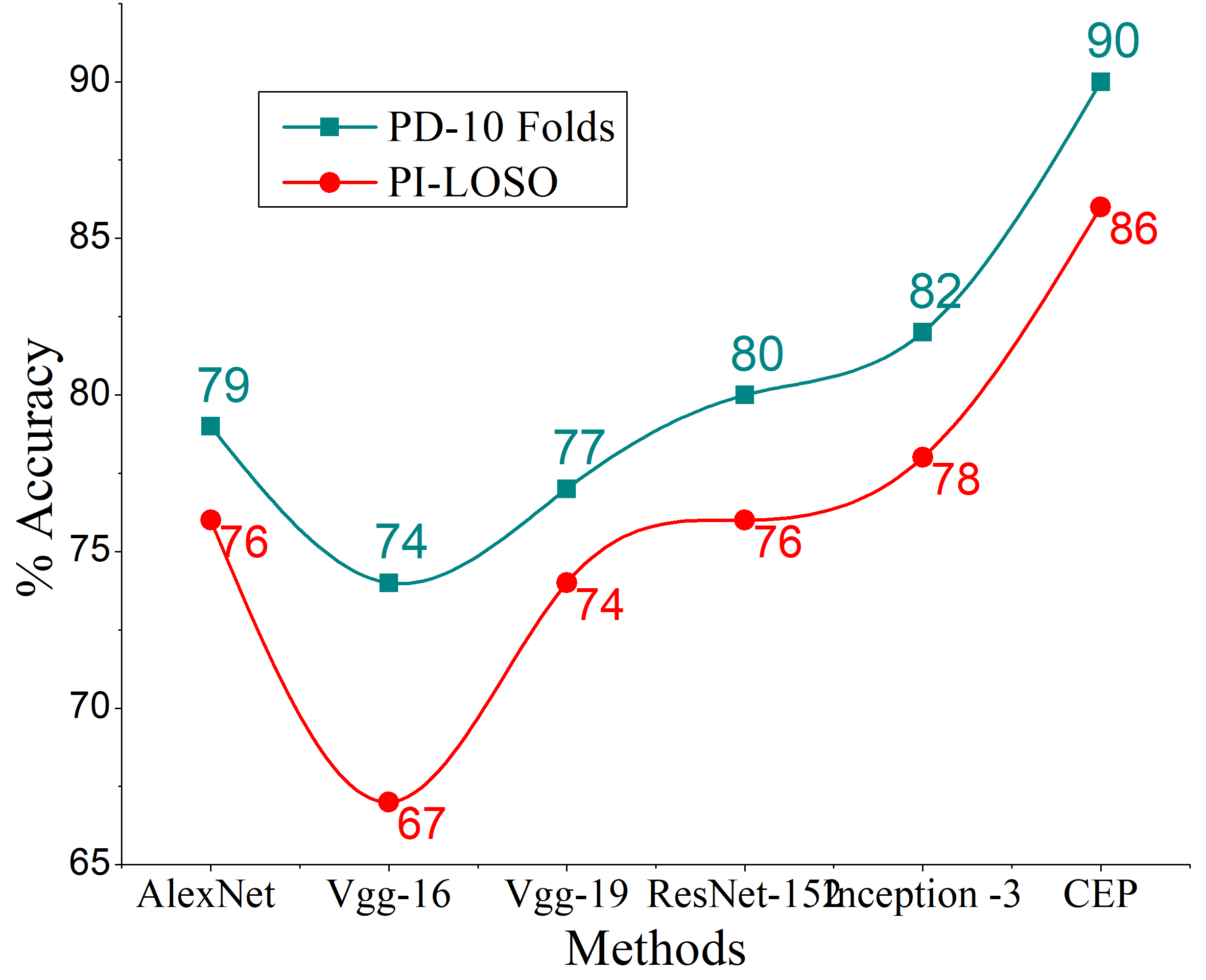}
         \caption{}
         \label{fig:17a}
     \end{subfigure}
     \hfill
     \begin{subfigure}[b]{0.49\linewidth}
         \centering
         \includegraphics[width=\textwidth]{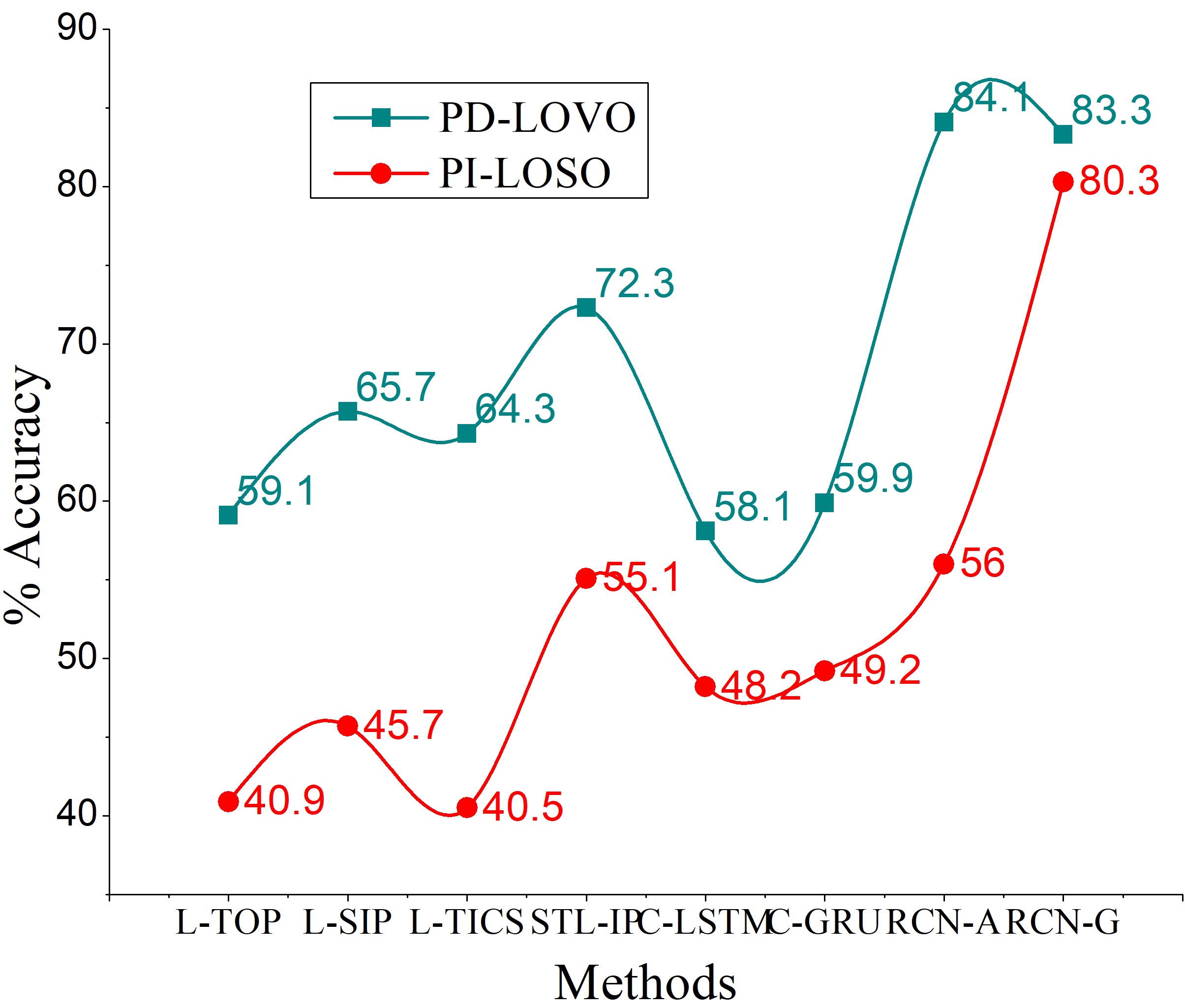}
         \caption{}
         \label{fig:17b}
     \end{subfigure}
     
        \caption{The performance analysis of different MER frameworks evaluated over a) PD-10 Folds vs PI-LOSO \cite{thuseethan2020complex} and b) PD-LOVO vs PI-LOSO\cite{xia2018spontaneous} setups.}
        \label{fig:17}
\end{figure}
\subsubsection{Person Independent Evaluation (PIE)}
The PIE setup follows a strict division for training and testing sets. It ensures evaluation over videos with unseen subject’s identities. Like PDE, the PIE setup can also be further categorized in single-domain leave-one-subject-out (SD-LOSO) and composite-domain leave-one-subject-out (CD-LOSO). In SD-LOSO, a single dataset is used, and samples are partitioned in such a way that all expressions of a particular subject at a particular iteration act as a testing set, and the remaining data is considered as the training set. The SD-LOSO is the most widely used validation strategy in the literature \cite{xia2019spatiotemporal, xia2020revealing, van2019capsulenet, xie2020assisted,  sun2020dynamic}. The list of models with SD-LOSO is tabulated in Table \ref{tab:3}. The CD-LOSO is one of the recently emerging validation setups to validate the robustness of the model with domain shifts. First MEGC 2018 \cite{yap2018facial} introduced the CD-LOSO setup for the MEs challenge. The CD-LOSO data division was introduced by combining CASME-II, SAMM, and SMIC with three emotion classes: positive, negative, and surprise. Further, the composite dataset is used by adopting an LOSO set up to evaluate the robustness and generalization of the model for MER with domain shifts. Based on literature study \cite{peng2018macro, khor2018enriched, van2019capsulenet, liu2019neural, zhou2019dual, xia2020learning, xia2020revealing} CD-LOSO is a highly recommended validation strategy as it ensures evaluation on unseen faces with ethnicity variations. It is observed from the literature, recent works on MER adopted the CD-LOSO along with SD-LOSO to validate the performance of the MER frameworks as shown in Fig. \ref{fig:16}.

\subsubsection{Cross Domain Evaluation (CDE)}
CDE is another setup that ensures PIE by training a model over a particular dataset and testing on a different dataset. The MEGC-2018 \cite{yap2018facial} used the CDE setup to evaluate the efficacy of the submitted MER frameworks. Peng \etal \cite{peng2018macro} and Khor \etal \cite{khor2018enriched} have successfully proven the performance of the models on the CDE setup of the MEGC-18 challenge. Wang \etal \cite{wang2020micro} performed two cross-domain experiments: CASME-II dataset is used for training and testing results are evaluated on SAMM dataset and vice versa. Choi \etal \cite{choi2020facial} have utilized the pair of CASME-II and SMIC dataset for CDE. First, CASME-II is considered for training, while SMIC is used for inference. Further, SMIC is used for training, and CASME-II is reserved for testing. Moreover, Verma \etal \cite{vermaAff}  have conducted nine experiments over four datasets: CASME-I, CASME-II, CAS(ME)$^2$, and SAMM. Three experiments were conducted by using CASME-I as a training dataset and CASME-II, CAS(ME)$^2$, SAMM as testing datasets, individually. Another three experiments were evaluated by using CASME-II as a training dataset and CASME-I, CAS(ME)$^2$, SAMM as testing datasets, individually. Similarly, the other three experiments were performed by using SAMM as a training dataset and the remaining three as testing datasets. Based on the literature, CDE setups are very less popular but have the most recommendable data validation strategies to gain much generalization capabilities of the FER models.

\begin{table*}[!t]
\centering
\caption{Experimental settings and performance based comparison of existing DL approaches on CD-LOSO validation setup.}
\label{tab:6}
\begin{tabular}{lcccccccccccc}
\hline
\multirow{2}{*}{\textbf{Pub-Yr}} & \multirow{2}{*}{\textbf{Input Size}} & \multirow{2}{*}{\textbf{LR}} & \multirow{2}{*}{\textbf{Data Aug.}}                                           & \multicolumn{3}{c}{\textbf{COMPOSITE}}         & \multicolumn{2}{c}{\textbf{SMIC}} & \multicolumn{2}{c}{\textbf{CASME-II}} & \multicolumn{2}{c}{\textbf{SAMM}} \\ \cline{5-13}
                                                                      &                              &                                                                                 &                                    & \textbf{Acc} & \textbf{UAR}   & \textbf{UF1}   & \textbf{UAR}    & \textbf{UF1}    & \textbf{UAR}      & UF1      & \textbf{UAR}    & \textbf{UF1}    \\ \hline \hline
FG-19 \cite{zhou2019dual}          & ${28\times 28}$                       & Fixed               & No                                                                                        & N/A          & 72.78          & 73.22          & 67.26           & 66.45           & 85.60             & 86.21             & 56.63           & 58.68           \\
FG-19 \cite{ststnet}         & $28\times 28$                       & Fixed               & No                                                                                      & 76.92        & 76.05          & 73.53          & 70.13           & 68.01           & 86.86             & 83.82             & 68.10           & 65.88           \\
FG-19 \cite{liu2019neural}         & {N/A}                         & {N/A}                 & N/A                                                                                      & N/A          & 78.24          & 78.85          & 75.30           & 74.61           & 82.09             & 82.93             & 71.52           & 77.54           \\
{FG-19 \cite{van2019capsulenet}}          & {Apex}                        & {Adaptive}            & F, R, T                  & N/A          & 65.06          & 65.20          & 58.77           & 58.20           & 70.18             & 70.68             & 59.89           & 62.09           \\
{ICBEA-19 \cite{xia2019cross}}       & $220\times 200$                     & {Adaptive}            & {N/A}                                                                                       & {N/A} & {60.37} & {59.79} & {63.27}  & {63.30}  & {57.97}    & {55.96}    & {53.03}  & {52.34}  \\
{MLSP-19 \cite{aouayeb2019spatiotemporal}}       & {N/A}                         &{N/A}                 & {N/A}                                                                                      & {N/A} & {90.18} & {90.22} & {88.28}  & {88.86}  & {98.57}    & {98.57}    & {81.03}  & {78.55}  \\
{TIP-20 \cite{xia2020revealing}}         & ${60\times 60}$                       & {Adaptive}            & {N/A}                                                                                     & {N/A} & {70.52} & {71.64} & {59.80}  & {59.91}  & {80.87}    & {85.63}    & {67.71}  & {69.76}  \\
{ACMMM-20 \cite{xia2020learning}}       &${224\times 224}$                     & {Fixed}               & {No}                                                                                        & {N/A} & {85.7}  & {86.4}  & {86.1}   & {86.4}   & {87.2}     & {87}       & {81.9}   & {82.5}   \\
{Arxiv-20 \cite{yu2020ice}}       & ${128\times 128}$                     &{Adaptive}            & {No}                                                                                     & {N/A} &{84.1}  &{84.5}  & {79.1}   &{79.0}   & {86.8}     & {87.6}     & {82.3}   & {84.5}   \\
{IEEE-Acc-20 \cite{choi2020facial}}    & ${21\times 21}$                       & {N/A}                 & {No}                                                                                        & {N/A} & 72.0  & {75.0}  &{71.0}   &{71.0}   & {77.0}     &{72.0}     & {51.0}   & {65.0}   \\

TAFF-21 \cite{gupta2021merastc}         & N/A                         & Adaptive            & F, R, T                     & N/A & 91.6  & 92.0  & 86.2   & 79.0   & 95.0 & 93.3     & 84.6   & 83.0  \\
Neu. Co-21 \cite{zhao2021two} & $112\times112$  & Adaptive            & Yes & N/A & 79.86  & 80.68   & 75.98  & 73.56 & 87.63     &88.18 & 72.80   & 74.75 \\ 
Sig. Proc. \cite{liu2021micro}         & $28\times 28$  & Fixed & No & 0.859 & 0.836&0.836&0.814& 0.812 &0.882& 0.891 &0.800& 0.790    \\ 
{PR-22 \cite{zhou2022feature}}       & ${28\times 28}$                       & {N/A}                 & N/A                                                                                               &{N/A} & 78.32 & {78.38} &{70.83}  & {70.11}  & {88.73}    & {89.15}    & 71.55  & 73.72  \\\hline
\end{tabular}

\end{table*}
\subsection{Discussion on experimental setups and validation strategies}\label{exstr}
In literature, we noticed that the standard evaluation protocols for MER techniques are not available as many authors follow contrast experimental setups to prove the robustness of the MER techniques. The term contrast is defined in terms of number of samples used for training and testing, input selection strategies, number of expression classes adopted in training or dropped some of emotion classes due to a smaller number of images \cite{xia2020revealing, vermaAff}, type of validation strategy adopted to prove the robustness, \textit{etc.} Based on the above contrast settings, it is harder to compare the performance of these techniques directly. Therefore, this section aims to provide a detailed study on experimental setups adopted in the literature. Also, analyzed the effect of adopting different evaluation setups like, PDE, PIE and CDE, data augmentation, input selection, number of emotion classes etc, over the performance of the MER approaches.

\subsubsection{Impact of PDE, PIE and CDE on MER performance}
In person dependent evaluation, the  training and testing set contains frames from the same category.  Thus, there is a possibility of similar samples of the same subject present on both training and testing data sets, which leads to inflated performance during testing. However, it may fail in real world scenarios. Thus, there is a need to evaluate the model performance over unseen or person-independent scenarios. This also makes the process of model design much more challenging to ensure robust performance even in real world scenarios. Therefore, the PIE ensures stable performance over PDE on unseen data. \par
In literature many existing works \cite{peng2018macro, khor2018enriched, van2019capsulenet, liu2019neural, zhou2019dual, xia2020learning, xia2020revealing, verma2021automer} have opted the PIE setup’s evaluation (More detailed categorization is indexed in Table \ref{tab:3}). Further, to study the effect of PDE as compared to PIE setup, we accumulated some results for PDE (including both 10-folds and LOVO) and PIE (SD-LOSO) setups as shown in Fig.\ref{fig:17}.  More specifically, Fig. \ref{fig:17a} illustrated the results analysis for the models \cite{thuseethan2020complex}, evaluated over 10-fold and SD-LOSO validation strategies. Whereas, Fig.  \ref{fig:17b} shows the comparative results for model \cite{xia2018spontaneous} evaluated over LOVO and SD-LOSO validation strategies. Based on Fig. \ref{fig:17}, it is evident that the models that adopted PDE setups outperform in accuracy over the PDE setup. 
Thereby, the PDE results are unreliable to validate the actual robustness of the deep learning models. Similarly, the model’s performance depends on a person’s identity, the cross culture and ethnicity variations. \par
Therefore, recent works \cite{ststnet, yu2020ice, gupta2021merastc, liu2021micro} also focused on the CD-LOSO validation strategy as it ensures person independence with sparse diversity in domains. However, it is hard to achieve impressive performance with the strictness of CD-LOSO. As we can see in Fig. \ref{fig:16}, the best accuracy results for CD-LOSO are 76.9\%, which has enough margin as compared to the results of PDE and SD-LOSO. Furthermore, some work \cite{peng2018macro,wang2020micro, vermaAff, khor2018enriched} also adopted the CDE validation strategy to evaluate the models with more challenging scenarios such as cross-ethnicities, out-group, illumination, and resolution variations. Therefore, the CDE validation protocol evaluates the model’s robustness more towards real-world scenarios. From Fig. \ref{fig:16}, the best accuracy over CDE setup is 62.8\%, which is very less as compared to PDE as well as PIE validation setups. Therefore, benchmarking the performances of different CNN, GAN models in a standard evaluation setup (PDE/PIE/CDE) is an important scope in micro expression research.

\begin{figure}[!bp]
     \centering
     \begin{subfigure}[b]{0.4\textwidth}
         \centering
         \includegraphics[width=\textwidth, height=1.9in]{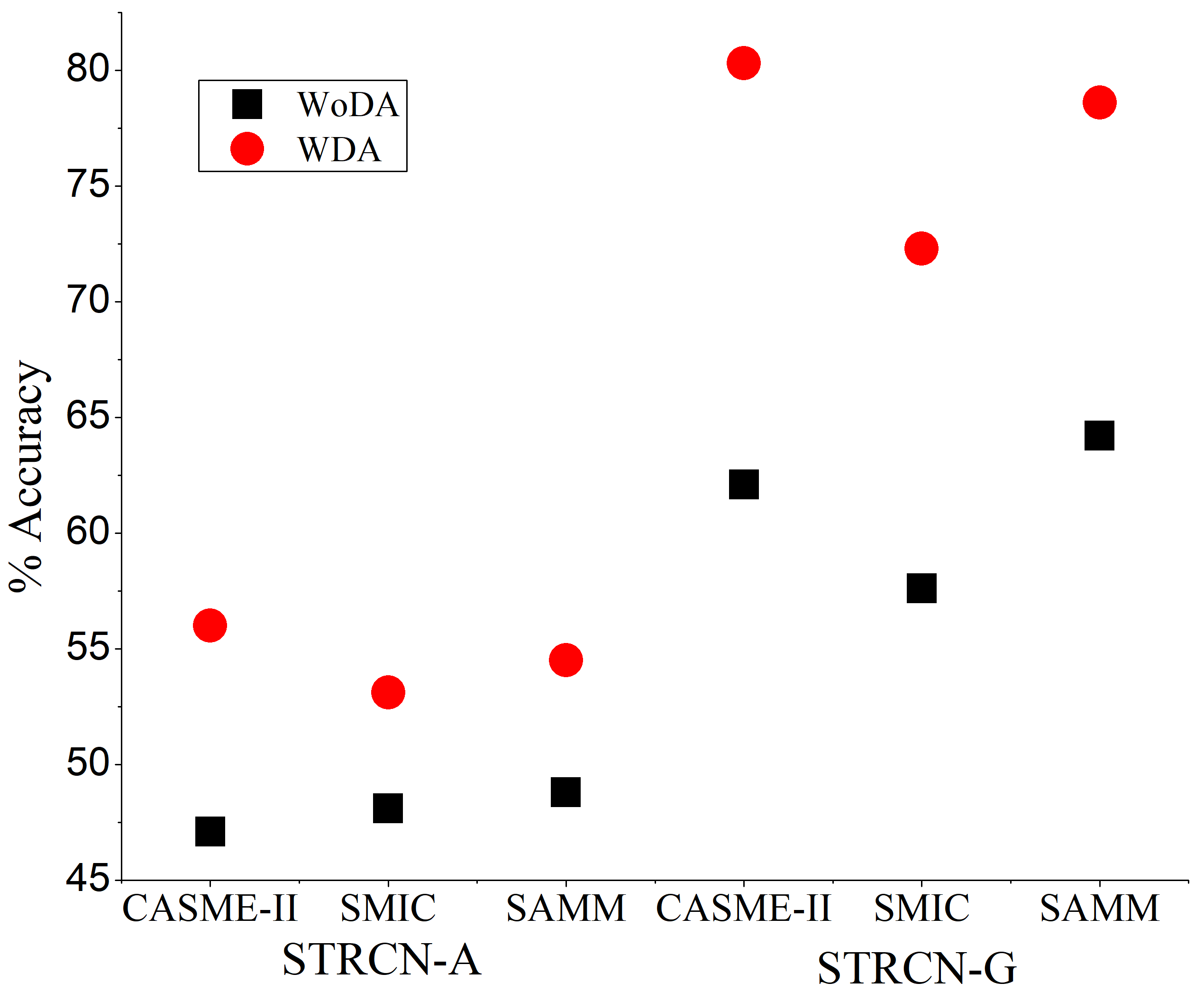}
         \caption{}
         \label{fig:y equals x}
     \end{subfigure}
     \smallskip
     \begin{subfigure}[b]{0.4\textwidth}
         \centering
         \includegraphics[width=\textwidth, height=1.9in]{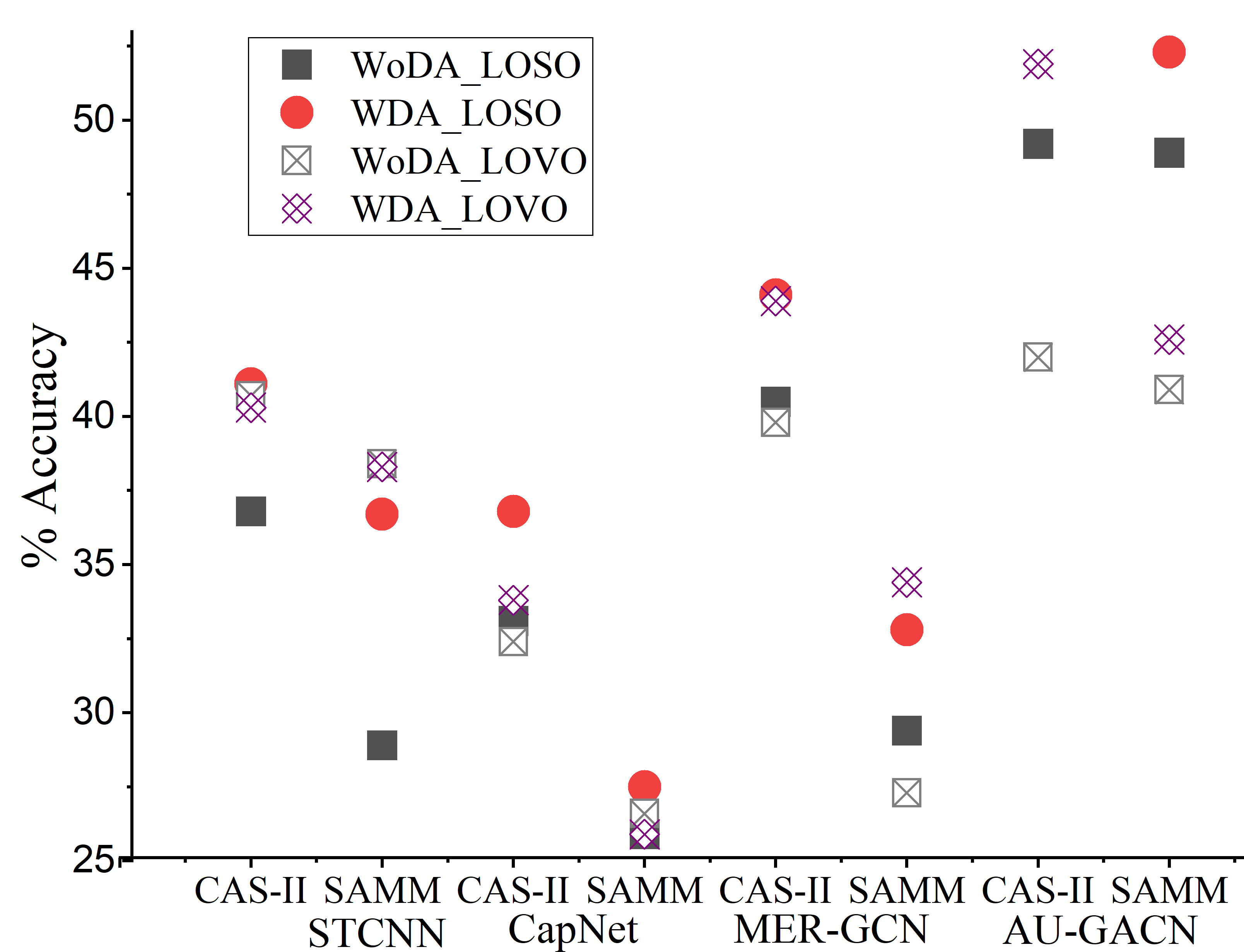}
         \caption{}
         \label{fig:three sin x}
     \end{subfigure}
    
        \caption{The performance of the (a) STRCN-A and STRCNG \cite{xia2019spatiotemporal}  models with or without data augmentation (based on temporal connectivity), over CASME-II, SMIC, and SAMM datasets, and (b) STCNN, CapsuleNet, MER-GCN, and AU-GACN \cite{xie2020assisted} models with or without data augmentation (based on GAN) over CASME-II, and SAMM datasets, in terms of recognition accuracy. \textit{\small Here, WDA, WoDA, LOSO, LOVO, CAS-II, and CapNet represent the with data augmentation, without data augmentation, leave-one-subject-out, leave-one-video-out, CASME-II, and CapsuleNet.}}
        \label{fig:18}
\end{figure}

\subsubsection{Impact of Data Augmentation on MER performance}
Since MEs datasets consist of a limited number of samples and imbalanced classes, which leads to model overfit. Many existing MER approaches \cite{kim2016micro, peng2018macro, hashmi2021larnet, van2019capsulenet, verma2020non, thuseethan2020complex, wang2018micro, lei2020novel,vermaAff, xia2019spatiotemporal, li2020joint} adopt the data augmentation techniques to create a sufficient pool of data samples for training. 2D-CNN based approaches \cite{kim2016micro, peng2018macro, hashmi2021larnet, van2019capsulenet, verma2020non, thuseethan2020complex, wang2018micro, lei2020novel, vermaAff} performed the basic operations like flipping, rotating, color shift, smoothing to increase the data samples. Xia \etal \cite{xia2019spatiotemporal} introduced the two new augmentation techniques with temporal connectivity. While the recent work \cite{xie2020assisted, yu2020ice} utilized the GAN based model to generate the synthetic MEs data for data augmentation. {In literature \cite{xie2020assisted, yu2020ice}, the patchGAN network is used to generate synthetic images/samples using the apex frame. The quality of the synthetic (fake/augmented) images are validated by using i.e., adversarial loss, consistency loss, attention loss, AU intensity loss, SSIM loss, ME loss and sequence authenticity loss. Similarly, \cite{xie2020assisted} utilizes a 3DConvNet architecture to distinguish the synthetic sequences from the real ones. Li \etal \cite{li2020joint} collect the five nearest frames to the apex frame to enhance the size of datasets. From literature \cite{kim2016micro, peng2018macro, hashmi2021larnet, van2019capsulenet, verma2020non, thuseethan2020complex, wang2018micro, lei2020novel,vermaAff, xia2019spatiotemporal, li2020joint, xie2020assisted}, we observed that data-augmentation resolves the issue of overfitting up to some extent and improves the performance of the model as shown in Fig. \ref{fig:18}(a) and Fig. \ref{fig:18}(b). Fig. \ref{fig:18}(a) depicted the results of STRCN-A and STRCN-G models over CASME-II, SMIC, and SAMM datasets with and without data augmentation (based on temporal connectivity) in terms of recognition accuracy. Fig. (b) represents the results of four other MER models: STCNN, CapsuleNet, MER-GCN, and AU-GACN models over CASME-II and SAMM datasets with and without data augmentation (based on GANs) in terms of recognition accuracy. From the analysis of the results, it is clear that the models trained over augmented datasets achieved higher accuracy as compared to models without augmented datasets.} However, some of the MER approaches \cite{khor2018enriched, Li3D-Flow, khor2019dual, ststnet, gan2019off} significantly improves the models performance without augmentation by introducing optical flow and shallow network designing. Therefore, it is observed that there is scope to design robust models without increasing the sample size.

\subsubsection{Impact of Input Selection Strategies on MER performance}
\begin{figure}[!t]
    \centering
    \includegraphics[width=0.8\linewidth, height=1.8in]{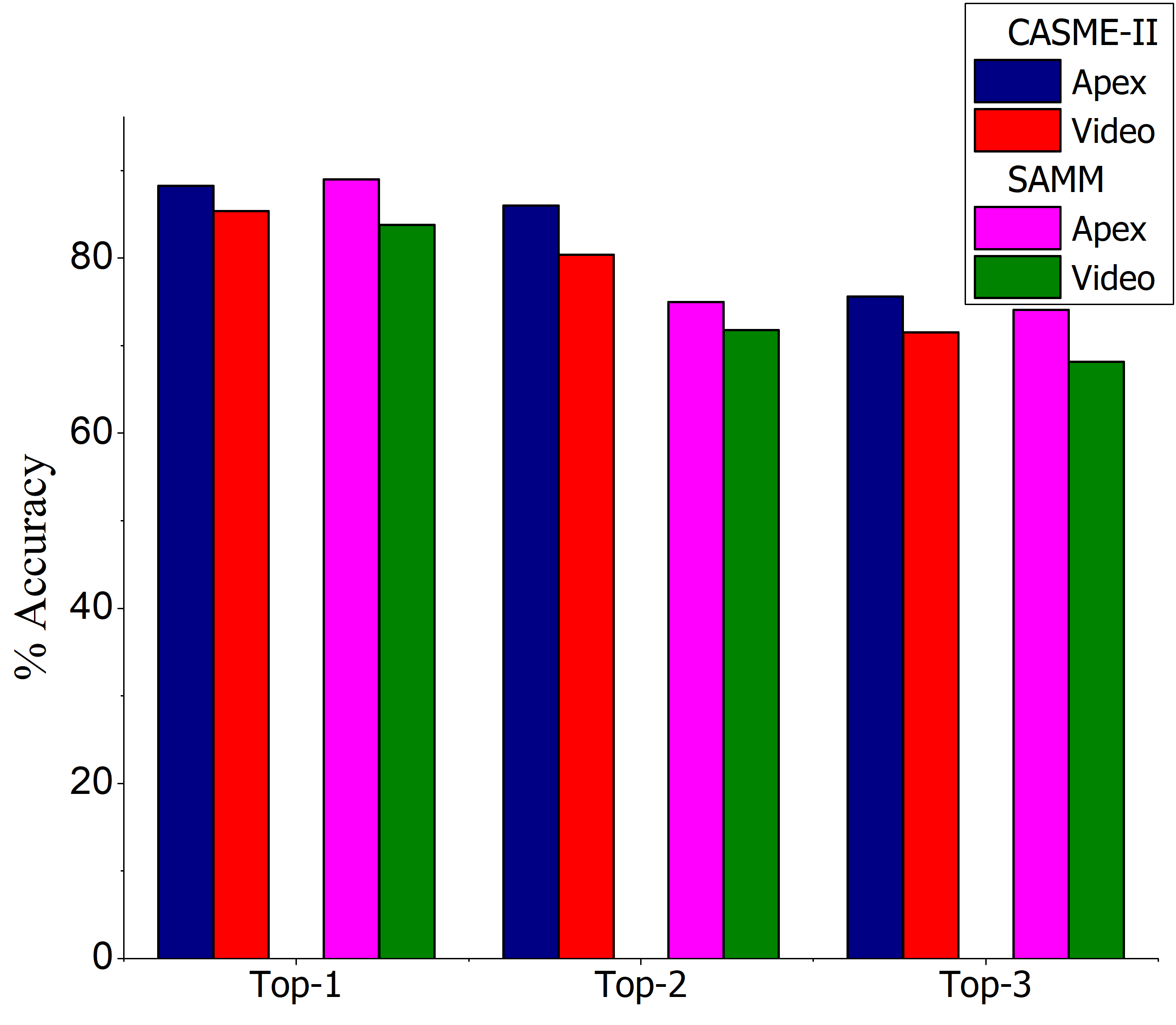}
    \caption{The {recognition rate} with two input selection strategies on two datasets: CASME-II and SAMM. \textit{Here, best three (top) \cite{gan2019off, xia2020learning, thuseethan2020complex, song2019recognizing, choi2020facial,zhou2022feature, gupta2021merastc, verma2021automer} accuracy results are selected for the comparative analysis.}}
    \label{fig:19}
\end{figure}

Most of the existing MER approaches follow three types of input formats: apex frame \cite{ststnet, lei2020novel, van2019capsulenet}, onset-apex-offset frames \cite{yu2020ice, sun2020dynamic} and whole videos \cite{Li3D-Flow, reddy2019spontaneous} as shown in Fig. \ref{fig:2}. Majority of the existing MER approaches \cite{li2020joint, liu2019neural, ststnet} rely only on the apex frame for the analysis. However, some studies emphasize the importance of dynamic aspects for detecting the subtle changes \cite{ambadar2005deciphering} and its effect on the performance of MER. In a MEs video, each frame has its own significance towards the identification of the emotion class. Therefore, apex frame-based approaches are lacking to analyze the motion information, which has its own potential to describe the MEs classes. Therefore, whole video input is more effective and reliable in MER. Nevertheless, the whole video input is incredibly challenging to handle and achieves less results as compared to apex frames input as shown in Fig. \ref{fig:19}. In Fig. \ref{fig:19}, we included top three accuracy results achieved by different models over CASME-II and SAMM datasets with apex and whole video inputs, respectively. From Fig. \ref{fig:19}, it is quite clear that models with whole video input formats acquire less accuracy results as compared to apex frames but have more reliability to the capability of delivering appearance information along with time variants. Therefore, utilizing complete frames in a video is more effective and reliable in MER than using a single apex frame.

\subsubsection{Do the number of emotion classes affect the MER performance?}
The number of emotion classes play a key role in the estimation of the MER's performance. In literature, there is no standard for the emotion class settings. Therefore,  it is difficult to compare the performance of the MER models directly. In literature, work has been done using 03 emotion classes (P, N, S), 04 emotion classes (P, N, S, O), 05 emotion classes (H, D, S, R, T) and 7 or 8 emotion classes for training the MER model. The list of several emotion classes adopted to test the MER model performance is Tabulated in Table \ref{tab:3}-\ref{tab:4}. \par
\begin{figure}[!t]
    \centering
    \includegraphics[width=0.68\linewidth, height=1.4in]{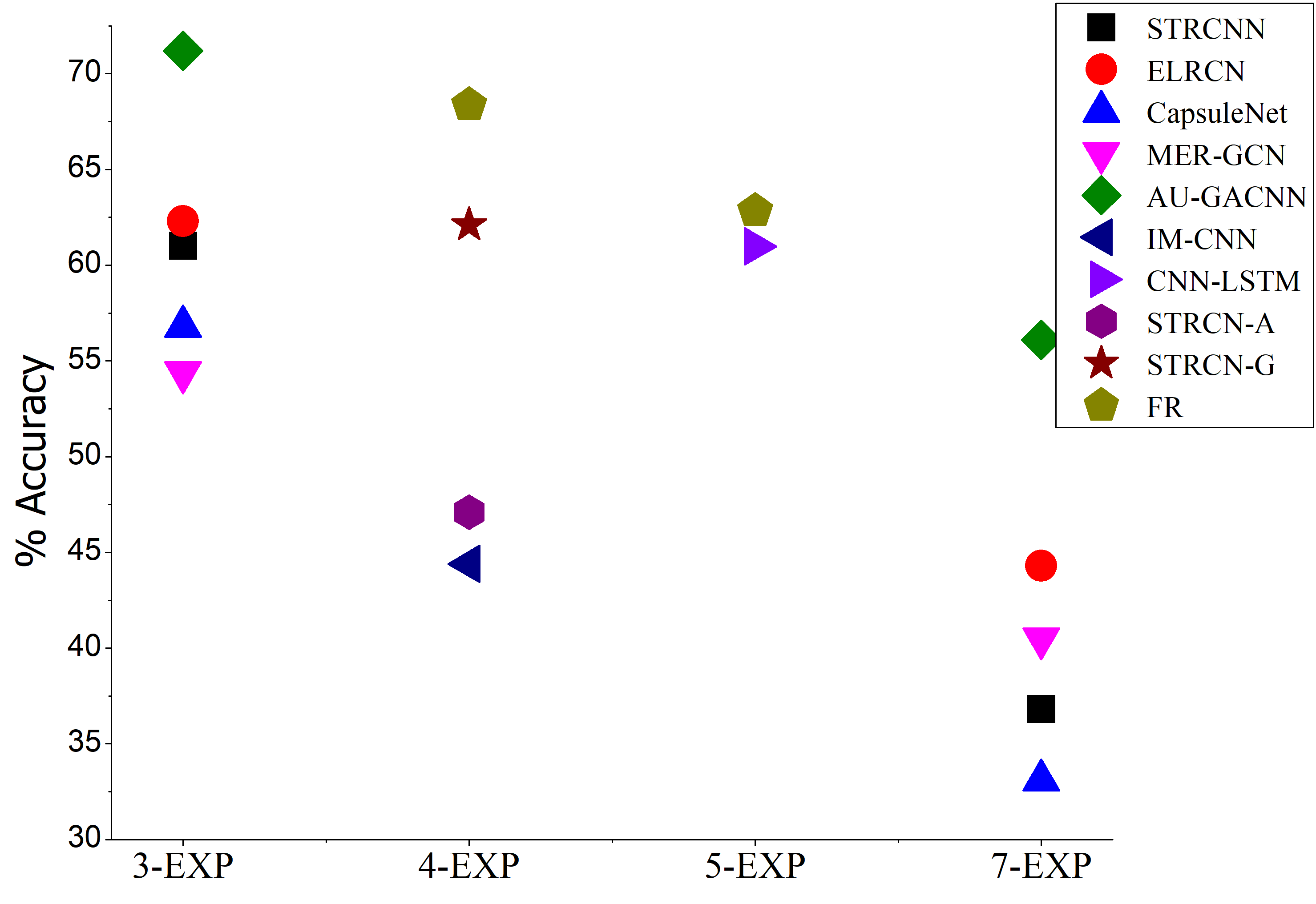}
    \caption{{The accuracy performance on four different emotion classes of CASME-II dataset for various existing MER frameworks: STRCNN \cite{xia2018spontaneous}, ELRCN \cite{khor2018enriched}, CapsuleNet \cite{van2019capsulenet}, MER-GCN  \cite{lo2020mer}, AU-GACNN \cite{xie2020assisted}, IM-CNN \cite{takalkar2017image}, CNN-LSTM \cite{kim2016micro}, STRCN-A \cite{xia2018spontaneous}, STRCN-G \cite{xia2018spontaneous}, and FR \cite{zhou2022feature}. \textit{Here, annotated results for STRCNN, ELRCN, CapsuleNet, MER-GCN, AU-GACNN are directly taken from the AU-GACNN \cite{xie2020assisted} published results. While, results of IM-CNN, CNN-LSTM, STRCN-A, STRCN-G, FR are grab from the FR \cite{zhou2022feature}.}}}
    \label{fig:20}
\end{figure}

To analyze the impact of emotion classes over the performance of a model, we analysed the existing MER approaches and results are highlighted in Fig. \ref{fig:20}. From Fig. \ref{fig:20}, it is clear that the performance of the models is improved by reducing the number of emotion classes. More specifically, models for 3- emotion classes gained the highest recognition accuracy and models for 4- emotion classes attained second highest and so on. Based on the observations and results of existing MER frameworks, we can conclude that a greater number of classes (7/8 emotion classes) create more confusion for models to generalize the emotion classes in true positives. However, in real life scenarios humans can exhibit a wide range of facial expressions. Thus, using a reduced set of emotions to carry on the experiments is a bit misleading. The fact that existing datasets lack enough labels for some emotions is a challenge that the research on the topic owes to face. Thereby, there is enough scope to provide a persuasive solution to handle such emotion classes instead of simply merging or dropping.

\section{Research Needs and Future Directions}\label{sec:fut}
This section describes the research needs and future directions of the MER approaches. More details of the critical issues in the literature which are unresolved or get least attention in the MER approaches are discussed. Therefore, there is scope for the upcoming researchers to design robust solutions for MER applications.

\subsection{Unbiased learning}
The publicly available ME’s datasets have imbalanced sample sizes in the emotion classes. The imbalancing nature of the dataset leads the CNN models bias towards a dominating class. Thus, there is an immense need to develop an unbiased learning algorithm to enhance the generalization ability of the model. To handle the imbalancing issue in the emotion classes, some of the existing MER approaches \cite{wang2018micro, Li3D-Flow} dropped the emotion classes, which contain very few samples. While some other approaches \cite{xia2018spontaneous, xie2020assisted} have created new emotion classes by merging the existing emotions as positive, negative, surprise, and other. However, combined emotion classes are still imbalanced. Therefore, Xia \etal \cite{xia2019spatiotemporal} took a step forward to resolve the imbalanced data samples problem and proposed temporal augmentation to create a balanced dataset. Furthermore, Xie \etal \cite{xie2020assisted} proposed a GAN based model to generate the synthetic data samples to alleviate the problem of unbalancing and trained a model with unbiased learning. From the above discussion it is evident that there is a need to develop a balanced dataset and robust deep learning technique to handle imbancing emotion class problems for MER. 

\subsection{Cross-cultured ME Dataset}
Most of the available ME datasets are developed in-lab environments with single cultured subjects. According to Xie \etal \cite{xie2002downside}, the study on single cultured samples lead to in-group downside problems. Thus, the model trained over single targeted ethnic participants may be biased and lead to under-perform on cross-cultured MEs. Therefore, there is a need to collect cross-cultured data samples of MEs. The cross-cultured ME datasets will ensure the fair analysis of MER frameworks and it is a great addition to the affective computing research community.

\begin{table*}[t!]
\centering
\caption{Experimental settings and  performance based comparison of existing DL approaches on CDE validation setup. }
\label{tab:7}
\begin{threeparttable}
\begin{tabular}{lccccccccccc}
\hline
\multirow{2}{*}{\textbf{Pub-Yr}} & \multirow{2}{*}{\textbf{Input Size}} & \multirow{2}{*}{\textbf{LR}} & \multirow{2}{*}{\textbf{Data Aug.}}  & \multicolumn{2}{c}{\textbf{CAS-II$\rightarrow$SMIC}} & \multicolumn{2}{c}{\textbf{CAS-II$\rightarrow$SAMM}} & \multicolumn{2}{c}{\textbf{SMIC $\rightarrow$CAS-II}} & \multicolumn{2}{c}{\textbf{SAMM$\rightarrow$CAS-II}} \\ \cline{5-12}
                                                                       &                              &                                     &                                    & \textbf{Acc}          & \textbf{F1}         & \textbf{Acc}         & \textbf{F1}          & \textbf{Acc}          & \textbf{F1}          & \textbf{Acc}          & \textbf{F1}          \\ \hline \hline
FG-18 \cite{khor2018enriched}                   & $224\times 224$                              & Adaptive                     & No                                                                & N/A                   & N/A                 & 43.4                & 34.1               & N/A                   & N/A                  & N/A                   & N/A                  \\
Neu.Co-19 \cite{wang2020micro}             & $224\times 224$                              & Adaptive                     & C.S, R, S                                  & N/A                   & N/A                 & 55.9                 & \textit{N/A}         & N/A                   & N/A                  & 58.4                  & \textit{N/A}         \\
ACMMM-20 \cite{xie2020assisted}               & N/A                                  & N/A                          & GAN Based                                                           & 34.4                  & 31.9                & N/A                  & N/A                  & N/A                   & N/A                  & N/A                   & N/A                  \\
IEEE-Acc-20 \cite{choi2020facial}             & $21\times 21$                                & N/A                          & No                                                                 & 55.5                 & 49.7               & N/A                  & N/A                  & 62.8                 & 57.8                & N/A                   & N/A                  \\
IEEE-MM-21 \cite{vermaAff}                & $112\times 112$                              & Fixed                        & F, R, T                                & N/A                   & N/A                 & 32.1                & N/A                  & N/A                   & N/A                  & 25.9                 & N/A          \\ \hline       
\end{tabular}
\begin{tablenotes}[para,flushleft]
   \textit{Here, LR and CAS implies for the learning rate and CASME dataset.}
  \end{tablenotes}
  \end{threeparttable}
\end{table*}
\subsection{Group Emotion Recognition}
Identifying the common emotion, shared among a group of people is known as group-level emotion recognition (GER). The GER plays a significant role in a wide variety of applications such as security, surveillance, early event prediction, image retrieval, and social era. Much study on GER using macro expression recognition is available in the literature \cite{dhall2017individual, veltmeijer2021automatic}. However, due to the challenges of MEs there is no group emotion dataset available for MER. Thus, there is scope to develop balanced and cross-cultured group emotion micro expression dataset.

\subsection{Motion Magnification in MER}
MEs are involuntary which cannot be captured in normal sight as they usually occur only for the minute interval. Therefore, it is hard to train a system to detect these variations and identify the relevant emotion class in MEs video sequences. In literature, Eulerian Video Magnification (EVM) \cite{wu2012eulerian} and learning-based MM \cite{oh2018learning} are adopted for magnifying the MEs. However, both approaches are not specifically designed for the MEs and sometimes destroy the emotional features. Therefore, there is a need to magnify the micro variations of emotion, developing application dependent  motion magnification algorithms could improve the performance of the model. 
\subsection{Multi-modal in MER}
Ongoing MER research is not achieving enough performance to use in real-time applications due to the low intensity and subtle nature. Moreover, the available dataset samples are not enough to train the MER frameworks. Therefore, to enhance the ability to recognize micro variations, in recent times multi-modal algorithms \cite{mittal2020m3er, mittal2020emoticon} gain the attention of researchers by supplying sufficient information to enhance the model performance. Thus, there is a need to develop an application specific (MER) dataset with multi-modalities like body gestures, eye gaze, Electrocardiogram (ECG), electroencephalogram (EEG), \textit{etc.} Also, there is an immense need to design a robust  algorithm to handle these multi modalities.

\subsection{MER in-wild}
In real world applications, to analyze the emotional state of a person, models designing in the lab environment may fail as the movement of the subject is dynamic, viewing angle of the camera is not static, illumination and lighting conditions are dynamic in nature. Due to the challenging nature of micro expressions, only one data set in-wild for MER \cite{husak2017spotting} is available in the literature. Also, extremely limited articles on MER in the wild are available in the literature.  Therefore, there is a lot of scope to develop a robust algorithm to handle the challenges of the in-wild and scope to develop cross-cultured dataset in the wild for MER. 

\subsection{MER in Psychological Disorders}
Psychological disorders such as autism spectrum, bipolar, anxiety and stress related disorders affect a person’s thoughts, behavior, feelings and sense of well-being. In these disorders people have a state of low mood and aversion to activity. In such a low mood and lack of interest, the facial expression appears different from the ones in normal states. Some research \cite{koizumi2014relationship, valstar2016avec} have been made for psychological disorders through facial expression recognition in past years. However, incredible challenges in dataset accumulation due to privacy of neurotic subjects, limited labeled datasets and lack of deep learning based solutions make it an appealing research area to be explored. The available datasets are not focusing on the privacy of neurotic subjects which is a very crucial aspect in social lives. The future work requires more attention towards both data collection and algorithm development for psychological disorders analysis through FER/MER.
\subsection{MER in Entertainment}
Online games have gained a lot of popularity due to the features of collaboration, communication, and interaction. However, the virtual world of gaming is still primitive and far from real-world communication. For example, players still communicate through text chats, avatars have no activities related to natural body gestures, facial expressions, and so forth. Therefore, there is a need for a robust automatic expression system that can be integrated into the gaming system to control the facial expressions of avatars and enhance the interest of players by providing a virtual interface near the real world. Moreover, facial expression analysis can be integrated into video-controlled devices for entertainment (music, movies, games, YouTube, \textit{etc.}) and the dynamic balancing system will automatically adjust the entertainment level based on the user’s facial expressions.

\subsection{MER in Education System}
In recent COVID pandemic years, the standard offline education systems transferred to the online education system. Effective online education is the primary need to maintain the education gap of each age group of students.  The online education system allows students from anywhere to access the classes as well as experimental work from a distance via the Internet. However, this is way far away from physical education. In such critical situations students as well as tutors face many challenges like lack of attention, high chances of distraction \textit{etc.} Resultant performance of the students degrades and it increases the chances of many mental problems such as stress, anxiety and depression. Thus, there is a lot of scope to develop a robust online education system by integration of facial expression analysis to scan the expressive features of the students. This will allow tutors to survey the attentiveness of the students in class and help them accordingly. 
\section{Conclusion}\label{sec:con}

This paper presents deep insights of learning-based MER frameworks with a perspective on promises in model designing, experiment strategies, challenges, and research needs. Particularly, the existing learning-based MER frameworks are analyzed in terms of model design and evaluation frameworks. The variety of existing deep learning architectures are examined and their effect on MER performances are discussed. The important paradigms in model designing and evaluation for MER are presented. Also, the impact of data division strategies like PDE, PIE and CDE on the model performance and the limitations of these strategies are discussed. The challenges in designing robust MER models and the current research needs are discussed. Further, the available datasets, challenges and the evaluation metrics utilized to test the efficacy of MER frameworks are discussed. In addition, presented useful insights of future needs and research guidance to carry forward the research in MER.

\bibliographystyle{IEEEtran}

\bibliography{egbib}
\end{document}

\title{Deep Insights of Learning based MER Frameworks: A Perspective on Promises, Challenges and Research Needs\\
(Supplementary Document)}
%
%
% author names and IEEE memberships
% note positions of commas and nonbreaking spaces ( ~ ) LaTeX will not break
% a structure at a ~ so this keeps an author's name from being broken across
% two lines.
% use \thanks{} to gain access to the first footnote area
% a separate \thanks must be used for each paragraph as LaTeX2e's \thanks
% was not built to handle multiple paragraphs
%
%
%\IEEEcompsocitemizethanks is a special \thanks that produces the bulleted
% lists the Computer Society journals use for "first footnote" author
% affiliations. Use \IEEEcompsocthanksitem which works much like \item
% for each affiliation group. When not in compsoc mode,
% \IEEEcompsocitemizethanks becomes like \thanks and
% \IEEEcompsocthanksitem becomes a line break with idention. This
% facilitates dual compilation, although admittedly the differences in the
% desired content of \author between the different types of papers makes a
% one-size-fits-all approach a daunting prospect. For instance, compsoc 
% journal papers have the author affiliations above the "Manuscript
% received ..."  text while in non-compsoc journals this is reversed. Sigh.

\author{Monu~Verma,
        Santosh~Kumar~Vipparthi,
        and~Girdhari~Singh,% <-this % stops a space
}
% note the % following the last \IEEEmembership and also \thanks - 
% these prevent an unwanted space from occurring between the last author name
% and the end of the author line. i.e., if you had this:
% 
% \author{....lastname \thanks{...} \thanks{...} }
%                     ^------------^------------^----Do not want these spaces!
%
% a space would be appended to the last name and could cause every name on that
% line to be shifted left slightly. This is one of those "LaTeX things". For
% instance, "\textbf{A} \textbf{B}" will typeset as "A B" not "AB". To get
% "AB" then you have to do: "\textbf{A}\textbf{B}"
% \thanks is no different in this regard, so shield the last } of each \thanks
% that ends a line with a % and do not let a space in before the next \thanks.
% Spaces after \IEEEmembership other than the last one are OK (and needed) as
% you are supposed to have spaces between the names. For what it is worth,
% this is a minor point as most people would not even notice if the said evil
% space somehow managed to creep in.

% The paper headers
\markboth{Journal of \LaTeX\ Class Files,~Vol.~XX, No.~XX, Month~XXXX}%
{}
% The only time the second header will appear is for the odd numbered pages
% after the title page when using the twoside option.
% 
% *** Note that you probably will NOT want to include the author's ***
% *** name in the headers of peer review papers.                   ***
% You can use \ifCLASSOPTIONpeerreview for conditional compilation here if
% you desire.

% The publisher's ID mark at the bottom of the page is less important with
% Computer Society journal papers as those publications place the marks
% outside of the main text columns and, therefore, unlike regular IEEE
% journals, the available text space is not reduced by their presence.
% If you want to put a publisher's ID mark on the page you can do it like
% this:
%\IEEEpubid{0000--0000/00\$00.00~\copyright~2015 IEEE}
% or like this to get the Computer Society new two part style.
%\IEEEpubid{\makebox[\columnwidth]{\hfill 0000--0000/00/\$00.00~\copyright~2015 IEEE}%
%\hspace{\columnsep}\makebox[\columnwidth]{Published by the IEEE Computer Society\hfill}}
% Remember, if you use this you must call \IEEEpubidadjcol in the second
% column for its text to clear the IEEEpubid mark (Computer Society jorunal
% papers don't need this extra clearance.)

% use for special paper notices
%\IEEEspecialpapernotice{(Invited Paper)}

% for Computer Society papers, we must declare the abstract and index terms
% PRIOR to the title within the \IEEEtitleabstractindextext IEEEtran
% command as these need to go into the title area created by \maketitle.
% As a general rule, do not put math, special symbols or citations
% in the abstract or keywords.
\IEEEtitleabstractindextext{%

% Note that keywords are not normally used for peerreview papers.
}

% make the title area
\maketitle

% To allow for easy dual compilation without having to reenter the
% abstract/keywords data, the \IEEEtitleabstractindextext text will
% not be used in maketitle, but will appear (i.e., to be "transported")
% here as \IEEEdisplaynontitleabstractindextext when the compsoc 
% or transmag modes are not selected <OR> if conference mode is selected 
% - because all conference papers position the abstract like regular
% papers do.
\IEEEdisplaynontitleabstractindextext
% \IEEEdisplaynontitleabstractindextext has no effect when using
% compsoc or transmag under a non-conference mode.

% For peer review papers, you can put extra information on the cover
% page as needed:
% \ifCLASSOPTIONpeerreview
% \begin{center} \bfseries EDICS Category: 3-BBND \end{center}
% \fi
%
% For peerreview papers, this IEEEtran command inserts a page break and
% creates the second title. It will be ignored for other modes.
\IEEEpeerreviewmaketitle
\begin{table}[!bp]
\caption{The computational complexity based comparison of existing DL approaches.}
\label{param}
\begin{tabular}{lccc}
\hline
Methods & Input Size &\# Parameters (M) & Ex. Time (s) \\ \hline \hline
AlexNet [M:75] & $227\times 227$ &61& 12.90 \\
Squeeze \cite{iandola2016squeezenet} & $227\times 227$ &1.24& 14.37\\
GoogleNet \cite{szegedy2015going} & $224\times 224$ &7& 29.30\\
VGG-16 [M:76] & $224\times 224$ &138& 95.44\\
ELRCN [M:36] & N/A &219&  N/A             \\
OffApex [M:34] & $28\times 28$ &2.77&  5.56             \\
STSTNet [M:39] & $28\times 28$ & 0.00167 & 5.74     \\
CNN-LSTM [M:45]& $64\times 64$ &4.52 &N/A\\
SSSN [M:32]& N/A & 0.63&N/A\\
DSSN [M:32]& N/A & 0.97&N/A\\
Micro-Attention [M:50]&$224\times 224$ &5.9  & 1.1\\
LEARNet [M:29]&$112\times 112$&1.8&N/A \\
AfffectiveNet [M:30]&$112\times 112$&2.2& 8.7      \\
OrigiNet [M:31]&$112\times 112$&1.8&N/A\\
FeatRef [M:61]&$28\times 28$ & 11.4231& 36.77 \\ \hline      
\end{tabular}
\end{table}
\begin{figure*}
    \centering
     \includegraphics[width=0.85\textwidth]{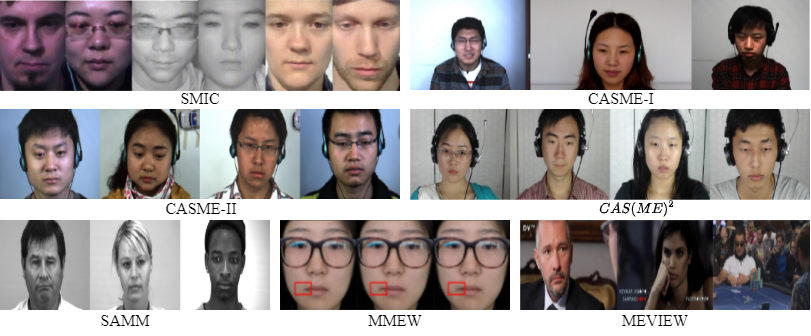}
    \caption{Data samples of all available MER datasets: a) SMIC, b) CASME-I, c) CASME-II, d) CAS(ME)$^2$, e) SAMM, f) MMEW, and g) MEVIEW.}
    \label{datasamples}
\end{figure*}

\begin{table*}[!t]
  \caption{Statistical details of seven publicly released micro-expression datasets.}\label{stas}
\begin{tabular}{lccccccccc}
\hline
\multirow{2}{*}{\textbf{ Features}} & \multicolumn{3}{c}{\textbf{SMIC} }  & \multirow{2}{*}{\textbf{CASME-I}}& \multirow{2}{*}{\textbf{CASME-II} } &\multirow{2}{*}{\textbf{CAS(ME)$^2$}}  & \multirow{2}{*}{\textbf{SAMM}}   & \multirow{2}{*}{\textbf{MMEW}}   & \multirow{2}{*}{\textbf{MEVIEW} }                                                              \\
& HS    & VIS    & NIR    &   &  & && &\\ \hline \hline
\textbf{\textit{Total Samples  }}  & 164& 71 & 71 & 195  & 247 & 57   & 159   & 300      & 40  \\ \hline
\textbf{\textit{Participants}}& 16  & 8   & 8  & 35  & 35 & 22 & 32 & 36  & 16  \\ \hline
\textbf{\textit{Mean Age}}  & \multicolumn{3}{l}{N/A}  & 22.03  & 22.03    & 22.59  & 33.24  & 22.35  & N/A  \\ \hline
\textbf{\textit{Environment}}  & \multicolumn{3}{l}{Lab}   & Lab   & Lab    & Lab    & Lab   & Lab  & Wild \\ \hline
\textbf{\textit{Ethnicities}}   & \multicolumn{3}{l}{3} & 1 & 1     & 1  & 13  & 1 & N/A  \\ \hline
\textbf{\textit{Image Resolution}}          & \multicolumn{3}{l}{$640\times 480$}& \begin{tabular}[c]{@{}l@{}}$640\times 480$ \& \\ $1280\times 720$ \end{tabular} & $640\times 480$  & $ 640\times 480$       & $2040\times 1088$ & $1920\times 1080$  & $1280\times 720$  \\ \hline
\textbf{\textit{Facial Resolution }}        &\multicolumn{3}{l}{$190\times 230$}                           & $150\times 190$  & $280\times 340$  & N/A   & $400\times 400$ & $400\times 400$ & N/A      \\ \hline
\textbf{\textit{Frame Rate}} & 100   & 25 & 25 & 60& 200& 30& 200  & 90 & 25    \\ \hline
\textbf{\textit{Emotion Categories}}        & \multicolumn{3}{l}{\begin{tabular}[c]{@{}l@{}}Positive:   107\\ Negative:   116\\ Surprise:   83\end{tabular}} & \begin{tabular}[c]{@{}l@{}}Amusement:   5\\  Disgust:   88\\    Sadness:   6\\ Contempt:   3\\    Fear:   2\\    Tense:   28\\    Surprise:   20    \\ Repression:   40\end{tabular} & \begin{tabular}[c]{@{}l@{}}Happiness:   33\\   Repression:   27\\    Surprise:   25\\    Disgust:   60\\    Others:   102\end{tabular} & \begin{tabular}[c]{@{}l@{}}Positive:   8\\    Negative:   21\\ Surprise:   9\\    Others:   19\end{tabular} & \begin{tabular}[c]{@{}l@{}}Happiness:   24\\   Surprise:   13\\ Anger:   20\\    Disgust:   8\\ Sadness:   3\\    Fear:   7\\   Others:   84\end{tabular} & \begin{tabular}[c]{@{}l@{}}Happiness:   36\\    Anger:   8\\    Surprise:   89\\    Disgust:   72\\    Fear:   16\\ Sadness:   13\\ Others:   66\end{tabular} & \begin{tabular}[c]{@{}l@{}}Happiness:   6\\ Anger:   2   \\ Surprise:   9   \\ Disgust:   1    \\ Fear:   3\\Contempt: 6\\     Unclear:   13\end{tabular} \\ \hline
\textit{\textbf{Available at: }}  & \multicolumn{3}{l}{\begin{tabular}[c]{@{}l@{}}https://wwww.cs\\e.oulu.fi/SMICD\\atabase\end{tabular}}                                                                  & \begin{tabular}[c]{@{}l@{}}http://fu.psych.ac\\.cn/CASME//cas\\me-en.php \end{tabular} & \begin{tabular}[c]{@{}l@{}}http://fu.psych.ac.\\cn/CASME//cas\\me2-en.php \end{tabular}  & \begin{tabular}[c]{@{}l@{}}http://fu.psych\\.ac.cn/CASME\\//cas(me)2-en.\\php \end{tabular}                                                                           & \begin{tabular}[c]{@{}l@{}}http://www2.\\docm.mmu.ac.\\uk/STAFF/m.\\yap/dataset.php  \end{tabular}  & \begin{tabular}[c]{@{}l@{}}https://www.\\dpailab.com/\\databas.html \end{tabular}  & \begin{tabular}[c]{@{}l@{}}https://cmp.felk\\.cvut.cz/$\sim$cechj\\/ME \end{tabular}  \\ \hline   
\end{tabular}
\end{table*}

\begin{table*}[!t]
\centering
\caption{Summery of the traditional features based MER approaches over PDE and PIE validation setups. }
\label{tab:1}
\begin{threeparttable}
\begin{tabular}{lcccccccccccccc}
\hline \noalign{\smallskip}
\multirow{2}{*}{\textbf{Pub-yr}}           & \multirow{2}{*}{\textbf{Input}}                                                                  & \multirow{2}{*}{\textbf{Prepr.}}                                                          & \multirow{2}{*}{\textbf{Features}}                                   & \multirow{2}{*}{\textbf{Classifier}}                                                      & \multirow{2}{*}{\textbf{Prot.}}                             & \multicolumn{5}{c}{\textbf{SMIC}}                                                             & \multicolumn{4}{c}{\textbf{CASME-II}}                                                       \\ \cline{7-15} \noalign{\smallskip}
                                  &                                                                                         &                                                                                  &                                                             &                                                                                  &                                                    & \textbf{E}                  & \multicolumn{3}{c}{\textbf{Acc.}}                 & \textbf{F1}                   & \textbf{E}                  & \multicolumn{2}{c}{\textbf{Acc.}}               & \textbf{F1}                   \\ \hline \hline
\multirow{2}{*}{IJCNN-14 \cite{guo2014micro}} & \multirow{2}{*}{Video}                                                                  & \multirow{2}{*}{-}                                                               & \multirow{2}{*}{LBP-TOP}                                    & \multirow{2}{*}{\begin{tabular}[c]{@{}c@{}}Nearest \\ Neighbors\end{tabular}}    & LOSO & \multirow{2}{*}{3} & \multicolumn{3}{c}{53.72}                & N/A                  & \multirow{2}{*}{-} & \multicolumn{2}{c}{\multirow{2}{*}{-}} & \multirow{2}{*}{-}   \\
                                  &                                                                                         &                                                                                  &                                                             &                                                                                  & PDE                                                 &                    & \multicolumn{3}{c}{65.83}                &                      &                    & \multicolumn{2}{c}{}                   &                      \\
ICPR-14 \cite{TICS}                    & Video (150F)                                                                            & \begin{tabular}[c]{@{}c@{}}ASM,\\  LI\end{tabular}                               & TICS                                                        & SVM                                                                              & LOSO                                               & -                  & \multicolumn{3}{c}{-}                    & -                    & 5                  & \multicolumn{2}{c}{61.76}              & N/A                  \\
\multirow{2}{*}{ACCV-14 [M:37]}     & \multirow{2}{*}{Video}                                                                  & \multirow{2}{*}{-}                                                               & \multirow{2}{*}{LBP-SIP}                                   & \begin{tabular}[c]{@{}c@{}}SVM-\\ Linear\end{tabular}                            & \multirow{2}{*}{LOSO}                              & \multirow{2}{*}{3} & \multicolumn{3}{c}{64.02}                & \multirow{2}{*}{N/A} & \multirow{2}{*}{5} & \multicolumn{2}{l}{63.56}              & \multirow{2}{*}{N/A} \\
                                  &                                                                                         &                                                                                  &                                                             & \begin{tabular}[c]{@{}c@{}}SVM-\\ RBF\end{tabular}                               &                                                    &                    & \multicolumn{3}{c}{62.80}                &                      &                    & \multicolumn{2}{c}{66.40}              &                      \\
ICCVW-15 [M:36]                 & Video                                                                                   & \begin{tabular}[c]{@{}c@{}}ASM, \\ BLI\end{tabular}                              & STLBP-IP                                                    & \begin{tabular}[c]{@{}c@{}}SVM-\\ Chi-Square\end{tabular}                        & LOSO                                               & 3                  & \multicolumn{3}{c}{57.93}                & N/A                  & 5                  & \multicolumn{2}{c}{59.51}              & N/A                  \\
TIP-15 [M:46]                   & Video (70 F)                                                                            & LI                                                                               & TICS                                                        & \begin{tabular}[c]{@{}c@{}}SVM-\\ Linear\end{tabular}                            & LOSO                                               & -                  & \multicolumn{3}{c}{-}                    & -                    & 4                  & \multicolumn{2}{l}{61.90}              & N/A                  \\
\multirow{2}{*}{TAFF-16 [M:43]}   & \multirow{2}{*}{Video (70 F)}                                                           & \multirow{2}{*}{\begin{tabular}[c]{@{}c@{}}V.J., O.F.,  \\ DRMF\end{tabular}} & \multirow{2}{*}{MDMO}                                       & \multirow{2}{*}{\begin{tabular}[c]{@{}c@{}}SVM-\\ Polynomial\end{tabular}} & LOVO & \multirow{2}{*}{3} & \multicolumn{3}{c}{\multirow{2}{*}{N/A}} & \multirow{2}{*}{N/A} & \multirow{2}{*}{4} & \multicolumn{2}{c}{70.34}              & \multirow{2}{*}{N/A} \\
                                  &                                                                                         &                                                                                  &                                                             &                                                                                  &  LOSO &                    & \multicolumn{3}{c}{}                     &                      &                    & \multicolumn{2}{c}{57.16}              &                      \\
TAFF-17 \cite{xu2017microexpression}                   & 10-20 F                                                                                 & \begin{tabular}[c]{@{}c@{}}LWM, \\ O. F., \\ LI\end{tabular}                     & FDM                                                         & \begin{tabular}[c]{@{}c@{}}SVM- \\ RBF\end{tabular}                              & LOSO                                               & 3                  & \multicolumn{3}{c}{54.88}                & 0.53                 & 7                  & \multicolumn{2}{c}{45.93}              & 0.40                 \\
\multirow{6}{*}{TAFF-17 [M:40]}   & \multirow{6}{*}{10-35 Frames}                                                           & \multirow{6}{*}{\begin{tabular}[c]{@{}c@{}}TIM, \\    \\ O. F.\end{tabular}}     & \multirow{6}{*}{FHOFO}                                      & \multirow{2}{*}{\begin{tabular}[c]{@{}c@{}}SVM- \\ Linear\end{tabular}}          & LOSO                                               & \multirow{6}{*}{3} & \multicolumn{3}{c}{51.22}                & 0.51                 & \multirow{6}{*}{4} & \multicolumn{2}{c}{55.86}              & 0.51                 \\
                                  &                                                                                         &                                                                                  &                                                             &                                                                                  & LOVO                                               &                    & \multicolumn{3}{c}{56.10}                & 0.55                 &                    & \multicolumn{2}{c}{64.06}              & 0.60                 \\
                                  &                                                                                         &                                                                                  &                                                             & \multirow{2}{*}{KNN}                                                             & LOSO                                               &                    & \multicolumn{3}{c}{42.68}                & 0.44                 &                    & \multicolumn{2}{c}{50.39}              & 0.40                 \\
                                  &                                                                                         &                                                                                  &                                                             &                                                                                  & LOVO                                               &                    & \multicolumn{3}{c}{50.61}                & 0.53                 &                    & \multicolumn{2}{c}{55.08}              & 0.47                 \\
                                  &                                                                                         &                                                                                  &                                                             & \multirow{2}{*}{LDA}                                                             & LOSO                                               &                    & \multicolumn{3}{c}{48.17}                & 0.48                 &                    & \multicolumn{2}{c}{49.22}              & 0.44                 \\
                                  &                                                                                         &                                                                                  &                                                             &                                                                                  & LOVO                                               &                    & \multicolumn{3}{c}{59.15}                & 0.58                 &                    & \multicolumn{2}{c}{62.50}              & 0.58                 \\
ToM-18 \cite{zong2018learning}                    & Video (20-60F)                                                                          & TBLI                                                                             & \begin{tabular}[c]{@{}c@{}}H.STLBP\\-RIP\end{tabular}                                                 & KGSL                                                                             & LOSO                                               & 3                  & \multicolumn{3}{c}{66.46}                & 0.65                 & 5                  & \multicolumn{2}{c}{65.18}              & 0.62                 \\ TIP-18 \cite{zong2018domain}                                                   & Video                                                                                   & \begin{tabular}[c]{@{}c@{}}ASM, \\ BLI\end{tabular}         & \begin{tabular}[c]{@{}c@{}}LBP- \\ Variants\end{tabular}    & DRL                                                  & CDE        & 3 & \textbf{\textit{C/S}}         & \multicolumn{2}{c}{62.16}   & N/A     & 3 & \textbf{\textit{S/C}}            & 54.88       & N/A    \\
SP:IC-18 \cite{liong2018less}                  & \begin{tabular}[c]{@{}c@{}}Onset-\\ Apex-\\ Offset frame\end{tabular}                   & O. F.                                                                            & Motion                                                      & \begin{tabular}[c]{@{}c@{}}PCA-\\ SVM-\\ RBF\end{tabular}                        & LOSO                                               & 3                  & \multicolumn{3}{c}{71.95}                & N/A                  & 5                  & \multicolumn{2}{c}{69.11\%}            & N/A                  \\

TAFF-18 [M:44]                  & Video & \begin{tabular}[c]{@{}c@{}}TIM, \\    \\ O. F.\end{tabular}                      & \begin{tabular}[c]{@{}c@{}}Sparse\\    \\ MDMO\end{tabular} & \begin{tabular}[c]{@{}c@{}}LIBSVM-\\ Polynomial\end{tabular}                     & LOSO                                               & 3                  & \multicolumn{3}{c}{70.51}                & 0.   70              & 5                  & \multicolumn{2}{c}{66.95}              & 0.69                 \\
ToC-19 \cite{zong2019toward}                                                    & Video                                                                                   & -                                                           & LBP-TOP                                                     & \begin{tabular}[c]{@{}c@{}}TTRM,\\ ASSM\end{tabular} & CDE & 3 & \textbf{\textit{C/S}}       & \multicolumn{2}{c}{59.46}   & N/A     & 3 & \textbf{\textit{S/C}}         & 53.05       & N/A    \\

IE. Acc.-20 \cite{zhang2020new}              & Video                                                                                   & AAM                                                                              & LBP-TOP                                                     & \begin{tabular}[c]{@{}c@{}}SVM-\\ Chi-Square\end{tabular}                        & LOSO                                               & 3                  & \multicolumn{3}{c}{59.71}                & N/A                  & 5                  & \multicolumn{2}{c}{59.51}              & N/A   \\ TKD-20 \cite{zhang2020cross}                                                   & Video (16F)                                                                             & TIM                                                         & \begin{tabular}[c]{@{}c@{}}LBP-\\ Variants\end{tabular}     & RSTR                                                 & CDE        & 3 & \multicolumn{2}{c} \textbf{\textit{C/S}} & 60.77         & 0.5622  & 3 & \textbf{\textit{S/C}}           & 54.27       & 0.5297
\\ \hline              
\end{tabular}
\begin{tablenotes}[para,flushleft]
   \textit{Here, Pub-Yr, Prepr., Prot., E, Acc., F1, O. F., C/S and S/C, represents the publication year, preprocessing techniques, evaluation protocols, expression classes, accuracy, f1 score, optical flow, training over CASME-II $\&$ test done using SMIC and vice versa. In addition, ASM, AAM, LI, V.J., F and N/A imply for active shape model, active appearance model, linear interpolation, Viola Jones, frame and not available respectively. M: implies for manuscript citations.}
  \end{tablenotes}
  \end{threeparttable}
\end{table*}

\begin{figure}

     \centering
     \begin{subfigure}[b]{1\linewidth}
         \centering
         \includegraphics[width=\textwidth,height=1.4in]{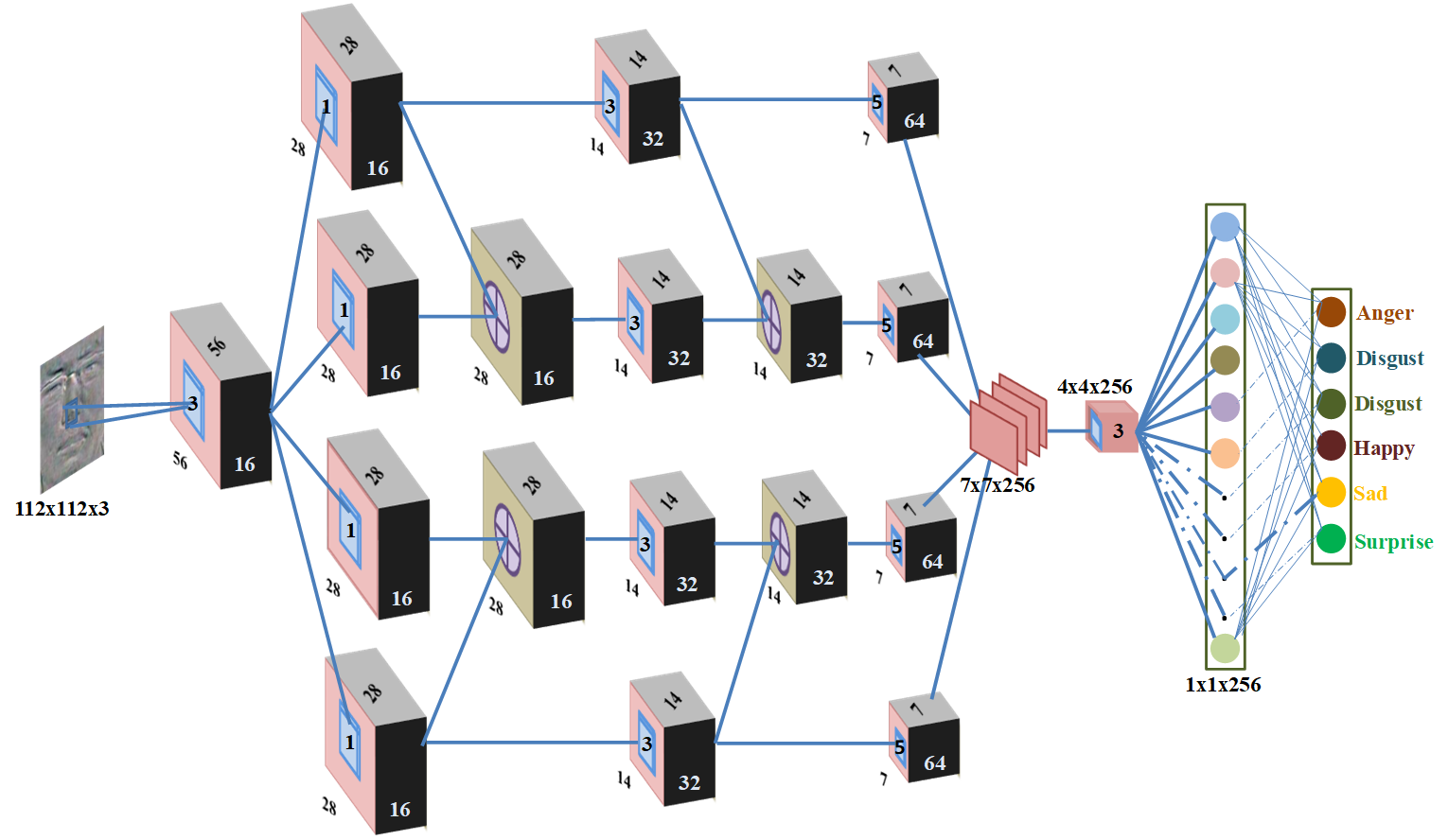}
         \caption{}
         \label{fig:4a}
     \end{subfigure}
     \hfill
     \begin{subfigure}[b]{1\linewidth}
         \centering
         \includegraphics[width=\textwidth,height=1.4in]{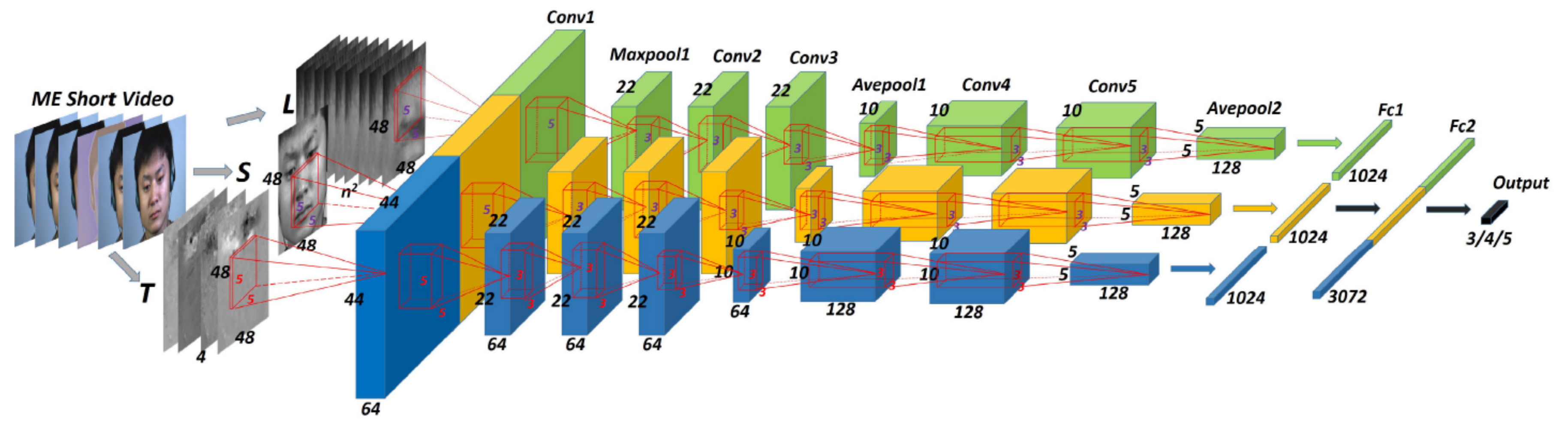}
         \caption{}
         \label{fig:4b}
     \end{subfigure}
     \hfill
     \begin{subfigure}[b]{1\linewidth}
         \centering
         \includegraphics[width=\textwidth, height=1.8in]{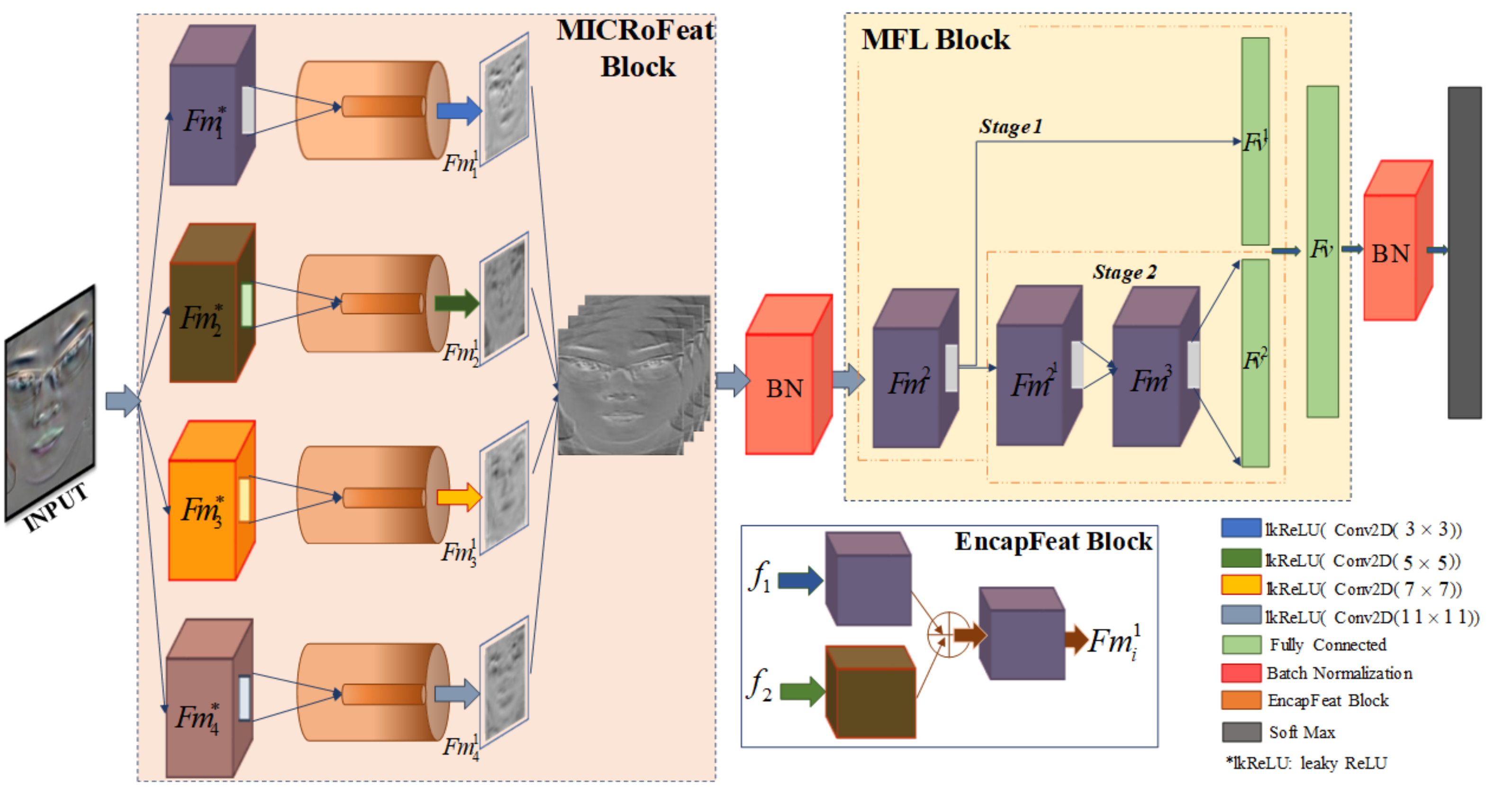}
         \caption{}
         \label{fig:4c}
     \end{subfigure}
        \caption{The two-stage MER approaches based on (a) 2D-CNN [M:62] (b) multi-stream [M:65] and (b) multi-scale [M:63] convolutional layers. These approaches also utilized different input formats: (a)dynamic imaging, (b) optical flow, and (c) affective motion images.}
        \label{fig:4}
\end{figure}

\begin{figure}
     \centering
     \begin{subfigure}[b]{1\linewidth}
         \centering
         \includegraphics[width=\textwidth]{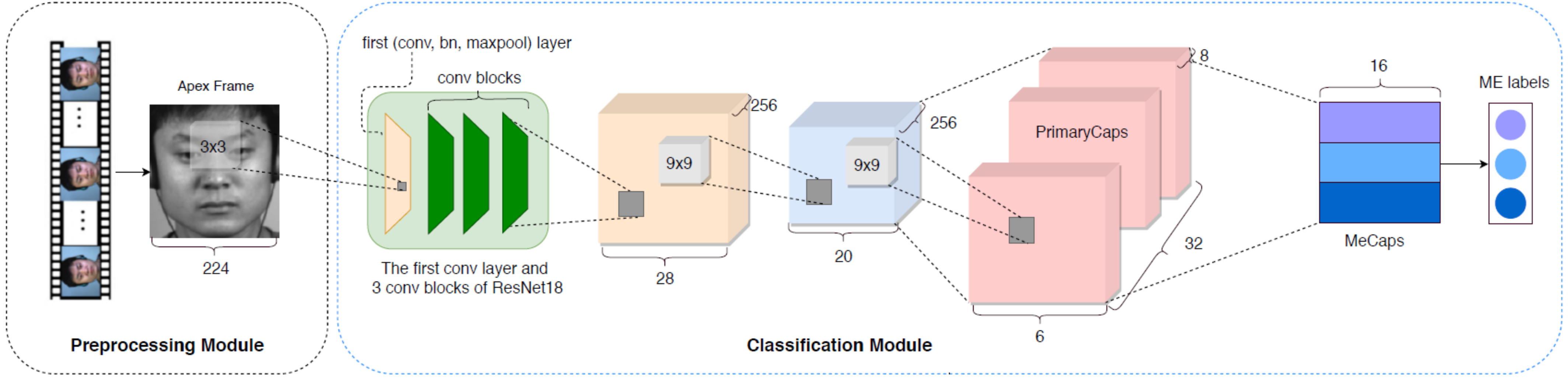}
         \caption{}
         \label{fig:5a}
     \end{subfigure}
     \hfill
     \begin{subfigure}[b]{1\linewidth}
         \centering
         \includegraphics[width=\textwidth]{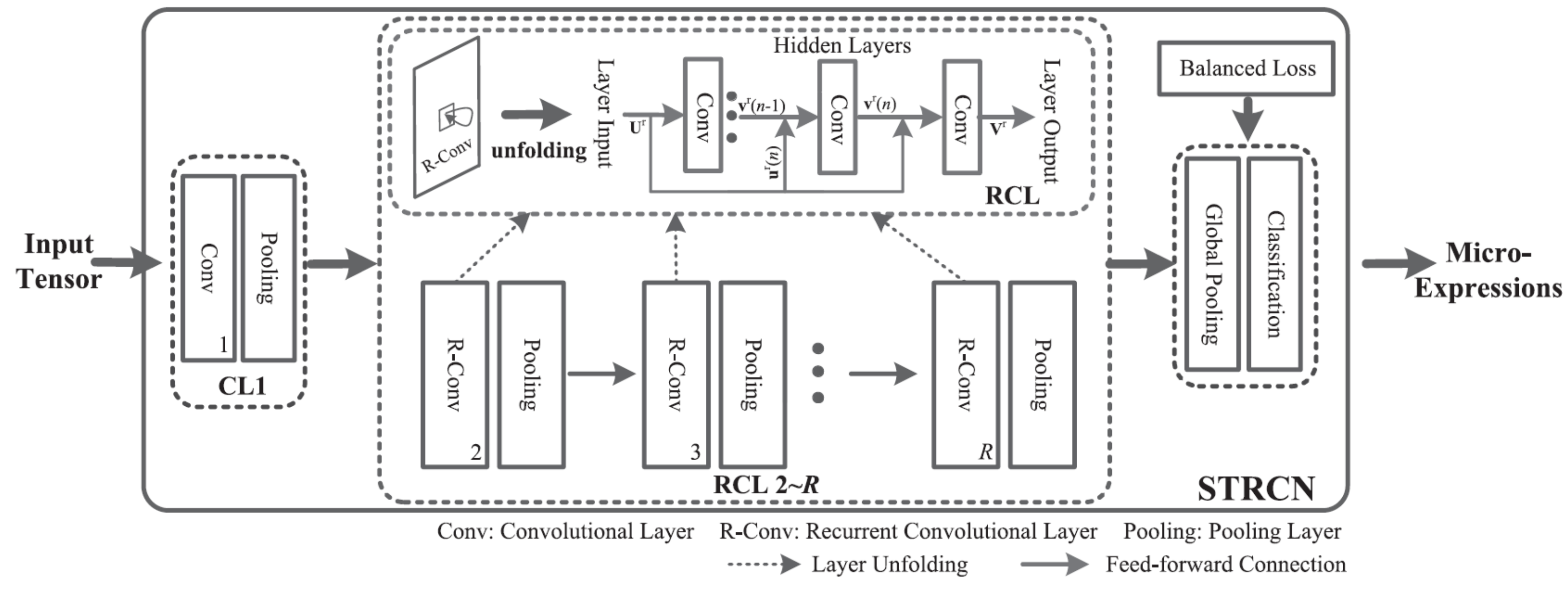}
         \caption{}
         \label{fig:5b}
     \end{subfigure}
     
        \caption{Multi-stage methods with (a) capsule module [M:83] and (b) RCN [M:31] module for MER.}
        \label{fig:5}
\end{figure}

\begin{figure}[]
     \centering
     \begin{subfigure}[b]{1\linewidth}
         \centering
         \includegraphics[width=\textwidth]{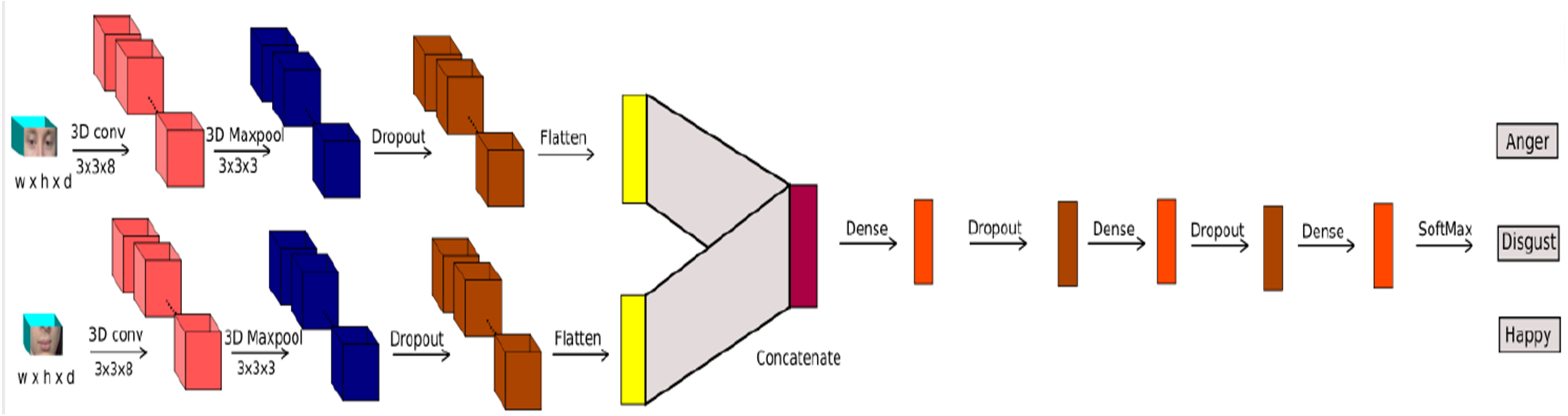}
         \caption{}
         \label{fig:6a}
     \end{subfigure}
     \hfill
     \begin{subfigure}[b]{1\linewidth}
         \centering
         \includegraphics[width=\textwidth]{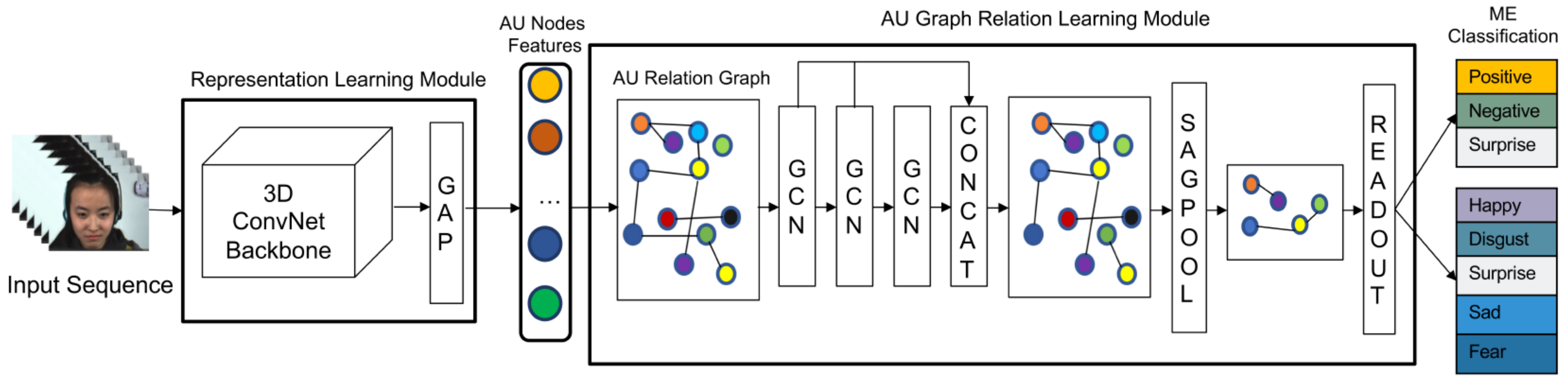}
         \caption{}
         \label{fig:6b}
     \end{subfigure}
     
        \caption{MER approaches (end-to-end) based on (a) 3D-CNN [M:90] and (b) AUs Graph relations [M:86]}
        \label{fig:6}
\end{figure}

\begin{figure}[!t]
     \centering
    
          \includegraphics[width=\linewidth, height=1.8in]{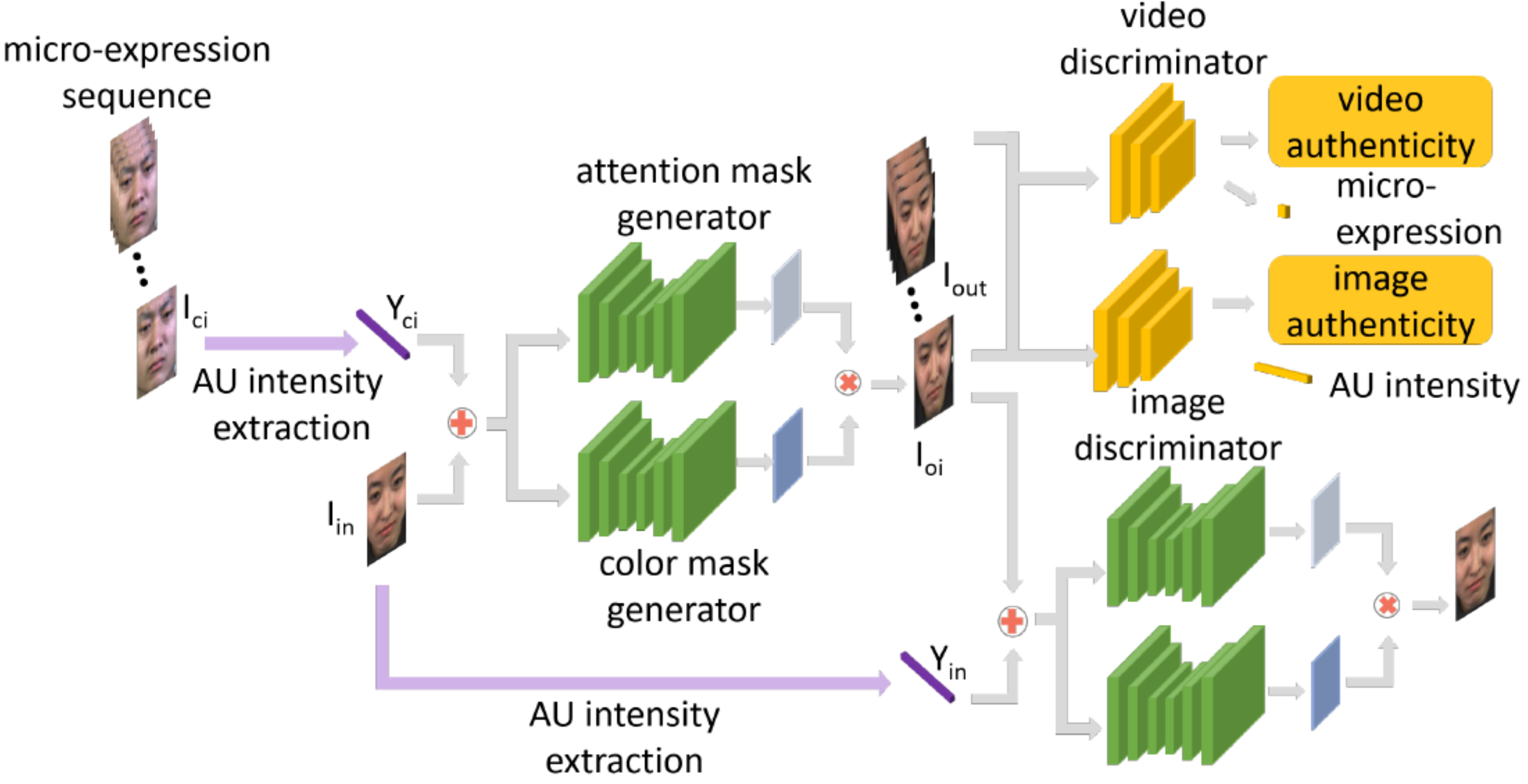}
     
        \caption{Representing GAN based synthetic image generation methods [M:53] for MEs.}
        \label{fig:7}
\end{figure}

\begin{figure}[!t]
     \centering
     \begin{subfigure}[b]{1\linewidth}
         \centering
         \includegraphics[width=\textwidth]{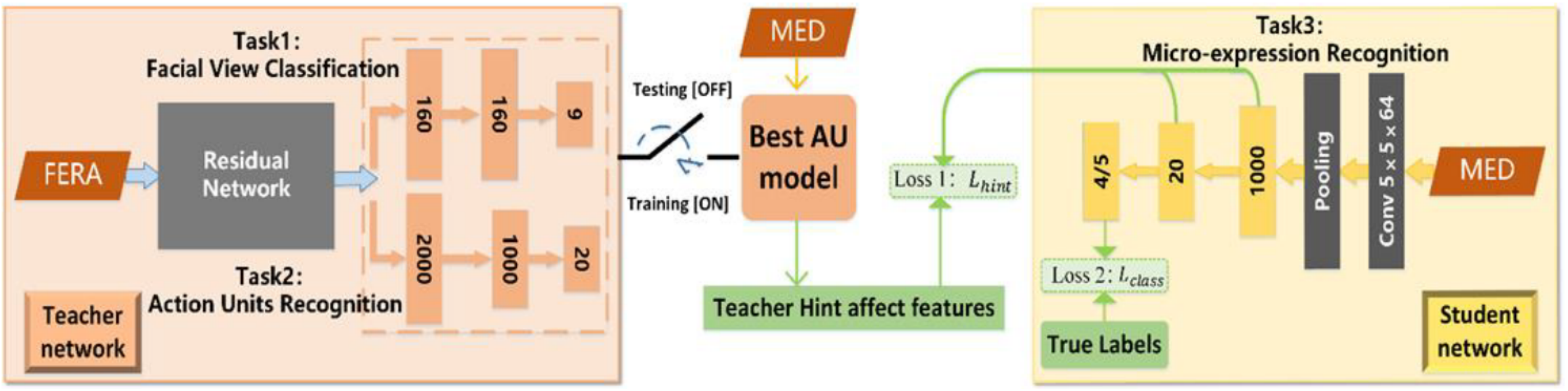}
         \caption{}
         \label{fig:8a}
     \end{subfigure}
     \hfill
     \begin{subfigure}[b]{1\linewidth}
         \centering
         \includegraphics[width=\textwidth]{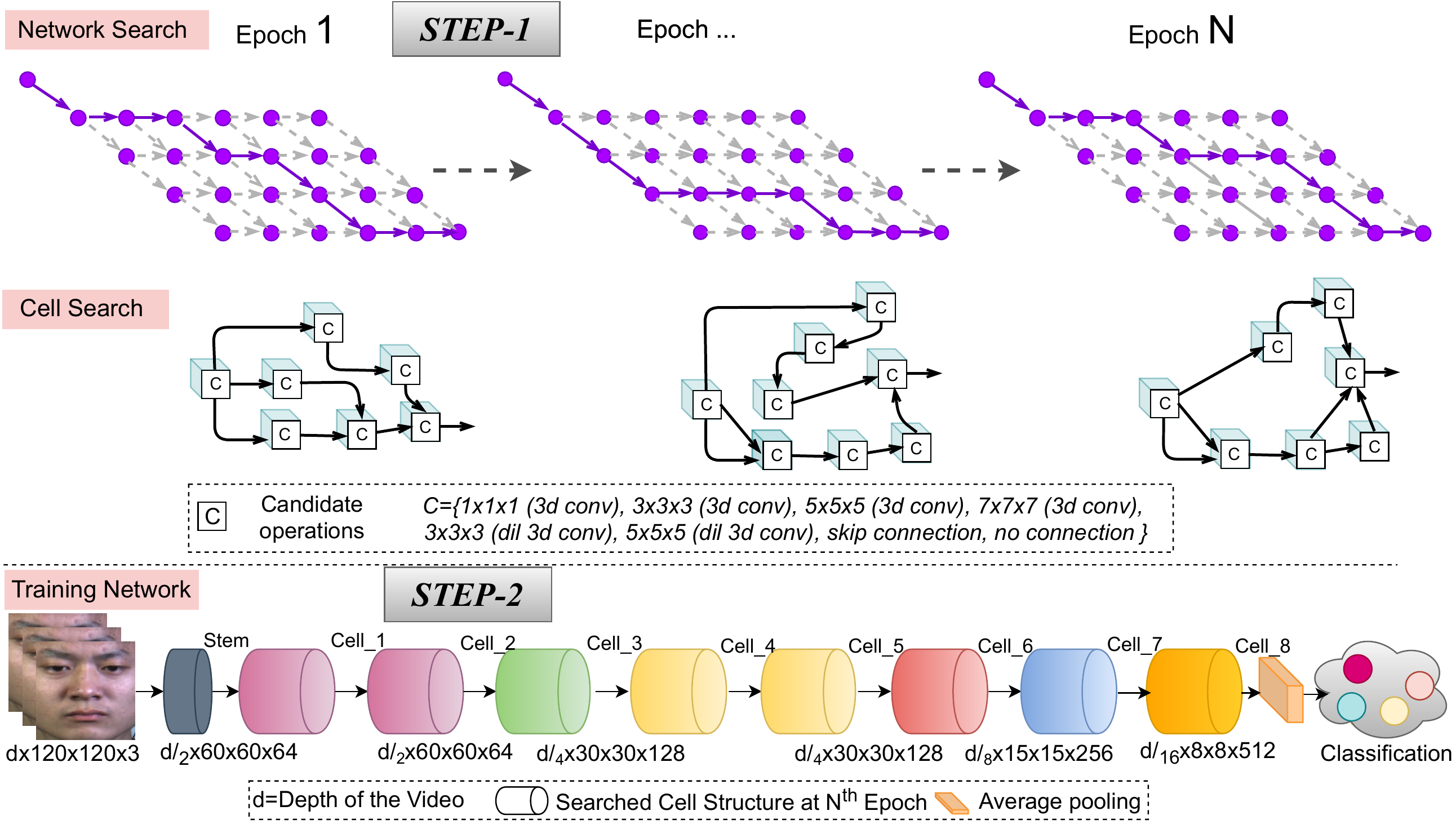}
         \caption{}
         \label{fig:8b}
     \end{subfigure}
     
        \caption{End-to-end methods (a) teacher-student based macro-to-micro knowledge transferring [M:18] and (b) NAS based framework for MER [M:81]}
        \label{fig:8}
\end{figure}

\IEEEraisesectionheading{\section{Introduction}\label{sec:introduction}}
This document first summarizes the computational complexity requirement of existing DL-based MER approaches in terms of total number of parameters and execution time in Table \ref{param}. Further,  Table \ref{stas} and Fig. \ref{datasamples} represents the data samples of the all MER datasets and statistical details of them, respectively. Further, the traditional handcrafted MER approaches are summarized and indexed in Table \ref{tab:1}. The documents detailed the differences between macro and micro expression. The document represents the different types of deep learning based MER approaches. The 2D-CNN based architecture are depicted in Fig. \ref{fig:4a} and \ref{fig:4c}. Moreover, these architectures also represents three different input formats: dynamic imaging, landmark points, affective motion imaging, applied to process the spatio-temporal data through two-dimensional convolutions networks. Similarly Fig. \ref{fig:4b} and Fig. \ref{fig:4c}, demonstrated the multi-stream and -scale based CNN architectures, respectively. The Fig. \ref{fig:5a} and \ref{fig:5b} represents the examples of capsule module and recurrent convolutional network (RCN) based multi-stage MER frameworks. Moreover, 3D-CNN and graph based MER frameworks are shown in the Fig. \ref{fig:6a} and \ref{fig:6b}. The recent evolving techniques: GAN, Student-Teacher and NAS for MER are demonstrated in Fig. \ref{fig:7}, \ref{fig:8a} and \ref{fig:8b}, respectively. Moreover, we discuss the differences between existing surveys and the proposed surveys in detail. More detailed comparison with proposed survey and recent survey [M:47] is tabulated in Table \ref{tab2}. 
\section{Macro vs Micro Expressions}
MEs occur with very low facial movements and intensities due to their subtle and involuntary nature. MEs emerge when a person tries to hide genuine emotions within their visage. However, MEs leak enough cues to understand humans' genuine emotions and deceitful behavior. In contrast, macro expressions (MaEs) exhibit expressive variations for a longer duration of 0.5 to 04 seconds on the face. While the MEs are swift in nature and do not last more than 1/25th seconds on the face \cite{matsumoto2011evidence, yan2013fast}. Therefore, developing generalized solutions for both MaEs and MEs is non-identical in terms of data accumulation, data processing, recognition model designing, etc. In literature, MaEs datasets emerged from scripted or posed (in-lab environments). The sample pose data sets are CK+, JAFFE, MUG, MMI, OULU, ISED, GEMEP-FERA, DISFA, etc. Further, the following datasets are developed to deal with real-world candid environments (in-wild)  AFEW, AffectNet, RAF-DB, FER, etc. However, the data collection/generation for MEs is challenging due to its dynamic nature, lack of experts in label generation, etc. Although, there are very few MEs datasets are publicly available in the literature. The MEs data sets are further classified into spontaneous and expression in the wild. The spontaneous ME datasets are SMIC \cite{li2013spontaneous}, CASME-I \cite{yan2013casme}, CASME-II \cite{yan2014casme}, MEVIEW \cite{husak2017spotting}, CAS(ME)$^2$ \cite{husak2017spotting}, SAMM \cite{davison2016samm} and MMEW \cite{ben2021video}. The spontaneous expression of the subject is captured while the subjects were watching videos. Also, the wild ME dataset is MEVIEW. Due to the above-said challenges, the ME datasets hold minimal and imbalanced data samples. Therefore, designing the deep CNN networks for MEs is challenging over MaEs. Also from the literature, it is evident that CNN models outperform while large sample sets are provided during the training. However, the imbalanced datasets with limited samples, and the subtle, rapid nature of the MEs, make it harder to design a model to capture micro variations on the face. Therefore, shallow CNN architectures are proposed in the literature by considering time-series data.  Whereas in the case of MER either shallow or deep CNN architectures are proposed based on the characteristics of the data. 
\section{Comparative Analysis of Existing and Proposed Survey}
There are very few surveys published in reputed journals as reported in Table. 1 (Manuscript). All most all surveys are focused more on the applications, datasets, preprocessing techniques, handcrafted descriptors based micro-expressions spotting, and feature extraction techniques. Moreover, surveys [M:35, 54, 29] published in the 20’s have aimed to solve some of the issues of deep learning MER approaches such as overfitting, data unbalancing, micro-to-macro feature adaption, recognition based on key apex frame and action units.  The recently published survey [M:42] in TPAMI is also focused on key differences between macro-and micro-expressions, video-based micro-expression analysis in a cascaded structure, encompassing the neuropsychological basis, datasets, features, spotting algorithms, recognition algorithms, applications, and evaluation of state-of-the-art approaches. The TPAMI paper is more dedicated to highlighting the issues in existing datasets and proposed a new MMEW dataset. The detailed technical comparison between TPAMI and proposed survey is as follows:\\
\textbf{\textit{TPAMI}}
\begin{enumerate}
\item  Section 1: Starting with the introduction and a brief discussion of Challenges and differences from conventional techniques. Further contributions point out.
\item	In Section 2: The key difference between macro and micro-expressions is defined in terms of the nature of muscles, which provoked emotions, Neuropsychology procedure, and duration.
\item	In Section 3: The detailed study of existing datasets, their properties, and challenges has been discussed. Based on challenges a new dataset MMEW has been proposed.
\item	In Section 4: Different types of features like dynamic texture (DT), frequency domain,  DT in tensor-domain space, optical flow features, deep features, feature interpretability has been discussed. This section is more towards the dynamic texture feature using handcrafted descriptors and very small details are noted down about the deep learning features.
\item	In Section 5: The traditional (handcrafted) ME spotting algorithms are presented.
\item	Section 6: presents a brief description of recognition algorithms by including traditional classifiers and deep learning. Further a short information about transfer learning is discussed.
\item	Section 7: represents the application of the MER.
\item	Section 8: Compare the performance of the proposed dataset MMEW over existing approaches (most of the algorithms are handcrafted) and indexed the base results for the same.
\item	In Section 9-10: future directions: Privacy-protection micro-expression analysis, Utilizability of macro-expressions for micro-expression analysis, Standardized datasets, Data augmentation, GAN-based sample generation, Multi-task learning, Explainable micro-expression analysis.
\end{enumerate}
However, the proposed survey highlights the comparative study of existing deep learning MER frameworks and their characteristics: \\

\textbf{\textit{Proposed}}
\begin{enumerate}
\item	Section 1: Starting with the introduction, steps involved in any MER approaches like pre-processing, feature extraction, and classification have been discussed in brief. Further, the differences in macro and micro-expressions in terms of muscles nature, durations, challenges in data collections, challenges in deep learning models’ designing are highlighted. Also, how the proposed survey is different from the existing surveys for MER is discussed in the introduction.
\item	In Section 2: A detailed survey has been presented by categorizing DL-based MER into four broad categories as multi-stage, end-to-end, transfer learning and GANs based MER. Based on technical characteristics multi-stage and end-to-end MER frameworks are also divided into subcategories as 2D-CNN, Multi-Stream, Multi-Scale, Capsule, RCNs, and Multi-Scale, CNN-LSTM, 3D-CNN, Graph based, NAS based MER approaches with pros and cons of each model. The survey collected the performance majors of existing DL based MER approaches (Table 3-7) and studied all technical aspects (two-stage vs end-to-end frameworks, the impact of downsampling with convolution and pooling, the impact of multi-scale/-stream and single-scale/linear CNN frameworks, does deeper network affect the performance of the MER?, does kernel sizes have any impact on CNN layers?) of the deep learning approaches and their effects on the performance of the MER, which will help the new researchers to design and develop robust MER frameworks.
\item	Section 3: represents the training and evaluation strategy. Initially, based on environmental challenges: lab and in-wild environments, MEs datasets are discussed. Further, the performance evaluation metrics and their significance in MER have been reported. Moreover, the various validation strategies: LOSO, LOVO, Composite, Cross-Domain, random split are discussed by broadly categorizing into three categories: PDE, PIE, and CDE. Finally, the effects of experimental settings: data augmentation, validation strategies, input selection, and the number of expression classes, over the performance of MER, are studied. 
\item	In Section 4: some critical issues in MER also point out which require the attention of researchers to improve the efficacy of the MER approaches. Moreover, future directions such as Group Emotion Recognition, Motion Magnification in MER, Multi-modal in MER, MER in-wild, MER in Psychological Disorders, are presented.
\item	In Section 5-6: future directions: are discussed  

The key difference between TPAMI and our proposed survey is detailed out in Supplementary document-Table \ref{tab2}.

\end{enumerate}

\begin{table*}[!t]
\centering
\caption{The key difference between TPAMI [M:42] and our proposed survey.}
\label{tab2}
\begin{tabular}{cc}
\hline
   \textbf{Proposed}                                                                                                                           & \textbf{TPAMI}                                                                                                                       \\ \hline \hline
 \begin{tabular}[l]{@{}l@{}}1. The detailed survey of deep learning MER approaches has\\ been done by categorizing them into four major categories:\\ multi-stage, end-to-end, transfer learning and GANs based.\\ These categories are further divided  into sub-categories such\\ as 2D-CNN, multi-stream/scale, capsule, 3D-CNN,  CNN-\\LSTM, Graph-based, etc., with benefits and limits of each\\ model.\end{tabular}  
 & \begin{tabular}[l]{@{}l@{}}1. Few deep learning MER approaches and transfer learning \\ approaches are explained in this.\end{tabular}      \\ \hline
 
 \begin{tabular}[l]{@{}l@{}}2. The survey collected the performance majors of existing \\ DL based MER  approaches (Table 3-7) and studied all \\ technical aspects (two-stage vs end-to-end frameworks, \\ the impact of downsampling with convolution and pooling\\, impact of multi-scale/-stream and single- scale/linear\\ CNN frameworks, does deeper network affect the perform-\\ance of the MER?, does  kernel sizes have any impact on \\CNN layers?) of the deep learning approaches and their eff-\\ects on the performance of the MER,  which will help  the new \\researchers to design and develop  robust MER frameworks.\end{tabular}    
 & \begin{tabular}[l]{@{}l@{}}2. There is no discussion about the technical characteristics \\ of the DL based MER approaches.\end{tabular}          \\ \hline
 
 \begin{tabular}[l]{@{}l@{}}3. The survey provides a detailed explanation of the evalua-\\tion metrics (accuracy, F1-score, Recall, UF1, UAR, confusion \\matrix) and validation strategies (PDE: LOVO, 80/20 split,\\ PIE: LOSO, composite LOSO, CDE), which plays a significant\\ role in MER approaches.\end{tabular}                                            & \begin{tabular}[l]{@{}l@{}}3. There is no discussion about evaluation metrics and vali-\\ dation strategies.\end{tabular}                     
 \\ \hline
 
 \begin{tabular}[l]{@{}l@{}}4. In literature, there are no standard protocols to evaluate \\the MER approaches. The MER approaches follow contrast\\ experimental setup in terms of total validation protocols,\\ number of samples, input selection, participants, expression \\classes,  etc., or dropped some of the emotion classes due to\\ a smaller number of images  [M:13],  [M:56].  Therefore, it is hard\\ to compare these approaches directly due to both positive \\and negative impacts on the  MER  performance.  Therefore,\\ in this survey, we have done a detailed study based on expe-\\rimental results of existing literature to analyses the effects \\of different evaluation setups:  PDE,  PIE,  and  CDE,  and other\\ experimental settings:  data augmentation,  input selection,\\ number of emotion classes, etc., over the performance of \\the MER approaches. Thus, new researchers will get awar-\\eness about the selection of experimental settings in DL-\\based MER.
 \end{tabular}  &  \begin{tabular}[c]{@{}l@{}}4. There is no discussion about the effects of experimental \\settings and protocols in this survey.\end{tabular} 
 \\ \hline
 
 \begin{tabular}[l]{@{}l@{}}5. The survey is briefly discussed the MEs datasets by categ-\\orizing them into two parts: lab and in-wild environment.\end{tabular}                                                                           & \begin{tabular}[c]{@{}l@{}}5. The survey is more towards MEs datasets, issues in exis-\\ting  datasets, and proposed a new data warehouse by in-\\corporating macro and micro-expressions.\end{tabular}               \\ \hline
 
 \begin{tabular}[l]{@{}l@{}}6. The survey detailed out the difference between macro and\\ micro in terms of muscles nature, duration, challenges in data\\ collections, challenges in deep learning models’ designing.\end{tabular}                                          & \begin{tabular}[l]{@{}l@{}}6. The survey detailed out the difference between macro \\and micro in terms of the nature of muscles, which prov-\\oked emotions, Neuropsychology procedure, and duration.\end{tabular}   \\ \hline
 \begin{tabular}[l]{@{}l@{}}7. The survey does not aim at conventional MER approaches\\ and briefly explained the handcrafted techniques.\end{tabular}                                                 & \begin{tabular}[l]{@{}l@{}}7. The survey categorize the features and aimed to explain \\more about the handcrafted features: DT, frequency doma-\\in, DT in tensor-domain space,optical flow features, deep\\ features, feature interpretability, instead of deep learning MER\\ approaches.\end{tabular}                                                      \\ \hline
 \begin{tabular}[l]{@{}l@{}}8. This survey is dedicated to deep learning-based MEs class-\\ification approaches only.   \end{tabular}    & \begin{tabular}[l]{@{}l@{}}8. The survey detailed the handcrafted MEs spotting appro-\\aches. \end{tabular}   \\ \hline
 
 \begin{tabular}[l]{@{}l@{}}9. This survey considered all the latest deep learning MER ap-\\proaches (2013-2021).\end{tabular}                   & \begin{tabular}[l]{@{}l@{}}9. This survey consider only few paper until early 2019\\ and does not include latest deep learning based MER\\ approaches such as:\\ MERTA-2019 [M:38], NMER-2019 [M:45], STSTNet-2019 [M:39]\\DSSN-2019 [M:32],
3D-Flow-2019 [M;46], OffApex-2019 [M:34]\\Dual Inc-2019 [M:12], DMERKD-2020 [M:52]\\OrigiNet-2021 [M:31],
AffectiveNet-2021 [M:30], LGConD-2021 [M:11] \\and many more as shown in Fig. 1 of the manuscript.
\end{tabular}                                           \\ \hline 
\end{tabular}

\end{table*}

\bibliographystyle{IEEEtran}
\bibliography{egbib}